\newif\ifconf
\newif\ificlrfinal

\conffalse

\ifconf
\documentclass{article} 
\usepackage{iclr2027_conference,times}

\usepackage{amsmath,amsfonts,bm}

\def\eqref#1{equation~\ref{#1}}

\def\1{\bm{1}}

\DeclareMathAlphabet{\mathsfit}{\encodingdefault}{\sfdefault}{m}{sl}
\SetMathAlphabet{\mathsfit}{bold}{\encodingdefault}{\sfdefault}{bx}{n}

\usepackage{hyperref}
\usepackage{url}
\usepackage{booktabs}
\usepackage{graphicx}
\usepackage{wrapfig}
\usepackage{listings}
\usepackage{subcaption} 

\usepackage{enumitem}

\title{Hardware-Aware Features for CUTLASS\\Kernel Selection} 

\author{Antiquus S.~Hippocampus, Natalia Cerebro \& Amelie P. Amygdale \thanks{ Use footnote for providing further information
about author (webpage, alternative address)---\emph{not} for acknowledging
funding agencies.  Funding acknowledgments go at the end of the paper.} \\
Department of Computer Science\\
Cranberry-Lemon University\\
Pittsburgh, PA 15213, USA \\
\texttt{\{hippo,brain,jen\}@cs.cranberry-lemon.edu} \\
\And
Ji Q. Ren \& Yevgeny LeNet \\
Department of Computational Neuroscience \\
University of the Witwatersrand \\
Joburg, South Africa \\
\texttt{\{robot,net\}@wits.ac.za} \\
\AND
Coauthor \\
Affiliation \\
Address \\
\texttt{email}
}

\else

\documentclass[sigconf,nonacm]{acmart}
\acmISBN{978-1-4503-XXXX-X/2018/06}

\usepackage{makecell}

\definecolor{vlgray}{rgb}{0.77 0.77 0.77}
\definecolor{ablack}{rgb}{0.2 0.2 0.2}

\usepackage{tikz}
\usetikzlibrary{tikzmark}

\usepackage{xpatch}
\expandafter\xpatchcmd
\csname pgfk@/tikz/every picture/.@cmd\endcsname
{\thepage}{\arabic{page}}{}{}

\usepackage{fontawesome}
\usepackage{pifont}
\usepackage{textcomp}
\usepackage{lipsum}

\usepackage{hyperref}
\usepackage{url}
\usepackage{booktabs}
\usepackage{graphicx}
\usepackage{wrapfig}
\usepackage{listings}
\usepackage{subcaption} 
\usepackage{enumitem}

\iclrfinalfalse

\newfloat{lstfloat}{htbp}{lop}
\floatname{lstfloat}{Listing}

\usepackage{hyphenat} 

\title{Hardware-Aware Features for CUTLASS Kernel Selection}

\author{Shriram Chandran$^{*\dagger}$}
\affiliation{%
  \institution{ETH Zurich}
  \city{Zurich}
  \country{Switzerland}
}

\author{Dominic Rinderer$^{\dagger}$}
\affiliation{%
  \institution{ETH Zurich}
  \city{Zurich}
  \country{Switzerland}
}

\author{Yakup Budanaz}
\affiliation{%
  \institution{ETH Zurich}
  \city{Zurich}
  \country{Switzerland}
}

\author{Alexandru Calotoiu}
\affiliation{%
  \institution{ETH Zurich}
  \city{Zurich}
  \country{Switzerland}
}

\author{Marcin Copik}
\affiliation{%
  \institution{ETH Zurich}
  \city{Zurich}
  \country{Switzerland}
}

\author{Torsten Hoefler}
\affiliation{%
  \institution{ETH Zurich}
  \city{Zurich}
  \country{Switzerland}
}

\fi

\newcommand{\citec}[1]{\ifconf(\cite{#1})\else\cite{#1}\fi}

\ifconf
\fi

\begin{document}

\begin{abstract}
GPU libraries such as CUTLASS expose tens of thousands of semantically equivalent kernels for a single operation, making exhaustive autotuning expensive and execution-free selection difficult. Existing analytical selectors require hand-designed performance rules, while learned selectors operate on raw configuration parameters and must infer hardware consequences from data. We introduce a hardware-aware representation for CUTLASS kernel selection that augments candidate configurations with statically computable estimates of induced hardware behavior. We construct a dataset of 4.9 million CUTLASS kernels and train gradient-boosted and neural learning-to-rank models to rank candidates within each problem. 
On held-out exhaustive evaluation problems, hardware-aware representations reduce selection regret by up to $40$\% relative to structural baselines and $64.2\%$ relative to NVIDIA's matrix-multiply heuristics. 
%
We further evaluate data-efficient cross-precision and epilogue-fusion transfer within CUTLASS GEMM, showing that explicitly representing candidate-induced hardware behavior provides a useful inductive bias for learned kernel selection.
\end{abstract}

\ifconf
\else
\include{mac-includes}
\maketitle
\begingroup
\renewcommand{\thefootnote}{}
\footnotetext{\raggedright $\dagger$ Equal contribution. $\ast$ Corresponding author.}
\addtocounter{footnote}{-1}
\endgroup
\fi

\section{Introduction}

Modern GPU libraries expose increasingly large spaces of implementations for a single operation. CUTLASS~\citec{nvidia_cutlass} is a particularly expressive instance of this problem -- a CUTLASS GEMM may be implemented in over sixty thousand ways using different tile shapes, instruction shapes, pipeline depths, schedules, cluster configurations, tile schedulers, and epilogue implementations, interacting non-trivially with the problem dimensions, layouts, data types, accumulation types, and target architecture~\citec{CUTLASS368:online}. The best configuration is determined by the interaction of these effects rather than by any one parameter in isolation. This flexibility is valuable to kernel developers, but it also creates a very large configuration space. Selecting an efficient implementation is therefore a central requirement for achieving near-peak performance on contemporary accelerators~\citec{bikshandi2023developing}. Empirical autotuning, by compiling and benchmarking candidate kernels, is expensive~\citec{Improvin92:online, bai2025learned, swann2025tritonblas}. Compilation and measurement must be repeated for each target GPU, datatype, epilogue, and problem shape. Empirical selection becomes impossible when a library requires several different GEMM shapes or must make selections dynamically. This motivates selectors that can rank candidates without executing them~\citec{Improvin92:online}. 

Existing execution-free selectors follow two main approaches. Analytical methods encode manually designed performance rules, such as preferences for tile sizes, occupancy limits, or wave quantization based on the problem instance~\citec{swann2025tritonblas, zhang2026wavetune}. These methods are fast but brittle, require substantial architecture-specific knowledge for development, and must be updated as hardware and kernel libraries evolve. Learned autotuners~\citec{ryu2021metatune, zhai2023tlp} reduce the need for manual rules, but typically represent candidates using raw problem dimensions and template parameters, leaving the model to then infer how those parameters affect hardware behaviour. Thus, raw parameters are a poor description of these effects.

We propose hardware-aware representations for kernel selection. Each candidate is represented using its structural configuration together with inexpensive estimates of its induced hardware behaviour. We construct a 4.9M-kernel training corpus, train ranking models, score the entire valid catalogue at inference, and obtain 6.2\% mean selection regret, compared with 17\% for nvMMH (nvMatmulHeuristics, NVIDIA's matrix multiply heuristics library), and 71.5\% for random selection -- a 64.2\% and 91.3\% relative reduction in selection regret, respectively.

%
Our approach combines the strengths of analytical selectors and measurement-driven learning: hardware behavior is exposed explicitly through static, mechanistic features, while the model learns how these effects interact from measured data rather than relying on a hand-written performance model.
This gives the selector the best of both worlds: strong architectural inductive bias without brittle decision rules, and data-driven adaptability without forcing the model to rediscover hardware behavior from raw parameters.
As a result, the same representation supports accurate selection, improved data efficiency, and transfer across related kernel spaces, establishing hardware-aware feature design as a practical foundation for learned performance modeling on modern accelerators.

\ifconf
\else 
\section{Background}
\label{sec:app-space}

\subsection{GPU Execution Model}

An NVIDIA GPU consists of many streaming multiprocessors (SMs), each containing the compute units and on-chip storage used for execution. A kernel launches a grid of thread blocks, and each block is assigned to an SM independently. Threads within a block execute in groups of 32 called warps. Warps can cooperate through on-chip shared memory and synchronization, while registers are private to individual threads. The output of a kernel is therefore determined not only by the arithmetic performed, but also by how work, data, and synchronization are distributed across threads, warps, blocks, and the memory hierarchy.

For GEMM, the main data movement path is from global memory through shared memory to registers. A thread block typically loads tiles of the input matrices into shared memory, computes on those tiles using registers as accumulators, and eventually stores its output tile back to global memory. Shared memory provides lower-latency reuse within a block, while the L2 cache and device memory serve data shared across blocks and data that does not fit on chip. The amount of work and data assigned to a block determines its arithmetic intensity, memory traffic, shared-memory footprint, register usage, and ability to keep the SM occupied.

Only a limited number of blocks can reside on an SM at once. This limit is determined by the block’s register and shared-memory requirements, its thread count, and architectural limits on blocks and warps. Higher residency can help hide memory and instruction latency, but increasing the amount of work assigned to a block often increases resource usage and can reduce the number of simultaneously resident blocks. Consequently, the best GEMM configuration must balance reuse and computational efficiency against resource pressure.

\subsection{Hopper Execution Mechanisms}

Our primary evaluation targets NVIDIA Hopper GPUs, whose SM90a architecture introduces several mechanisms that substantially affect GEMM performance~\citec{luo2024benchmarkingdissectingnvidiahopper}. Hopper tensor-core operations are issued using warpgroup matrix multiply-accumulate (WGMMA) instructions, comprising four consecutive warps, or 128 threads, cooperating to issue an asynchronous matrix operation. The instruction shape and warpgroup organization impose further constraints on the tile shapes that a kernel can use. Hopper also provides the Tensor Memory Accelerator (TMA), which asynchronously transfers multidimensional tensor regions between global memory and shared memory, saving several individual thread address computations and data loads.

\ifconf
\else
\begin{figure}
\centering
\label{fig:overview}
\includegraphics[width=\linewidth]{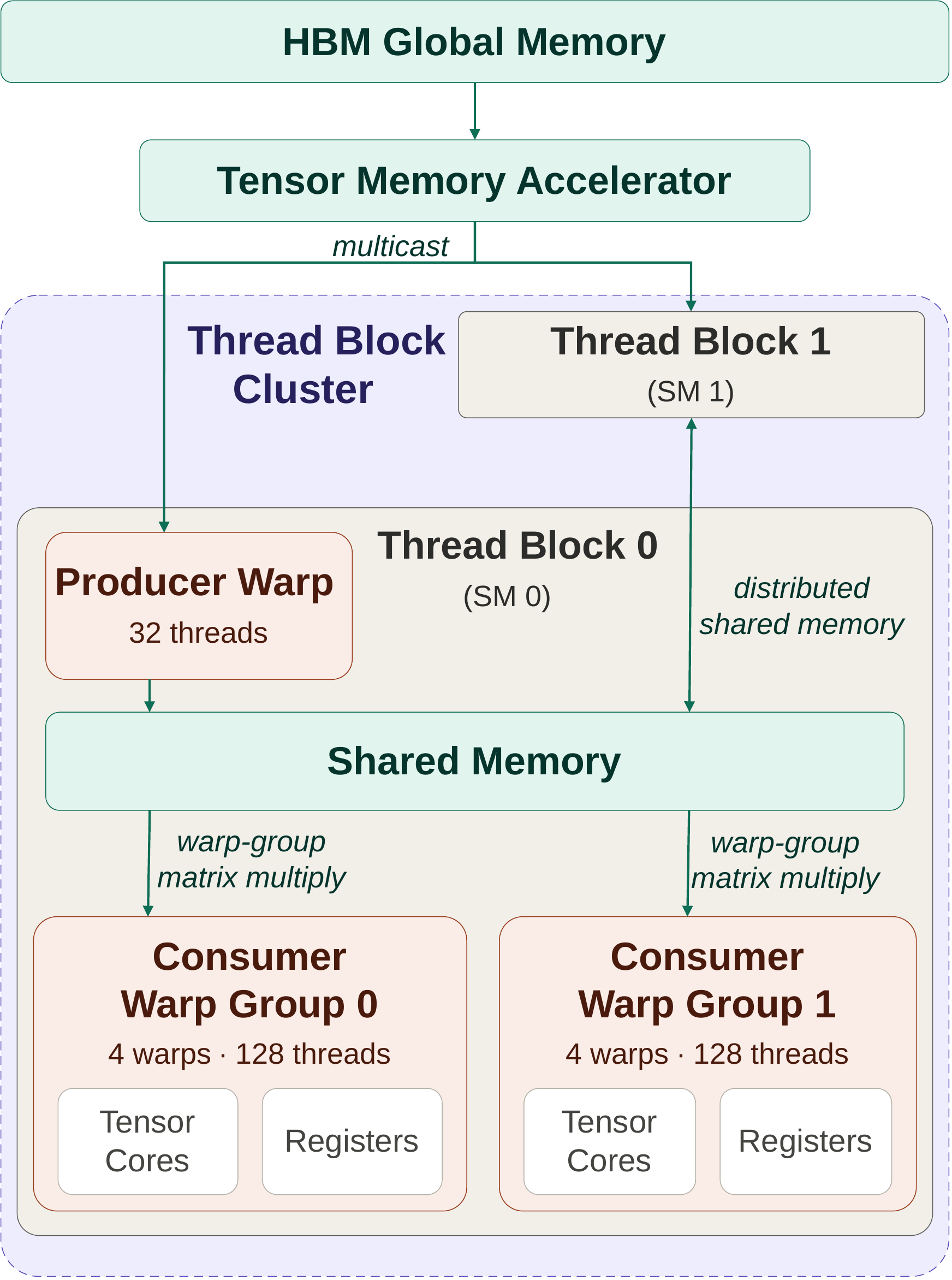}
\caption{H100 Programming Model}
\end{figure}
\fi

\ifconf
\begin{wrapfigure}{r}{0.40\textwidth}
\centering
\label{fig:overview}
\includegraphics[width=0.40\textwidth]{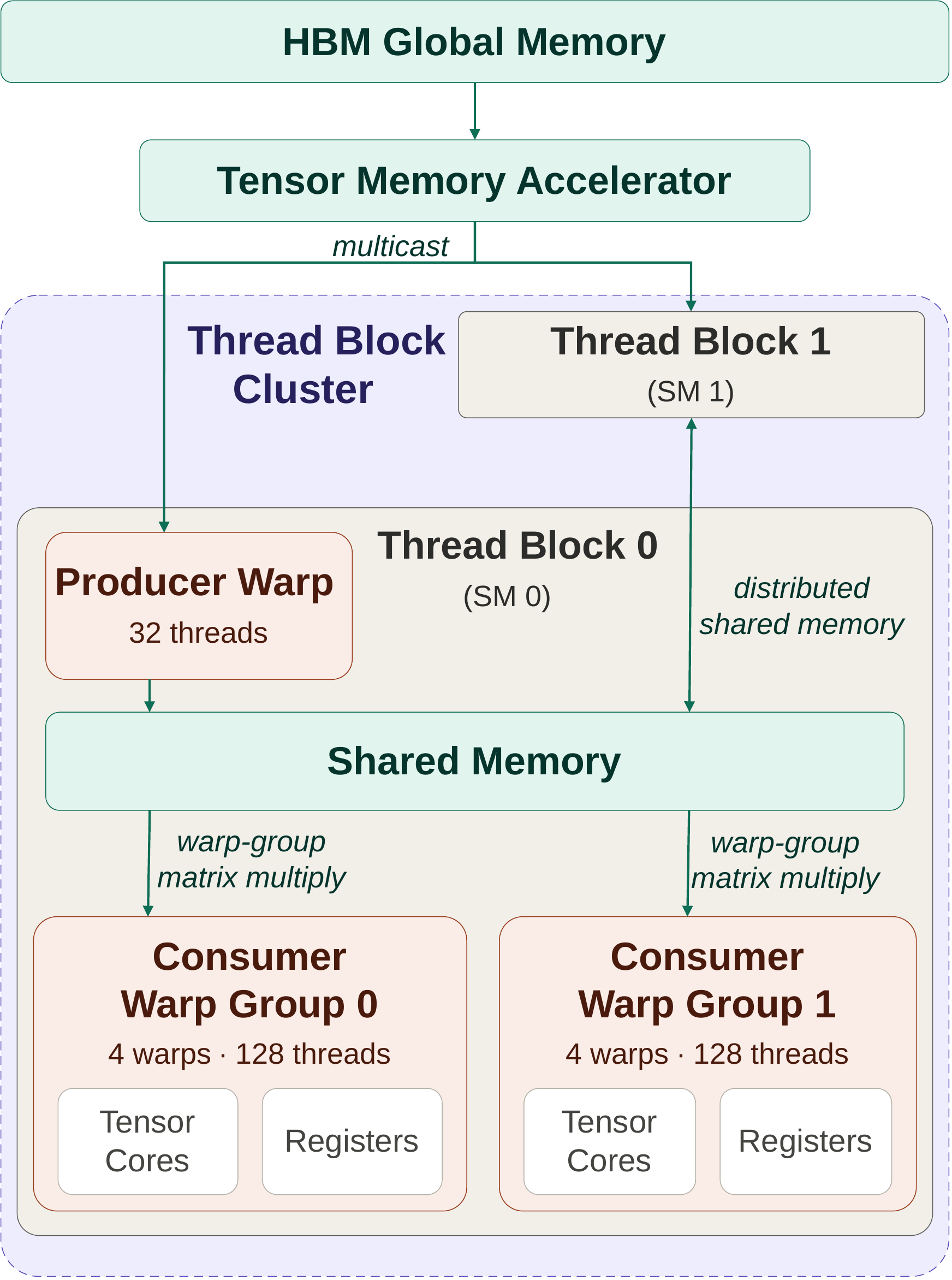}
\caption{H100 Programming Model}
\end{wrapfigure}
\else
\fi

TMA and WGMMA enable warp-specialized pipelines~\citec{soi2025optimalsoftwarepipeliningwarp}, where different warps assume different duties. Producer warps issue asynchronous TMA transfers, while consumer warpgroups perform WGMMA on previously loaded tiles. The shared-memory buffer is divided into a multi-stage pipeline, allowing memory movement to run ahead of computation. Increasing the number of stages can hide transfer latency but increase shared-memory consumption.

Hopper additionally supports thread-block clusters, which can cooperate through distributed shared memory, or through TMA multicast which can deliver the same data to multiple blocks. Clusters can reduce memory traffic but impose scheduling and residency constraints. Whether clustering is beneficial therefore depends on the problem shape, tile decomposition, and resource requirements of the candidate kernel.

These mechanisms make the relationship between a CUTLASS configuration and hardware performance highly indirect. A configuration parameter such as pipeline depth, schedule, or cluster shape changes several hardware-level properties simultaneously. The resulting effects cannot generally be inferred from the parameter value alone.

\subsection{GEMM Tiling and CUTLASS}

We consider the general matrix multiplication
$
C = \alpha AB + \beta C,
$
where $A \in \mathbb{R}^{M \times K}$, $B \in \mathbb{R}^{K \times N}$, and $C \in \mathbb{R}^{M \times N}$. The operation performs ($2MNK$) floating-point operations. A GPU implementation partitions $C$ into output tiles. A thread block computes one or more such tiles by iterating over the reduction dimension in chunks.

A candidate configuration is characterized primarily by a tile shape $(T_M,T_N,T_K)$. The first two dimensions determine the tile computed by a block, while $T_K$ determines the amount of the reduction processed in one mainloop iteration. Each iteration loads an $T_M \times T_K$ tile of $A$ and a $T_K \times T_N$ tile of $B$, performs a matrix multiply-accumulate, and advances to the next $K$-tile. The number of iterations is thus $\lceil K/T_K\rceil$. The tile shape therefore affects the number of output tiles, boundary tile waste, effective arithmetic intensity, memory traffic, shared-memory and register usage, and the number of mainloop iterations.
\ifconf
\else
For one shape, our BF16 enumerator exposes $534{,}600$ raw combinations per layout before validity filtering, and $2{,}138{,}400$ kernels throughout.
\fi

CUTLASS builds GEMM kernels by composing reusable mainloop and epilogue components. The mainloop loads input tiles and performs the matrix multiplication, while the epilogue applies operations such as scaling, bias addition, activation, residual addition, type conversion, and the final store. CUTLASS exposes all aforementioned configurations as template parameters, and 
these choices are coupled -- for e.g., a larger tile may increase data reuse but increase register and shared-memory usage. More pipeline stages can make the pipeline more efficient, but consume additional resources and reduce occupancy. A schedule may require a particular tile divisibility or warpgroup arrangement. Clusters may reduce redundant loads while limiting which blocks can be resident together. Epilogues add their own computation, memory traffic, and resource requirements, and can change which mainloop configuration is best. %
Not 
%
every combination of choices produces a valid kernel. CUTLASS implicitly imposes architectural and implementation constraints on several of the parameters. The valid candidate set is therefore a tiny constrained subset of the cartesian product of the parameters. 
\ifconf
\else
Static CUTLASS legality and shared-memory checks
retain around $11.36\%$, leaving a catalogue of around
$61{,}000$ per layout
($243{,}000$ throughout), on the basis of which the selector must therefore rank candidates.

\fi
The configuration does not directly specify performance -- making selecting an efficient candidate from this valid set highly non-trivial.
\ifconf
\else
On a GH200, measured CUTLASS throughputs on BF16 GEMMs in our sweep span
$0.01$--$763$\,Tflop/s depending on problem size and configuration.
For one given GEMM shape and data layout, the empirical best and worst
valid kernels can differ by a factor of $1{,}070\times$ in throughput (median across $68$ evaluated groups; $3.9$\,million measured pairs),
a uniform random pick achieves only $28.5\%$ of the in-group oracle on average, and choosing a static best configuration only $71\%$. Less than $2.4\%$ of configurations get performance within $5\%$ of the best kernel on average.
Mis-selection is therefore not a rounding error; it can waste most of the hardware budget on
every GEMM in a compiled program. Exhaustively compiling and benchmarking the full catalogue is impractical at scale:
our evaluation set contains only $17$ shapes, yet requires a benchmarking sweep over 
the $3.9$\,million distinct (kernel, shape) GPU measurements
($68$ unique groups, up to $61{,}000$ candidates per group).
Compiling the heavily templated configuration catalogue once alone costs on the order of hours of CPU time
even before any GPU benchmarking begins.
\fi

\subsection{A Generated Kernel}

\ifconf
The snippet below is a representative CUTLASS kernel: BF16 operands with FP32 accumulation, tile $128\times128\times64$, a $1\times1$ cluster, four mainloop pipeline stages, the plain TMA warp-specialized mainloop, a TMA warp-specialized epilogue, persistent tile scheduling, and the \texttt{TN} operand layout (A row-major, B column-major). The configuration name is recorded in the generated comment header.
\fi

\definecolor{codegreen}{rgb}{0,0.6,0}
\definecolor{codegray}{rgb}{0.5,0.5,0.5}
\definecolor{codepurple}{rgb}{0.58,0,0.82}
\definecolor{backcolour}{rgb}{0.95,0.95,0.92}

\lstdefinestyle{mystyle}{
    backgroundcolor=\color{backcolour},
    commentstyle=\color{codegreen},
    keywordstyle=\color{magenta},
    numberstyle=\tiny\color{codegray},
    stringstyle=\color{codepurple},
    basicstyle=\ttfamily\scriptsize,
    breakatwhitespace=false,
    breaklines=true,
    captionpos=b,
    keepspaces=true,
    numbers=left,
    numbersep=5pt,
    showspaces=false,
    showstringspaces=false,
    showtabs=false,
    tabsize=2
}

\lstset{style=mystyle}

\begin{lstlisting}[language=C++]
// ... unrelated generated boilerplate omitted (includes, CUDA/CUTLASS checks) ...

// Config: cutlass3x_sm90_tensorop_bf16_bf16_f32_bf16_bf16_
//         128x128x64_1x1x1_4_tnn_align64_warpspecialized_epi_tma

namespace Kernel_0 {

    using           ElementA    = cutlass::bfloat16_t;
    using           LayoutATag  = cutlass::layout::RowMajor;
    constexpr int   AlignmentA  = 64;

    using           ElementB    = cutlass::bfloat16_t;
    using           LayoutBTag  = cutlass::layout::ColumnMajor;
    constexpr int   AlignmentB  = 64;

    using           ElementD    = cutlass::bfloat16_t;
    using           ElementC    = cutlass::bfloat16_t;
    using           LayoutCTag  = cutlass::layout::ColumnMajor;
    using           LayoutDTag  = cutlass::layout::ColumnMajor;
    constexpr int   AlignmentD  = 64;
    constexpr int   AlignmentC  = 64;

    using ElementAccumulator    = float;
    using ElementCompute        = float;
    using ArchTag               = cutlass::arch::Sm90;
    using OperatorClass         = cutlass::arch::OpClassTensorOp;

    using MmaTileShape          = Shape<_128, _128, _64>;
    using ClusterShape          = Shape<_1, _1, _1>;
    using StageCount            = cutlass::gemm::collective::StageCount<4>;
    using TileSchedulerType     = cutlass::gemm::PersistentScheduler;

    using MainloopScheduleType  = cutlass::gemm::KernelTmaWarpSpecialized;
    using EpilogueScheduleType  = cutlass::epilogue::TmaWarpSpecialized;
    using EpilogueTile          = cutlass::epilogue::collective::EpilogueTileAuto;

    using CollectiveEpilogue = typename cutlass::epilogue::collective::CollectiveBuilder<
        ArchTag, OperatorClass,
        MmaTileShape, ClusterShape,
        EpilogueTile,
        ElementAccumulator, ElementAccumulator,
        ElementC, LayoutCTag, AlignmentC,
        ElementD, LayoutDTag, AlignmentD,
        EpilogueScheduleType
    >::CollectiveOp;

    using CollectiveMainloop = typename cutlass::gemm::collective::CollectiveBuilder<
        ArchTag, OperatorClass,
        ElementA, LayoutATag, AlignmentA,
        ElementB, LayoutBTag, AlignmentB,
        ElementAccumulator,
        MmaTileShape, ClusterShape,
        StageCount,
        MainloopScheduleType
    >::CollectiveOp;

    using GemmKernel = cutlass::gemm::kernel::GemmUniversal<
        Shape<int,int,int,int>,
        CollectiveMainloop,
        CollectiveEpilogue,
        TileSchedulerType>;

    using Gemm = cutlass::gemm::device::GemmUniversalAdapter<GemmKernel>;
}

\end{lstlisting}

%


\ifconf
\else
\begin{table*}
\centering
\footnotesize
\setlength{\tabcolsep}{3pt}
\renewcommand{\arraystretch}{1.05}
\caption{Configuration knobs exposed by the BF16 SM90 generator (\texttt{generate\_search\_space}).}
\label{tab:config-knobs}
\begin{tabular}{@{}p{0.24\linewidth} p{0.36\linewidth} p{0.36\linewidth}@{}}
\toprule
Knob & Values / domain & Meaning \\
\midrule
\texttt{tile\_m} & $\{64,128,192,256\}$ & Thread-block $M$ tile extent \\
\texttt{tile\_n} & $\{16,32,48,64,80,96,128,192,256\}$ & Thread-block $N$ tile extent \\
\texttt{tile\_k} & $\{32,64,128,256,512\}$ & Thread-block $K$ tile / pipeline atom \\
\texttt{stages} & $\{2,\ldots,12\}$ & Mainloop pipeline stage count \\
\texttt{cluster\_m}, \texttt{cluster\_n} & $\{1,2,4,8,16\}$, $C_M C_N \le 16$ & CTA cluster shape in $M$ and $N$ \\
\texttt{cluster\_k} & fixed $1$ & $K$-clustering not searched \\
\texttt{kernel\_schedule} & WS / Pingpong / Cooperative & TMA warp-specialized mainloop policy \\
\texttt{epilogue\_schedule} & NoSmem / TMA / TMA-Cooperative & Epilogue store path \\
\texttt{scheduler} & Persistent / Stream-K & Tile scheduling across the output grid \\
\texttt{layout} & TN, TT, NN, NT & Operand layouts for $A$ and $B$ \\
\addlinespace
\multicolumn{3}{@{}l}{\textit{Fixed (not searched independently)}} \\
Operand / acc.\ types & BF16 in, FP32 acc.\ (BF16 sweep) & CUTLASS element aliases \\
\texttt{alignment\_*} & $128\,\text{bit}$ / element size & TMA vectorization requirement \\
$C/D$ layout & ColumnMajor & Row-major output via transpose trick \\
\texttt{ArchTag} / \texttt{OpClass} & Sm90 / TensorOp & Target architecture \\
\texttt{EpilogueTile} & \texttt{EpilogueTileAuto} & Sub-tile shape chosen by CUTLASS \\
\bottomrule
\end{tabular}
\end{table*}

\begin{table*}
\centering
\footnotesize
\setlength{\tabcolsep}{3.5pt}
\renewcommand{\arraystretch}{1.05}
\caption{Static validity constraints for BF16 SM90 GEMM configurations.}
\label{tab:validity-constraints}
\begin{tabular}{@{}p{0.48\linewidth} p{0.48\linewidth}@{}}
\toprule
Constraint & Reason \\
\midrule
Stream-K $\Rightarrow$ cooperative mainloop & CUTLASS Stream-K scheduler requires the cooperative TMA mainloop. \\
\addlinespace
Cooperative mainloop $\Rightarrow T_M \equiv 0 \pmod{128}$ & Two consumer warpgroups split the $M$ tile; each half must stay 64-row WGMMA aligned. \\
\addlinespace
$S_{\mathrm{mainloop}} + S_{\mathrm{epi}} + 2\,\mathrm{KiB} \le 227\,\mathrm{KiB}$ & Hopper shared-memory capacity per CTA; $S_{\mathrm{epi}}$ depends on epilogue schedule and tile via \texttt{estimate\_epilogue\_smem\_bytes}. \\
\addlinespace
If $A$ is column-major: $T_K \le 256\,C_N$ & TMA smem box $K$ extent is multicast-split across the $N$ cluster dimension. \\
\addlinespace
If $B$ is row-major: $T_K \le 256\,C_M$ & Symmetric $K$-box limit along the $M$ cluster dimension for the transposed operand. \\
\bottomrule
\end{tabular}
\end{table*}
\fi

\ifconf
\else
The snippet above is a representative CUTLASS kernel: BF16 operands with FP32 accumulation, tile $128\times128\times64$, a $1\times1$ cluster, four mainloop pipeline stages, the plain TMA warp-specialized mainloop, a TMA warp-specialized epilogue, persistent tile scheduling, and the \texttt{TN} operand layout (A row-major, B column-major). The configuration name is recorded in the generated comment header.
\fi

\sloppy Reading top to bottom: \texttt{ElementA/B/C} and \texttt{ElementAccumulator} fix the operand and accumulation types; \texttt{LayoutA/B} are the searched operand layouts while \texttt{LayoutC/D} are always column-major in our generator (row-major $C$ is obtained offline via the transpose--swap identity $C^\top=B^\top A^\top$). \texttt{AlignmentA/B/C} are derived as $128/\text{sizeof(element)}$ bytes of vectorization. \texttt{ArchTag} and \texttt{OperatorClass} pin the kernel to SM90 tensor-core WGMMA. \texttt{MmaTileShape} and \texttt{ClusterShape} are the thread-block tile $T_M\times T_N\times T_K$ and cluster $C_M\times C_N\times C_K$; \texttt{StageCount} sets the software-pipelined mainloop depth. \texttt{MainloopScheduleType}, \texttt{EpilogueScheduleType}, and \texttt{TileSchedulerType} select the Hopper TMA warp-specialized mainloop variant, epilogue implementation, and persistent vs.\ Stream-K tile scheduling. The two \texttt{CollectiveBuilder} aliases materialize epilogue and mainloop collectives; \texttt{GemmKernel} and \texttt{Gemm} wrap them into a universal device GEMM. The setup excerpt shows the runtime path our benchmark uses: packed CuTe strides, \texttt{GemmUniversalMode::kGemm}, alpha/beta epilogue scalars, a hardware-info struct, then \texttt{can\_implement} and \texttt{initialize} before timed \texttt{run}. Every generated source has essentially the same structure; the search space is formed by varying a small set of template/configuration choices while leaving the GEMM semantics unchanged. Table~\ref{tab:config-knobs} lists the configuration space.

\subsection{Configuration Space}

For a fixed GEMM problem shape, BF16 precision, and operand-layout tag, the generator varies tile geometry, cluster shape, pipeline depth, mainloop and epilogue schedules, and the tile scheduler. Operand element types, accumulator type, architecture tag, operator class, $C/D$ layout, alignments, and epilogue-tile sizing are not independent search axes: they are fixed by the precision tuple or left to CUTLASS defaults (\texttt{EpilogueTileAuto}). Rasterization order, swizzle patterns, and classic split-$K$ are likewise not exposed by our API; Stream-K is the only searched alternative to persistent scheduling, and CUTLASS resolves remaining lowering details internally.

\ifconf
\begin{table*}[h]
\centering
\footnotesize
\setlength{\tabcolsep}{3pt}
\renewcommand{\arraystretch}{1.05}
\caption{Configuration knobs exposed by the BF16 SM90 generator (\texttt{generate\_search\_space}).}
\label{tab:config-knobs}
\vspace{0.25em}
\begin{tabular}{@{}p{0.24\linewidth} p{0.36\linewidth} p{0.36\linewidth}@{}}
\toprule
Knob & Values / domain & Meaning \\
\midrule
\texttt{tile\_m} & $\{64,128,192,256\}$ & Thread-block $M$ tile extent \\
\texttt{tile\_n} & $\{16,32,48,64,80,96,128,192,256\}$ & Thread-block $N$ tile extent \\
\texttt{tile\_k} & $\{32,64,128,256,512\}$ & Thread-block $K$ tile / pipeline atom \\
\texttt{stages} & $\{2,\ldots,12\}$ & Mainloop pipeline stage count \\
\texttt{cluster\_m}, \texttt{cluster\_n} & $\{1,2,4,8,16\}$, $C_M C_N \le 16$ & CTA cluster shape in $M$ and $N$ \\
\texttt{cluster\_k} & fixed $1$ & $K$-clustering not searched \\
\texttt{kernel\_schedule} & WS / Pingpong / Cooperative & TMA warp-specialized mainloop policy \\
\texttt{epilogue\_schedule} & NoSmem / TMA / TMA-Cooperative & Epilogue store path \\
\texttt{scheduler} & Persistent / Stream-K & Tile scheduling across the output grid \\
\texttt{layout} & TN, TT, NN, NT & Operand layouts for $A$ and $B$ \\
\addlinespace
\multicolumn{3}{@{}l}{\textit{Fixed (not searched independently)}} \\
Operand / acc.\ types & BF16 in, FP32 acc.\ (BF16 sweep) & CUTLASS element aliases \\
\texttt{alignment\_*} & $128\,\text{bit}$ / element size & TMA vectorization requirement \\
$C/D$ layout & ColumnMajor & Row-major output via transpose trick \\
\texttt{ArchTag} / \texttt{OpClass} & Sm90 / TensorOp & Target architecture \\
\texttt{EpilogueTile} & \texttt{EpilogueTileAuto} & Sub-tile shape chosen by CUTLASS \\
\bottomrule
\end{tabular}
\vspace{-0.75em}
\end{table*}
\fi

The raw Cartesian product over the searched axes alone contains $2{,}138{,}400$ combinations.
Static validity filtering leaves $242{,}971$ BF16 configurations.

\subsection{Validity Constraints}

The Cartesian product contains many combinations that cannot instantiate as valid SM90 kernels. Enumeration is therefore followed by our set of static predicates; we also statically remove wasteful candidates with $T_N > 2M$ or $T_N > 2N$. Table~\ref{tab:validity-constraints} lists the constraints active for BF16.

\ifconf
\begin{table*}
\centering
\footnotesize
\setlength{\tabcolsep}{3.5pt}
\renewcommand{\arraystretch}{1.05}
\caption{Static validity constraints for BF16 SM90 GEMM configurations.}
\label{tab:validity-constraints}
\vspace{0.25em}
\begin{tabular}{@{}p{0.48\linewidth} p{0.48\linewidth}@{}}
\toprule
Constraint & Reason \\
\midrule
Stream-K $\Rightarrow$ cooperative mainloop & CUTLASS Stream-K scheduler requires the cooperative TMA mainloop. \\
\addlinespace
Cooperative mainloop $\Rightarrow T_M \equiv 0 \pmod{128}$ & Two consumer warpgroups split the $M$ tile; each half must stay 64-row WGMMA aligned. \\
\addlinespace
$S_{\mathrm{mainloop}} + S_{\mathrm{epi}} + 2\,\mathrm{KiB} \le 227\,\mathrm{KiB}$ & Hopper shared-memory capacity per CTA; $S_{\mathrm{epi}}$ depends on epilogue schedule and tile via \texttt{estimate\_epilogue\_smem\_bytes}. \\
\addlinespace
If $A$ is column-major: $T_K \le 256\,C_N$ & TMA smem box $K$ extent is multicast-split across the $N$ cluster dimension. \\
\addlinespace
If $B$ is row-major: $T_K \le 256\,C_M$ & Symmetric $K$-box limit along the $M$ cluster dimension for the transposed operand. \\
\bottomrule
\end{tabular}
\vspace{-0.75em}
\end{table*}
\fi

\fi
\section{From CUTLASS Candidates to Hardware-Aware Features}
\label{sec:feature-design}

\ifconf
\else
\begin{figure*}
\centering
\label{fig:overview}
\includegraphics[width=\textwidth]{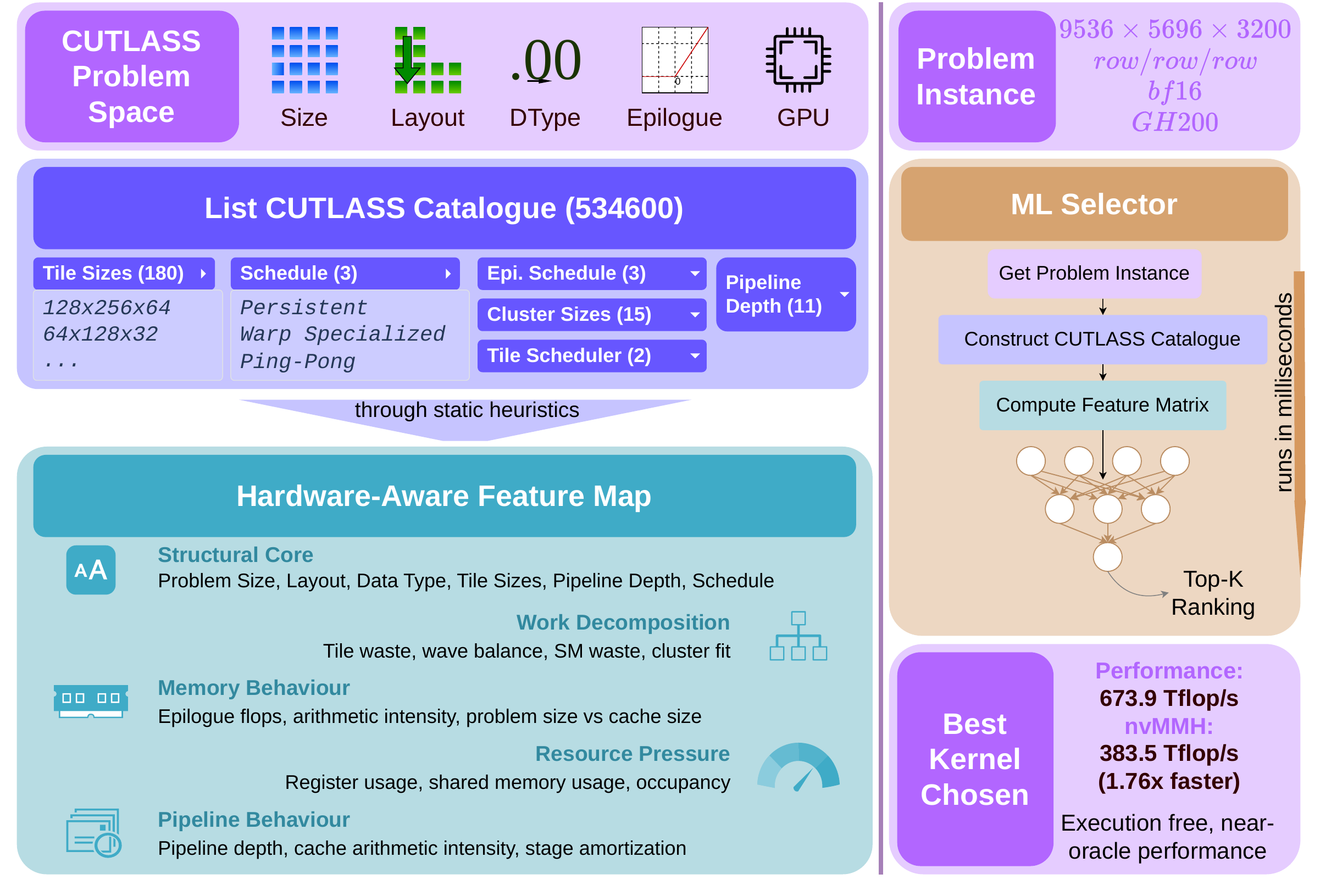}
\caption{Overview of our execution-free CUTLASS kernel selector, which maps each problem and candidate catalogue to hardware-aware features and ranks kernels without benchmarking.}
\vspace{-0.5em}
\end{figure*}
\fi

For each candidate configuration, we compute a feature map entirely from the problem description, configuration parameters, and hardware constants. This constraint is essential for an execution-free selector.
To decide which execution effects are worth representing, we first profile kernels.
%
%
\ifconf
\else

\ifconf
\section{Profiling Analysis for Feature Construction}
This section gives the detailed profiling analysis underlying the feature-design methodology in Section~\ref{sec:feature-design}.
\else
\subsection{Profiling Analysis for Feature Construction}
\fi
\label{sec:app-profiling}
The purpose of profiling is to identify execution effects that distinguish fast and slow CUTLASS configurations and can subsequently be approximated statically.

\ifconf
\subsection{Profiling setup}
\else
\subsubsection{Profiling setup}
\fi

We collected 408 Nsight Compute profiles of BF16 CUTLASS GEMM kernels on GH200 across 17 problem shapes. Profiles are associated with fixed $(M,N,K,\mathrm{layout})$ groups, within which only the CUTLASS candidate configuration changes. For every profiled candidate, we pair the collected hardware counters with the throughput measured by our benchmark harness.

Because many hardware counters scale strongly with the amount of work in the GEMM, correlations computed over all profiles can be misleading. For example, larger problems can simultaneously execute more tensor-core instructions, generate more memory traffic, and achieve higher sustained throughput. To separate these problem-size effects from candidate quality, for each metric $x$ we compute
\[
\rho_g(x)=
\operatorname{Spearman}
\!\left(
\{x(c)\}_{c\in C_g},
\{T(c)\}_{c\in C_g}
\right),
\]
where $g$ is a fixed shape--layout group, $C_g$ is the set of profiled candidates in that group, and $T(c)$ is measured benchmark throughput.

Table~\ref{tab:app-counter-correlation} reports the complete set of counters used in this analysis.

\begin{table}[!h]
\centering
\footnotesize
\setlength{\tabcolsep}{4pt}
\caption{Spearman correlation between profiled execution metrics and measured benchmark throughput.}
\ifconf
\vspace{0.5em}
\fi
\label{tab:app-counter-correlation}
\begin{tabular}{@{}lrrr@{}}
\toprule
Metric & Global $\rho$ & Mean $\rho_g$ & Median $\rho_g$ \\
\midrule
\multicolumn{4}{@{}l}{\textit{Throughput and pipeline activity}} \\
\texttt{tensor\_active\_pct}              & $+0.992$ & $+0.392$ & $+0.492$ \\
\texttt{dram\_throughput\_pct}            & $+0.817$ & $+0.408$ & $+0.500$ \\
\texttt{compute\_throughput\_pct}         & $+0.977$ & $+0.235$ & $+0.382$ \\
\texttt{tma\_active\_pct}                 & $+0.792$ & $+0.181$ & $+0.289$ \\
\texttt{sm\_active\_pct}                  & $+0.910$ & $-0.001$ & $+0.024$ \\

\addlinespace
\multicolumn{4}{@{}l}{\textit{Memory traffic and hierarchy}} \\
\texttt{dram\_read\_bytes}                & $+0.903$ & $-0.055$ & $-0.093$ \\
\texttt{dram\_write\_bytes}               & $+0.817$ & $+0.000$ & $+0.086$ \\
\texttt{l2\_read\_sectors}                & $+0.942$ & $-0.102$ & $-0.238$ \\
\texttt{l2\_write\_sectors}               & $+0.908$ & $-0.151$ & $-0.184$ \\
\texttt{l2\_throughput\_pct}              & $+0.868$ & $+0.142$ & $+0.314$ \\
\texttt{l2\_hit\_rate}                    & $-0.393$ & $-0.058$ & $-0.087$ \\
\texttt{l2\_read\_hit\_rate}              & $+0.003$ & $-0.086$ & $-0.194$ \\
\texttt{l2\_write\_hit\_rate}             & $+0.243$ & $+0.110$ & $+0.000$ \\

\addlinespace
\multicolumn{4}{@{}l}{\textit{Occupancy and resource use}} \\
\texttt{achieved\_occupancy\_pct}         & $+0.388$ & $+0.140$ & $+0.128$ \\
\texttt{registers\_per\_thread}           & $+0.736$ & $-0.151$ & $-0.315$ \\
\texttt{warps\_active}                    & $+0.388$ & $+0.112$ & $+0.145$ \\
\texttt{eligible\_warps\_per\_cycle}      & $+0.337$ & $-0.148$ & $-0.191$ \\

\addlinespace
\multicolumn{4}{@{}l}{\textit{Warp stalls}} \\
\texttt{stall\_barrier\_pct}              & $+0.894$ & $+0.156$ & $+0.143$ \\
\texttt{stall\_long\_scoreboard\_pct}     & $-0.679$ & $+0.193$ & $+0.267$ \\
\texttt{stall\_short\_scoreboard\_pct}    & $-0.863$ & $+0.029$ & $+0.029$ \\
\texttt{stall\_gmma\_pct}                 & $+0.540$ & $-0.158$ & $-0.217$ \\
\texttt{stall\_mio\_pct}                  & $-0.100$ & $+0.217$ & $+0.227$ \\
\texttt{stall\_wait\_pct}                 & $-0.519$ & $-0.113$ & $-0.143$ \\
\texttt{stall\_not\_selected\_pct}        & $+0.054$ & $-0.224$ & $-0.238$ \\

\bottomrule
\end{tabular}
\ifconf
\vspace{-1em}
\fi
\end{table}

\ifconf
\subsection{Global correlation is often a problem-size effect}
\else
\subsubsection{Global correlation is often a problem-size effect}
\fi

Several metrics appear to be excellent performance predictors when all profiles are pooled, yet lose almost all ranking power once the problem is fixed. For example, \texttt{sm\_active\_pct} has global $\rho=+0.910$ but mean within-group $\rho_g=-0.001$. Likewise, \texttt{l2\_read\_sectors} drops from $+0.942$ globally to $-0.102$ within groups, and \texttt{dram\_read\_bytes} drops from $+0.903$ to $-0.055$. These quantities primarily distinguish different amounts of work rather than good candidates from bad candidates for the same GEMM.

Two execution quantities remain substantially correlated with candidate quality after fixing the problem. \texttt{dram\_throughput\_pct} has mean within-group $\rho_g=+0.408$, and \texttt{tensor\_active\_pct} has mean $\rho_g=+0.392$; both have positive correlation in $83.7\%$ of computable groups. These measurements suggest that successful candidates simultaneously sustain the memory system and keep the tensor-core pipeline active. Since both quantities require executing the kernel, we instead construct static proxies for the mechanisms that produce them: data reuse, traffic, cache-relative working sets, feed--compute balance, pipeline fill, and pipeline amortization.

\ifconf
\subsection{Measured symptoms versus controllable causes}
\else
\subsubsection{Measured symptoms versus controllable causes}
\fi

Several counters are useful diagnostically but unsuitable as direct selector features. Barrier stalls, for example, have global $\rho=+0.894$ but only mean within-group $\rho_g=+0.156$. Long-scoreboard stalls even change apparent interpretation: their global correlation is $-0.679$, while the mean within-group correlation is $+0.193$. Such counters describe the state of an asynchronous execution pipeline after scheduling, resource allocation, and overlap have already taken effect.

This distinction is also visible when relating configuration parameters to measured counters. Pipeline depth has mean within-group correlation $-0.357$ with long-scoreboard stalls (median $-0.439$), consistent with additional stages being associated with fewer exposed memory dependencies. However, stage count itself has only weak correlation with throughput. We therefore represent the quantities that determine the consequences of staging, such as bytes per stage, total shared-memory demand, residency, pipeline fill, and stage amortization, rather than treating the stall counter or stage count as a complete performance signal.

\ifconf
\subsection{Static Configuration Effects}
\else
\subsubsection{Static Configuration Effects}
\fi
\label{sec:app-config-effects}

We additionally apply the same within-group analysis to candidate parameters and statically derived quantities. Table~\ref{tab:app-config-correlation} summarizes the strongest effects.

\begin{table}[h]
\centering
\footnotesize
\caption{Correlation between selected static candidate quantities and measured throughput.}
\ifconf
\vspace{0.5em}
\fi
\label{tab:app-config-correlation}
\begin{tabular}{@{}lrrr@{}}
\toprule
Feature & Global $\rho$ & Mean $\rho_g$ & Median $\rho_g$ \\
\midrule
\texttt{tile\_m}                  & $+0.626$ & $-0.261$ & $-0.409$ \\
\texttt{tile\_n}                  & $+0.827$ & $+0.080$ & $+0.091$ \\
\texttt{tile\_k}                  & $-0.415$ & $+0.257$ & $+0.272$ \\
\texttt{k\_iters}                 & $+0.826$ & $-0.260$ & $-0.289$ \\
\texttt{reg\_pressure\_proxy}     & $+0.827$ & $-0.222$ & $-0.319$ \\
\texttt{cluster\_size}            & $-0.337$ & $-0.247$ & $-0.223$ \\
\texttt{stages}                   & $-0.001$ & $-0.111$ & $-0.145$ \\
\texttt{bytes\_per\_stage}        & $+0.396$ & $+0.054$ & $+0.092$ \\
\texttt{pipeline\_fill\_frac}     & $+0.721$ & $+0.233$ & $+0.171$ \\
\texttt{restream\_factor}         & $+0.399$ & $+0.115$ & $+0.258$ \\
\bottomrule
\end{tabular}
\ifconf
\vspace{-1em}
\fi
\end{table}

These results illustrate why raw template parameters alone are insufficient. For example, \texttt{stages} has almost zero global correlation and only weak negative within-group correlation with throughput, even though stage depth directly affects the asynchronous mainloop. Its performance effect depends on how much storage each stage requires, how many $K$-iterations are available to amortize pipeline startup, and whether the resulting resource footprint reduces residency. The representation therefore includes both the raw stage count and derived quantities describing these interactions.

Similarly, $T_K$ has positive mean within-group correlation with throughput ($+0.257$), while the corresponding number of reduction iterations has correlation $-0.260$. However, $T_K$ is not itself an arithmetic-intensity proxy: in the tile-level ratio
\[
I_{\mathrm{tile}}
=
\frac{2T_MT_NT_K}
     {b_A T_MT_K + b_B T_NT_K}
=
\frac{2T_MT_N}
     {b_A T_M + b_B T_N},
\]
$T_K$ cancels. Instead, changing $T_K$ modifies reduction granularity, the number of mainloop iterations, per-stage storage, and the amount of computation available between pipeline synchronization points.

\ifconf
\subsection{Regime Dependence}
\else
\subsubsection{Regime Dependence}
\fi
\label{sec:app-regime-effects}

The effect of a configuration parameter need not be uniform over the shape space. To test this, we repeat the within-group analysis after partitioning the held-out problems by size, geometry, and arithmetic-intensity regime.

The strongest example is $T_K$. Its effect remains positive across coarse size regimes, with mean within-group correlations of $+0.217$, $+0.307$, and $+0.245$ for small, medium, and large problems respectively, and $+0.265$ for compute-heavy problems. The magnitude changes substantially with geometry: the mean correlation is $+0.549$ for wide GEMMs, $+0.116$ for square GEMMs, and $+0.021$ for tall GEMMs. Skinny-$K$ problems yield $+0.268$.

\begin{table}[h]
\centering
\footnotesize
\caption{Within-group correlation between $T_K$ and throughput under selected problem regimes.}
\ifconf
\vspace{0.5em}
\fi
\label{tab:app-tk-regimes}
\begin{tabular}{@{}lrr@{}}
\toprule
Regime & Mean $\rho_g$ & Median $\rho_g$ \\
\midrule
Small          & $+0.217$ & $+0.262$ \\
Medium         & $+0.307$ & $+0.412$ \\
Large          & $+0.245$ & $+0.273$ \\
Compute-heavy  & $+0.265$ & $+0.289$ \\
Memory-heavy   & $+0.238$ & $+0.262$ \\
\addlinespace
Wide           & $+0.549$ & $+0.577$ \\
Square         & $+0.116$ & $+0.237$ \\
Tall           & $+0.021$ & $-0.010$ \\
Skinny-$K$     & $+0.268$ & $+0.268$ \\
\bottomrule
\end{tabular}
\ifconf
\vspace{-1em}
\fi
\end{table}

These results show that the strength of its effect depends strongly on problem geometry and execution regime. More generally, this motivates exposing both candidate choices and the problem quantities that determine how those choices should be interpreted.

The same principle guides the complete feature map: a feature is retained when it is available before execution, distinguishes configurations or identifies the regime in which they operate, and has a plausible connection to an execution mechanism. \ifconf Appendix~\ref{sec:app-proxies} gives the resulting analytical definitions.\fi

\fi

\ifconf
\else
\subsection{Feature Map}
\fi

Table~\ref{tab:feature-groups} summarizes our chosen feature set. The structural core identifies the problem and candidate. The remaining groups estimate the hardware behavior. For each problem \(p\), candidate \(c\), and target \(h\), this procedure produces the feature vector \(x=\phi(p,c,h)\); the feature matrix of the candidate catalogue is the sole input to the selector described in Section~\ref{sec:training}.

\begin{table*}[h]
\centering
\caption{Feature groups.}
\label{tab:feature-groups}
\small
\begin{tabular}{lp{10.5 cm}}
\toprule
Group & Features and purpose \\
\midrule
Structural core
& \(\log_2 M,\log_2 N,\log_2 K\); operand layouts; input, accumulation, and output types; tile and instruction shapes; pipeline stages; mainloop and epilogue schedules; cluster shape; tile scheduler; and architecture identifiers. \\

    Work decomposition
    & Output-tile counts, reduction iterations, edge-tile waste, SM subscription, final-wave efficiency, Stream-K applicability, and cluster fit. \\

    Memory behavior
    & Problem and tile arithmetic intensity, operand re-streaming, working-set and resident-panel size relative to L2, and input, data/epilogue/conversion traffic. \\

    Resource pressure
    & Bytes per pipeline stage, total mainloop and epilogue shared-memory storage, register-pressure proxy, and shared-memory- and register-limited occupancy. \\

    Pipeline behavior
    & Estimated TMA load time, WGMMA compute time, producer--consumer ratio, pipeline-fill fraction, stage amortization, and interactions between tile depth, stage count, and occupancy. \\
    \bottomrule
\end{tabular}
\end{table*}

\ifconf
\else
\ifconf
\section{Feature Set Design}
\label{sec:app-proxies}
\fi

\ifconf
\subsection{Structural Core}
\else
\subsubsection{Structural Core}
\fi

The structural core contains the quantities required to identify a candidate: problem dimensions, operand layouts and types, threadblock and instruction tile shapes, stage count, mainloop and epilogue schedules, cluster shape, tile scheduler, and operator and compute-engine identifiers. Architecture descriptors provide the capacities and rates used by the derived features, including the number of SMs, shared-memory and register-file capacity per SM, LLC capacity, HBM bandwidth, and the peak throughput of the selected compute engine.

These features are necessary but insufficient. A stage count of four does not itself state whether four stages are enough to overlap TMA and WGMMA, whether the associated buffers fit in shared memory, or whether the problem has enough \(K\)-iterations to amortize pipeline fill. The derived groups make these consequences explicit.

\ifconf
\subsection{Work Decomposition and Wave Quantization}
\else
\subsubsection{Work Decomposition and Wave Quantization}
\fi

Let \((T_M,T_N,T_K)\) be the candidate tile shape. The output grid and number of mainloop iterations are

$$
n_M = \left\lceil \frac{M}{T_M} \right\rceil,\qquad
n_N = \left\lceil \frac{N}{T_N} \right\rceil,\qquad
n_K = \left\lceil \frac{K}{T_K} \right\rceil.
$$

The candidate computes \(n_M n_N\) output tiles. The edge tiles may execute masked work that is discarded; we encode this separately along each output dimension:

$$
w_M = \frac{n_M T_M - M}{M},
\qquad
w_N = \frac{n_N T_N - N}{N}.
$$

To estimate wave quantization, we divide the number of output tiles by the number of concurrently resident blocks across the GPU. If \(S_{\mathrm{SM}}\) is the SM count and \(b_{\mathrm{res}}\) the estimated resident blocks per SM, then

$$
q_{\mathrm{SM}}
=
\left\lceil\frac{n_M n_N}{S_{\mathrm{SM}} b_{\mathrm{res}}}\right\rceil
$$

is the number of passes that the resident blocks make over the tile list. The corresponding wave efficiency is

$$
\eta_{\mathrm{last}}
=
\frac{n_M n_N}
{q_{\mathrm{SM}}
 S_{\mathrm{SM}} b_{\mathrm{res}}}.
$$

A value near one indicates that there are several waves and/or the final wave fills the GPU, while a small value indicates a ragged tail. This directly motivates the Stream-K proxy:

$$
\texttt{streamK\_applicability}
=
\max(0, 1-\eta_{\mathrm{last}}),
$$

together with a feature indicating whether the candidate actually uses Stream-K. Cluster features similarly encode the cluster dimensions, number of tiles available to fill each cluster dimension, and whether the cluster overshoots the grid.

\ifconf
\subsection{Memory Behavior}
\else
\subsubsection{Memory Behavior}
\fi

We estimate problem arithmetic intensity using the bytes required by the input matrices and the complete epilogue:

$$
I_{\mathrm{problem}}
=
\frac{2MNK}
{b_A MK + b_B KN + Q_{\mathrm{epilogue}}},
$$

where \(b_A\) and \(b_B\) are input element sizes and \(Q_{\mathrm{epilogue}}\) includes output stores and any reads or writes induced by the output type, \(\beta C\), residual, bias, activation, or conversion operations.

A complementary tile-level feature captures the amount of tensor-core work obtained from a loaded operand footprint:

$$
I_{\mathrm{mainloop}}
=
\frac{2T_M T_N}
{T_M b_A + T_N b_B}.
$$

Here \(T_K\) cancels. This is useful: \(T_K\) affects per-stage storage and pipeline depth, whereas \(T_M\) and \(T_N\) determine how much reuse the tile geometry obtains from each loaded operand row or column.

We estimate the re-streaming incurred if operand panels cannot remain resident in LLC:

$$
r_{\mathrm{stream}}
=
\frac{
n_N b_A MK + n_M b_B KN
}{
b_A MK + b_B KN
}.
$$

The remaining cache features compare the complete operand working set and a single resident operand panel with LLC capacity. Together, these features distinguish a problem whose repeated operand reads are absorbed by LLC from one that must repeatedly fetch them from HBM.

\ifconf
\subsection{Resource Pressure and Occupancy}
\else
\subsubsection{Resource Pressure and Occupancy}
\fi

The shared-memory buffer required for one mainloop stage is

$$
B_{\mathrm{stage}}
=
T_M T_K b_A + T_N T_K b_B.
$$

For \(S\) pipeline stages, the mainloop requires \(S B_{\mathrm{stage}}\) bytes. The epilogue additionally requires schedule-dependent storage \(B_{\mathrm{epi}}\), including output fragments and any fused epilogue state. For a warp-specialized schedule, mainloop and epilogue storage may overlap; cooperative and ping-pong schedules may require both simultaneously. We therefore compute the corresponding schedule-specific total shared-memory requirement \(B_{\mathrm{smem}}\) and its fraction of per-SM capacity:

$$
f_{\mathrm{smem}}
=
\frac{B_{\mathrm{smem}}}{C_{\mathrm{smem}}}.
$$

This bounds shared-memory-limited residency:

$$
b_{\mathrm{smem}}
=
\left\lfloor
\frac{C_{\mathrm{smem}}}{B_{\mathrm{smem}}}
\right\rfloor.
$$

We approximate register pressure from the output elements accumulated per thread,

$$
r_{\mathrm{proxy}}
=
\frac{T_M T_N}{T_{\mathrm{block}}},
$$

and use the register-file capacity \(C_{\mathrm{reg}}\) to derive an independent residency bound \(b_{\mathrm{reg}}\). The estimated resident-block count is

$$
b_{\mathrm{res}}
=
\min(b_{\mathrm{smem}}, b_{\mathrm{reg}}, b_{\mathrm{arch}}).
$$

Keeping the separate bounds matters, because increasing the stage count tightens the shared-memory bound but does not directly alter accumulator pressure. The model can therefore distinguish a kernel limited by registers from one that falls off a shared-memory size limitation.

\ifconf
\subsection{Pipeline Balance}
\else
\subsubsection{Pipeline Balance}
\fi

Hopper warp-specialized kernels overlap TMA loads from HBM to shared memory with WGMMA execution from shared memory. We estimate the producer and consumer times for one stage as

$$
t_{\mathrm{producer}}
=
\frac{n_{SM}B_{\mathrm{stage}}}{\widehat{BW}_{\mathrm{TMA}}}
\qquad
t_{\mathrm{consumer}}
=
\frac{2T_M T_N T_K}
{n_{\mathrm{consumer}}\,\widehat{P}_{\mathrm{WGMMA}}}
$$

where \(n_{\mathrm{consumer}}\) is the number of consumer warpgroups actively executing the mainloop for the selected schedule, $\widehat{BW}_{\mathrm{TMA}}$ is the bandwidth of the TMA operation, and $\widehat{P}_{\mathrm{WGMMA}}$ is the throughput of the WGMMA operation. The producer-consumer ratio

$$
r_{\mathrm{pc}}
=
\frac{t_{\mathrm{consumer}}}{t_{\mathrm{producer}}}
$$

is a static proxy for whether TMA can supply WGMMA at the required rate. Values far below one indicate that consumers are likely to wait for data; values far above one indicate that loads are completed faster than arithmetic consumes them.

Two additional features describe whether the pipeline depth is useful for the problem:

$$
f_{\mathrm{fill}}
=
\frac{\min(S,n_K)}{S},
\qquad
a_{\mathrm{stage}}
=
\frac{n_K}{S}.
$$

The first falls below one when the contraction is too shallow to prime a deep pipeline. The second measures how often the pipeline fill cost is amortized. Finally,
$
S \cdot f_{\mathrm{smem}}
$
exposes the interaction between stage count and shared-memory pressure. This interaction captures both the stage-count cliff and the regime in which a larger \(T_K\) becomes harmful because its enlarged stage buffers force a shallower pipeline.

\fi
\section{Training the Selector}
\label{sec:training}

Given an input configuration, we list the valid CUTLASS kernel catalogue, and use a machine learning model to rank the catalogue and choose the best kernel. In this section, we explain the data collection process for training and the model design. 

\subsection{Constructing The Training Dataset}
\label{sec:dataset}

The CUTLASS configuration space is too large to measure exhaustively for every problem shape. Even after validity filtering, a single BF16 GEMM layout exposes over sixty thousand candidate configurations. Exhaustively sweeping this space over thousands of shapes would require thousands of compiles and hundreds of millions of benchmarks.

We therefore separate the collection into two tasks. Our training dataset covers a broad range of GEMM shapes while also sampling candidates within each shape-layout group.
The held-out evaluation dataset covers fewer shapes, but sweeps their valid candidates exhaustively to establish a trustworthy in-space oracle.
This section describes the training GEMM collection; the same collection design is used when extending the dataset to additional types, epilogues, and operator families.

\subsubsection{Shape Selection}







A GEMM problem shape is a triple $(M,N,K)$.
We define our structured anchor grid in 3 dimensions as [$32$,$16384$]$^3$ with each dimension snapped to a multiple of $32$ -- roughly $134$ million lattice points.
The preferred CUTLASS kernel changes with problem geometry (square, tall, wide, skinny-$K$), reduction depth ($\lceil K/T_K\rceil$), and cache residency; a single random draw over this box would leave boundary (extreme large/small values of one/more dimensions) and hardware transition points severely underrepresented.
We therefore use two complementary (but non-disjoint) construction axes: a \emph{geometry class} (interior versus boundary/extreme) and, where needed, targeted \emph{supplements} (anchor neighbours and transition points). The final dataset contains 593 base shapes
($\times$ 4 layouts $=$ 2372 unique groups):

\begin{itemize}
\item \textbf{Primary random coverage (469 shapes).}
297 \emph{interior} shapes: log-uniform draws from the anchor grid.
172 \emph{boundary} shapes: deliberate hull extremes; very tall, wide, skinny-$K$, and corner cases up to 16384 on an axis.

\item \textbf{Targeted supplements (124 shapes).}
92 \emph{anchor} shapes: Neighbours formed by ablating shapes. For every data point, the training set includes shapes that vary one dimension at a time where the domain permits it. This helps the model learn how changing different dimensions of the input shape changes performance and behaviour. This adds 47 interior and 45 boundary shapes.
32 \emph{transition} shapes: points near the roofline ridge (BF16 arithmetic intensity $\approx 247.3$\,flop/byte on GH200) and the 50\,MiB LLC-residency boundary. This adds 14 interior and 18 boundary shapes.
\end{itemize}

\ifconf
\else
\subsubsection{Candidate Sampling}
\fi

For a shape-layout group, we measure only a budgeted subset of the full catalogue during training. The default budget is 2000 candidates per group. It is reduced to 1000 for the largest shapes, where each benchmark is substantially more expensive, and increased to 3000 for atypical or transition shapes, where the configuration ranking changes more sharply. \ifconf
\else
\ifconf
\section{Sampling Kernels - Analysis}
\fi
\label{sec:app-sampling}

Exhaustive measurement is impractical for training. Therefore we study how the measurement budget should be allocated across candidates. The goal of this experiment is to quantify how effectively a sampling policy exposes high-performing kernels within a fixed measurement budget.

We use the exhaustively measured BF16 evaluation groups as ground truth. For each group \(g\), let \(C_g\) denote the complete valid candidate catalogue, \(T(c)\) the measured throughput of candidate \(c\), and
\[
T_g^\star = \max_{c \in C_g} T(c)
\]
the exhaustive oracle throughput. Given a sampling policy that selects a subset \(S_B(g) \subseteq C_g\) of \(B\) candidates, we define the \emph{sampling regret}
\[
R_B(g)
=
1 -
\frac{
\max_{c \in S_B(g)} T(c)
}{
T_g^\star
}.
\]
This measures the quality lost purely because the measurement subset failed to contain the best available kernels. For stochastic policies, we repeat each sampling experiment $100$ times per group and report mean results.

\ifconf
\subsection{Sparsity of the High-Performance Tail}
\else
\subsubsection{Sparsity of the High-Performance Tail}
\fi

We first quantify how many kernels in the valid catalogue are actually competitive. For each group, we compute the fraction of candidates whose throughput lies within \(\epsilon\) of the exhaustive oracle:
\[
q_\epsilon(g)
=
\frac{
\left|
\left\{
c \in C_g :
T(c) \ge (1-\epsilon)T_g^\star
\right\}
\right|
}{
|C_g|
}.
\]

Across the evaluation groups, only 0.43\% of candidates are within 1\% of the oracle on average, 2.4\% are within 5\%, and 10.9\% are within 10\%. The corresponding median fractions are 0.17\%, 1.0\%, and 4.4\%, respectively. In the most selective groups, fewer than 0.33\% of valid kernels lie within 5\% of the optimum.

These results show that high-performing kernels occupy a sparse tail of the catalogue. A uniformly sampled training subset can therefore contain thousands of measured candidates while still failing to observe any configuration close to the true optimum.

\begin{table}
\centering
\footnotesize
\caption{Fraction of valid kernels close to the exhaustive oracle.}
\ifconf
\vspace{0.5em}
\fi
\label{tab:sampling-tail}
\begin{tabular}{@{}lccc@{}}
\toprule
Threshold & Mean fraction & Median fraction & Minimum fraction \\
\midrule
Within 1\%  & 0.43\% & 0.17\% & 0.17\% \\
Within 5\%  & 2.41\% & 1.00\% & 0.33\% \\
Within 10\% & 10.91\% & 4.42\% & 1.17\% \\
\bottomrule
\end{tabular}
\ifconf
\vspace{-1em}
\fi
\end{table}

\ifconf
\subsection{Sampling Policies}
\else
\subsubsection{Sampling Policies}
\fi

We compare the following candidate-selection policies.

\paragraph{Uniform.}
Candidates are sampled uniformly at random from the complete valid catalogue.

\paragraph{Static top-\(B\).}
Candidates are ranked by the final static sampling score from\ifconf~Section~\ref{sec:dataset}\else~Section~\ref{sec:sampling-summary}\fi, and the highest-ranked \(B\) candidates are selected deterministically.

\paragraph{Biased-diverse sampling.}
This is the policy used to construct the training set. A fraction 75\% of the budget is sampled from the high-ranked portion of the static score, while the remaining 25\% is sampled uniformly from the rest of the catalogue. The exploratory quarter is not overhead for oracle discovery alone: the throughput predictor must also observe poorly ranked configurations during training, so a purely heuristic-biased corpus would under-represent the long tail of bad kernels.

We evaluate budgets
\[
B \in \{100, 250, 500, 1000, 2000, 3000, 5000, 10000\}.
\]
The values \(1000\), \(2000\), and \(3000\) correspond directly to the candidate budgets used in our training collection.

\ifconf
\subsection{Sampling Efficiency}
\else
\subsubsection{Sampling Efficiency}
\fi

Figure~\ref{fig:sampling-regret} reports sampling regret as a function of measurement budget. Uniform sampling improves steadily with larger budgets, but converges slowly because the near-optimal tail is sparse. At a budget of 1000 candidates, uniform sampling leaves 11.3\% mean regret; at 2000 candidates the regret is 8.5\%, and at 3000 candidates it remains 7.1\%.

The static top-\(B\) policy improves over uniform sampling at small budgets, reaching 0.6\% regret at \(B=2000\). However, its deterministic bias can exclude competitive kernels that receive a poor heuristic score. The biased-diverse policy combines both effects and achieves 1.5\%, 0.9\%, and 0.5\% mean regret at budgets of 1000, 2000, and 3000, respectively.

At the default training budget of 2000 candidates, biased-diverse sampling reduces sampling regret by 89.4\% relative to uniform sampling (from 8.5\% to 0.9\%). Static top-\(B\) attains marginally lower regret (0.6\%) at this budget, but it is unsuitable as a training policy because it never deliberately measures configurations the heuristic ranks poorly. Oracle-discovery metrics and training-corpus construction therefore pull in opposite directions under a fixed budget.

\begin{figure}[t]
\centering
\caption{Sampling regret versus candidate measurement budget on the exhaustive BF16 evaluation groups.}
\ifconf
\includegraphics[width=0.6\linewidth]{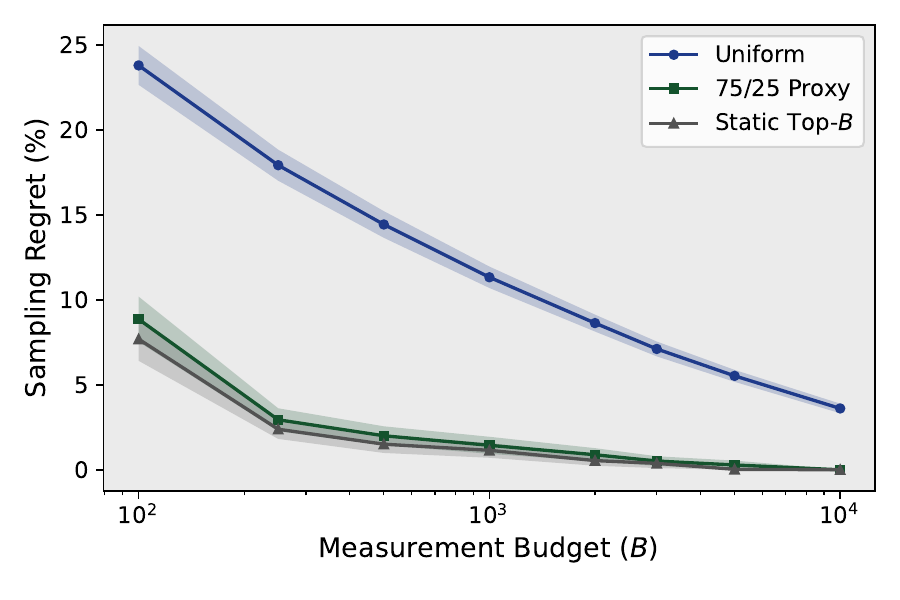}
\else
\includegraphics[width=\linewidth]{figures/sampling_regret_vs_budget.pdf}
\fi
\label{fig:sampling-regret}
\end{figure}

\ifconf
\subsection{Probability of Observing a Near-Optimal Kernel}
\else
\subsubsection{Probability of Observing a Near-Optimal Kernel}
\fi

Sampling regret summarizes the quality of the best observed configuration, but it is also useful to ask how often a sampling policy discovers at least one near-optimal kernel. We therefore measure
\[
P_\epsilon(B)
=
\Pr
\left[
\max_{c\in S_B(g)} T(c)
\ge
(1-\epsilon)T_g^\star
\right]
\]
for \(\epsilon \in \{0.01,0.05,0.10\}\).

Figure~\ref{fig:sampling-success} reports \(P_{0.05}(B)\) as a function of measurement budget. At \(B=2000\), uniform sampling observes at least one kernel within 5\% of the oracle in 25.6\% of groups, compared with 97.1\% for static top-\(B\) and 91.3\% for biased-diverse sampling. At \(B=3000\), the corresponding fractions are 35.3\%, 97.1\%, and 97.3\%. Uniform sampling improves with budget but remains far below the heuristic-biased policies until \(B \ge 3000\), where biased-diverse sampling essentially matches static top-\(B\).

\begin{figure}
\centering
\caption{Probability that the sampled candidate pool contains a near-optimal kernel.}
\ifconf
\includegraphics[width=0.6\linewidth]{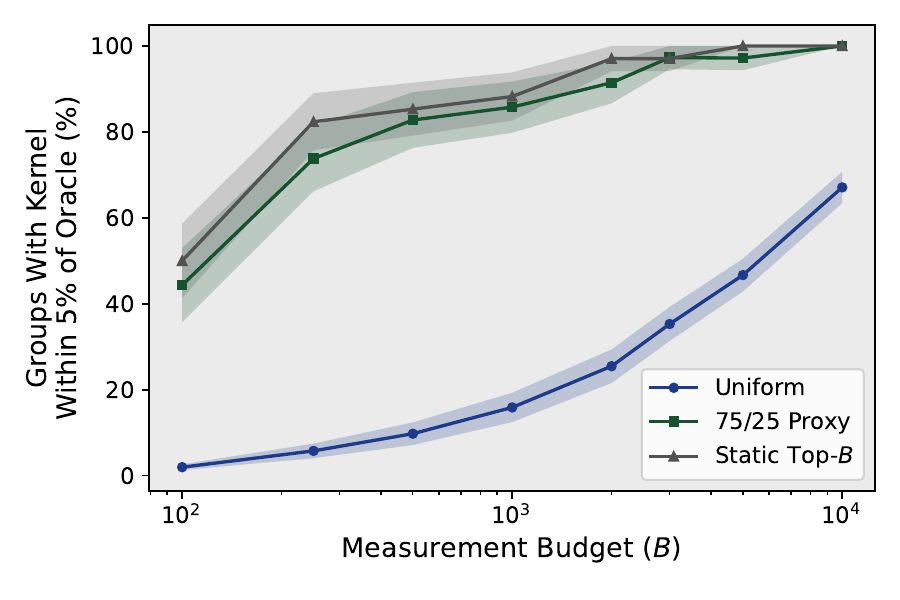}
\else
\includegraphics[width=\linewidth]{figures/sampling_within5_vs_budget.pdf}
\fi
\label{fig:sampling-success}
\end{figure}

\ifconf
\subsection{Effect of the Exploration Fraction}
\else
\subsubsection{Effect of the Exploration Fraction}
\fi

\begin{table}[h]
\centering
\footnotesize
\caption{Ablation of the exploitation fraction at \(B=2000\).}
\label{tab:sampling-mixture}
\begin{tabular}{@{}ccc@{}}
\toprule
Heuristic-biased fraction \(\alpha\) & Mean sampling regret & Groups within 5\% of oracle \\
\midrule
0.00 & 8.5\% & 25.6\% \\
0.25 & 1.4\% & 87.0\% \\
0.50 & 1.1\% & 88.9\% \\
0.75 & 0.9\% & 91.3\% \\
1.00 & 0.6\% & 97.1\% \\
\bottomrule
\end{tabular}
\end{table}

The biased-diverse policy trades off exploitation of the static score against broader exploration of the remaining catalogue. We ablate this mixture at a fixed budget of \(B=2000\). Let \(\alpha\) denote the fraction of the budget drawn from the heuristic-preferred region. We test
\[
\alpha \in \{0,0.25,0.50,0.75,1.0\}.
\]

Pure exploration (\(\alpha=0\)) reduces to broad random sampling and yields 8.5\% mean regret. Pure exploitation (\(\alpha=1\)) achieves 0.6\% regret and 97.1\% within-5\% coverage, the best values \emph{for oracle discovery} on this evaluation set. Increasing \(\alpha\) monotonically improves both metrics, yet \(\alpha=1\) is not a viable training policy: the model would rarely see measured examples of low-throughput kernels and could not learn to discriminate bad configurations from good ones.

This retrospective analysis validates the \(\alpha=0.75\) policy used for corpus construction. At \(B=2000\) this yields 0.9\% mean regret and 91.3\% within-5\% coverage---a modest cost relative to \(\alpha=1\)---while guaranteeing that one quarter of every measurement budget samples outside the heuristic-preferred region.

\ifconf
\subsection{Summary}
\else
\subsubsection{Summary}
\label{sec:sampling-summary}
\fi

The exhaustive study confirms that the candidate-selection problem is dominated by a sparse high-performance tail: only 2.4\% of valid kernels lie within 5\% of the oracle on average. Consequently, uniform sampling requires substantially larger measurement budgets to expose competitive configurations. At the default budget of 2000 candidates per shape--layout group, our biased-diverse policy reduces sampling regret from 8.5\% to 0.9\% and increases the probability of observing a kernel within 5\% of the oracle from 25.6\% to 91.3\%.

These results retroactively validate the candidate-sampling strategy used for the training corpus. Most measurements are directed toward configurations with favorable static efficiency estimates, while the fixed 25\% exploratory fraction ensures the predictor is trained on bad as well as good kernels---a requirement that pure top-\(B\) or \(\alpha=1\) sampling cannot satisfy under the same budget.

\fi
Instead, we rank the valid candidates using a cheap, static efficiency score that ranks kernels likely to give better performance. Let \(c_M,c_N\) be the thead-block cluster shape, \(S_{\mathrm{SM}}\) the number of SM clusters, $S$ the pipeline stage count, and
$$
n_c =
\left\lceil \frac{M}{c_MT_M} \right\rceil
\left\lceil \frac{N}{c_NT_N} \right\rceil
$$
be the number of clusters required by the launch. The number of cluster waves is thus 
$
W = \lceil \frac{n_c}{S_{\mathrm{SM}}} \rceil.
$
We use the following sampling score to rank kernels:
$$
s_{\mathrm{sample}}
=
\frac{n_c}
{W S_{\mathrm{SM}} c_M c_N}
\cdot
S
\cdot
\sqrt{T_M T_N T_K}
$$
The first term favours efficient wave filling, penalizing oversized clusters that are too wasteful. The remaining terms favour configurations with deeper pipelines and sufficiently large tiles, which are likely to have higher reuse and provide better throughput. For a group budget, three quarters of the sampled configurations are drawn from the top of this ranking, while the remaining quarter is sampled across the rest of the ranked space. This preserves examples of poor and mediocre configurations. If the number of valid candidates is below the assigned budget, we collect the entire valid pool. The final training collection contains approximately 4.9 million measured kernels.

\subsubsection{Collection Pipeline}

We perform data collection on
\ifconf
\else
a GH200 node of the Daint supercomputer on CSCS Alps,
\fi
equipped with 4 GH200 GPUs. We compile CUTLASS SM90a kernels from the commit \texttt{3476ddb7} using CUDA 13.1.
Each measurement records the behavior of one candidate on one shape-layout group. Each successful benchmark consists of five warm-up iterations followed by five rounds of three timed iterations. Every attempted measurement is recorded, including successful executions, compilation failures, launch failures, crashes, and timeouts.
\ifconf
\else
To make the dataset practical to collect, we implement the following optimizations:

\begin{enumerate}
    \item \textbf{Batched compilation.} Kernels that share a subset of configurations are compiled together into one shared library. A single exported entry point dispatches by kernel index inside the batch, amortizing template instantiation and \texttt{nvcc} startup cost across many configurations.

    \item \textbf{Executable caching.} Each batch library is stored under a name derived from a hash of its member configurations, and kernels are reused instead of recompilation.

    \item \textbf{Buffer caching.} Each GPU worker pre-allocates a fixed device memory pool and serves all kernels from slices of that pool rather than allocating fresh tensors per kernel. We replicate the matrices enough times from this pool until the working set exceeds L2 capacity.

    \item \textbf{Node-local database storage.} Benchmark results and compile metadata are written to a registry database kept in node-local memory for the duration of the job, and checkpointed periodically to scratch and to durable home storage.

    \item \textbf{Parallel compilation and benchmarking.} Compilation and GPU benchmarking run concurrently: a pool of CPU workers builds shared libraries while one persistent process per GPU consumes a queue of newly compiled kernels. A monitor recovers from worker crashes and hung kernels by attributing the failure to the active measurement, recording it, and respawning the worker to continue with the remainder of the queue.
\end{enumerate}
\fi

\ifconf
\else
\ifconf
\section{Evaluation Dataset}
\else
\subsection{Evaluation Dataset}
\fi
\label{sec:app-evalset}

Unlike the training corpus, where each problem is measured on a proxy shortlist of $\approx 2{,}000$ configurations, the evaluation corpus is \emph{exhaustively} enumerated: for every held-out shape $(M,N,K)$ in Table~\ref{tab:eval-shapes} and every layout tag $\ell\in\{\mathrm{TN},\mathrm{TT},\mathrm{NN},\mathrm{NT}\}$, the candidate set $\mathcal{C}(M,N,K,\ell)$ is the full catalogue after static validity filtering (Section~\ref{sec:app-space}) and shape-aware plan guards. 
No further proxy pruning is applied. 
The $17$ shapes are listed exhaustively; together with four layouts they define $68$ evaluation groups and $3{,}922{,}067$ distinct (kernel, shape) benchmark tasks.

\begin{table}[t]
\centering
\footnotesize
\setlength{\tabcolsep}{4pt}
\caption{Held-out evaluation shapes and valid kernel counts per layout.}
\label{tab:eval-shapes}
\begin{tabular}{@{}rrr rrrrr@{}}
\toprule
\multicolumn{3}{c}{Shape} & \multicolumn{5}{c}{Kernel count} \\
\cmidrule(lr){1-3}\cmidrule(lr){4-8}
$M$ & $N$ & $K$ & TN & TT & NN & NT & Total \\
\midrule
2048 & 2048 & 2048 & 60{,}795 & 60{,}740 & 60{,}740 & 60{,}696 & 242{,}971 \\
4096 & 4096 & 4096 & 60{,}795 & 60{,}740 & 60{,}740 & 60{,}696 & 242{,}971 \\
64 & 64 & 64 & 32{,}190 & 32{,}135 & 32{,}135 & 32{,}091 & 128{,}551 \\
128 & 128 & 128 & 60{,}795 & 60{,}740 & 60{,}740 & 60{,}696 & 242{,}971 \\
256 & 256 & 256 & 60{,}795 & 60{,}740 & 60{,}740 & 60{,}696 & 242{,}971 \\
512 & 512 & 512 & 60{,}795 & 60{,}740 & 60{,}740 & 60{,}696 & 242{,}971 \\
32 & 128 & 4096 & 60{,}795 & 60{,}740 & 60{,}740 & 60{,}696 & 242{,}971 \\
2048 & 128 & 4096 & 60{,}795 & 60{,}740 & 60{,}740 & 60{,}696 & 242{,}971 \\
4096 & 128 & 4096 & 60{,}795 & 60{,}740 & 60{,}740 & 60{,}696 & 242{,}971 \\
12{,}288 & 128 & 4096 & 60{,}795 & 60{,}740 & 60{,}740 & 60{,}696 & 242{,}971 \\
256 & 4096 & 4096 & 60{,}795 & 60{,}740 & 60{,}740 & 60{,}696 & 242{,}971 \\
256 & 12{,}288 & 4096 & 60{,}795 & 60{,}740 & 60{,}740 & 60{,}696 & 242{,}971 \\
64 & 12{,}288 & 4096 & 37{,}290 & 37{,}235 & 37{,}235 & 37{,}191 & 148{,}951 \\
2048 & 2048 & 128 & 60{,}795 & 60{,}740 & 60{,}740 & 60{,}696 & 242{,}971 \\
4096 & 11{,}008 & 4096 & 60{,}795 & 60{,}740 & 60{,}740 & 60{,}696 & 242{,}971 \\
256 & 256 & 8192 & 60{,}795 & 60{,}740 & 60{,}740 & 60{,}696 & 242{,}971 \\
512 & 3072 & 768 & 60{,}795 & 60{,}740 & 60{,}740 & 60{,}696 & 242{,}971 \\
\midrule
\multicolumn{3}{l}{Sum (68 groups)} & 981{,}405 & 980{,}470 & 980{,}470 & 979{,}722 & 3{,}922{,}067 \\
\bottomrule
\end{tabular}
\end{table}

\fi

\subsection{Learning to Rank Candidates}
\label{sec:model}

We study two model families for mapping the hardware-aware representation
$\phi(p,c,h)$ to a scalar candidate score.  In both cases, candidates are
grouped by GEMM problem $p$ and supervision is derived from
measured kernel throughput within that group. 
\ifconf
\else
\ifconf
\section{Model Design}
\fi
\label{sec:app-model}

\ifconf
\subsection{Problem and Candidate Sets}
\else
\subsubsection{Problem and Candidate Sets}
\fi

A GEMM instance is a tuple $p=(M,N,K,\ell)$, where $\ell$ denotes the operand layout tag. $h$ be a hardware descriptor.
For each $p$ we maintain a \emph{finite candidate set} $\mathcal{C}(p)\subset\mathcal{K}$, where $\mathcal{K}$ is the statically valid CUTLASS configuration catalogue (Section~\ref{sec:app-space}).
Every $c\in\mathcal{C}(p)$ is described by a feature vector $\mathbf{x}(p,c)=\phi(p,c)\in\mathbb{R}^{d}$ computed from problem geometry, tile and cluster parameters, schedule choices, and the analytical hardware proxies of Section~\ref{sec:feature-design}; no feature depends on measured runtime counters.
A learned selector is a scoring function $f_\theta:\mathbb{R}^{d}\to\mathbb{R}$.
The deployed decision rule is always
\begin{equation}
  \hat{c}(p)\;=\;\arg\max_{c\in\mathcal{C}(p)} f_\theta\!\bigl(\mathbf{x}(p,c, h)\bigr),
  \label{eq:selection}
\end{equation}
i.e.\ we reduce selection to within-problem ranking and do not use a predicted throughput value directly at inference time.

\ifconf
\subsection{Measured Labels}
\else
\subsubsection{Measured Labels}
\fi

For each successfully benchmarked pair $(p,c)$ the autotuner records a mean throughput estimate $\widehat{T}(p,c)$ and an empirical standard deviation $\widehat{\sigma}(p,c)$ over repeated timed rounds.
Group all rows that share the same $p$ and index such a group by $g$.
The \emph{empirical group maximum} is
\begin{equation}
  T^{\star}_g \;=\; \max_{c\in\mathcal{C}(g)} \widehat{T}(g,c),
  \qquad
  y(g,c) \;=\; \frac{\widehat{T}(g,c)}{T^{\star}_g} \in (0,1],
  \label{eq:ynorm}
\end{equation}
so that $y(g,c)=1$ for at least one measured candidate in the group and $y(g,c)<1$ otherwise.
On training problems $\mathcal{C}(g)$ is a proxy shortlist (Section~\ref{sec:dataset}); on evaluation problems it is a near-exhaustive sweep over the catalogue, so $T^{\star}_g$ approximates the in-space empirical oracle for that shape--layout.

\ifconf
\subsection{Relevance grades (ranking supervision)}
\else
\subsubsection{Relevance grades (ranking supervision)}
\fi

Ranking objectives do not use raw $\widehat{T}$ directly; they use discrete grades derived from it.
Within group $g$, sort candidates by decreasing $\widehat{T}(g,c)$ and partition the sorted list into contiguous \emph{bands} by merging adjacent entries $i$ and $i{+}1$ whenever their throughput gap is smaller than twice the pooled measurement uncertainty,
\begin{equation}
  \widehat{T}_i - \widehat{T}_{i+1} < 2\sqrt{\widehat{\sigma}_i^{\,2}+\widehat{\sigma}_{i+1}^{\,2}}.
  \label{eq:tie-merge}
\end{equation}
Let $b(c)\in\{0,1,\ldots\}$ be the band index of $c$ ($0$ for the top band).
We assign integer grades $\gamma(g,c)=G_{\max}-b(c)$ with $G_{\max}=31$.
For listwise tree training we further compress grades to a top-focused relevance scale with $B$ bands ($B{=}2$ by default):
\begin{equation}
  \mathrm{rel}(g,c) \;=\; \mathrm{clip}\bigl(\gamma(g,c)-(G_{\max}-B),\,0,\,B\bigr).
  \label{eq:rel-compress}
\end{equation}
Thus only the highest-throughput band(s) carry non-zero relevance, concentrating learning on near-optimal orderings.

\ifconf
\subsection{Model A (gradient-boosted ranker)}
\else
\subsubsection{Model A (gradient-boosted ranker)}
\fi

Model~A is an additive ensemble of regression trees $f_\theta=\sum_m h_m$ fitted with XGBoost.
Numeric features enter splits directly; categorical schedule and layout fields use native categorical handling with a fixed level ordering shared across train, validation, and deployment.
Let $\mathcal{G}_{\mathrm{tr}}$ denote the set of training groups and list all rows in a group-contiguous order with group-id vector $\mathbf{g}$.
We train with one of three objectives:
\begin{itemize}
\item \textbf{Listwise ranking} (default): minimise a surrogate for NDCG at cut-off $|\mathcal{C}(g)|$ using targets $\mathrm{rel}(g,c)$ and the standard $\mathrm{rank\!:\!ndcg}$ objective;
\item \textbf{Pairwise ranking}: $\mathrm{rank\!:\!pairwise}$ on the same relevance labels;
\item \textbf{Pointwise regression} (ablation): minimise $\sum_{i} \bigl(f_\theta(\mathbf{x}_i)-y_i\bigr)^2$ with $\mathrm{reg\!:\!squarederror}$, treating $y_i$ as a continuous target independent of group structure.
\end{itemize}
Only the first two objectives align the training loss with the argmax decision rule in~\eqref{eq:selection}; the regression ablation tests whether accurate throughput prediction implies good selection.

\ifconf
\subsection{Model B (feed-forward ranker)}
\else
\subsubsection{Model B (feed-forward ranker)}
\fi

Model~B is a multilayer perceptron $f_\theta:\mathbb{R}^{d'}\to\mathbb{R}$ applied to a transformed feature matrix.
Let $\mathbf{x}^{\mathrm{num}}$ denote numeric columns and $\mathbf{x}^{\mathrm{cat}}$ categorical indicators obtained by one-hot encoding with the same frozen level sets as in Model~A.
Training applies per-column standardisation to $\mathbf{x}^{\mathrm{num}}$ (zero mean, unit variance, with $\mathrm{NaN}/\pm\infty$ mapped to zero before scaling); categoricals are left unscaled.
We consider three training losses:
\begin{align}
  \mathcal{L}_{\mathrm{MSE}}(\theta)
  &= \frac{1}{N}\sum_{i=1}^{N}\bigl(f_\theta(\mathbf{x}'_i)-y_i\bigr)^2,
  \label{eq:mse-loss}
\\
  \mathcal{L}_{\mathrm{RankNet}}(\theta)
  &= \frac{1}{|\mathcal{P}|}\sum_{(i,j)\in\mathcal{P}}
     \log\bigl(1+\exp\bigl(f_\theta(\mathbf{x}'_j)-f_\theta(\mathbf{x}'_i)\bigr)\bigr),
  \label{eq:ranknet}
\\
  \mathcal{L}_{\mathrm{LambdaRank}}(\theta)
  &= \frac{1}{|\mathcal{P}|}\sum_{(i,j)\in\mathcal{P}}
     \Delta\mathrm{NDCG}_{ij}\,
     \log\bigl(1+\exp\bigl(f_\theta(\mathbf{x}'_j)-f_\theta(\mathbf{x}'_i)\bigr)\bigr),
  \label{eq:lambdarank}
\end{align}
where $\mathbf{x}'_i$ is the scaled/encoded feature vector, and $\mathcal{P}$ is a set of ordered pairs within a group such that $y_i>y_j$ (higher-throughput index $i$, lower $j$).
For LambdaRank we use gain values $g_i=y_i$ (linear mode) or $g_i=2^{y_i}-1$ (exponential mode), ideal discounted cumulative gain $\mathrm{IDCG}=\sum_{r=1}^{n} g_{\pi(r)}/\log_2(1{+}r)$ for the throughput-sorted permutation $\pi$, and pair weights
\begin{equation}
  \Delta\mathrm{NDCG}_{ij} \;=\;
  \frac{\bigl|g_i-g_j\bigr|\,\bigl|D(r_i)-D(r_j)\bigr|}{\mathrm{IDCG}},
  \qquad
  D(r)=\frac{1}{\log_2(1+r)},
  \label{eq:delta-ndcg}
\end{equation}
with ranks $r_i,r_j$ computed from the current model scores (detached from the gradient graph when evaluating $D$).
Pairs are sampled per group with a hard cap; pairs touching any of the top eight $y$-values are always retained (\emph{needle guarantee}) before random subsampling of the remainder.
At deployment, numeric standardisation is folded into the exported inference graph so that~\eqref{eq:selection} can be evaluated on raw feature rows.

\ifconf
\subsection{Linear Ridge Baseline}
\else
\subsubsection{Linear Ridge Baseline}
\fi

The ridge baseline tests how much of the learned selectors' advantage comes from
the hardware-aware representation and \emph{globally linear} feature
combinations, without tree splits or hidden layers.
It uses the same candidate feature matrix as Models~A and~B: numeric columns
from $\phi(p,c,h)$ are standardised with training-set mean and variance (with
non-finite values mapped to zero before scaling), and categorical schedule and
layout fields are one-hot encoded with the same frozen level sets.
Let $\mathbf{x}'_i$ denote the resulting vector for row $i$.

The scorer is ordinary least squares with optional $\ell_2$ shrinkage:
\begin{equation}
  f_{\mathbf{w},b}(\mathbf{x}')
  \;=\;
  \mathbf{w}^{\top}\mathbf{x}' + b,
  \qquad
  \min_{\mathbf{w},b}\;
  \sum_{i\in\mathcal{P}_{\mathrm{tr}}}
  \bigl(y_i - f_{\mathbf{w},b}(\mathbf{x}'_i)\bigr)^2
  + \alpha\,\|\mathbf{w}\|_2^2.
  \label{eq:ridge-loss}
\end{equation}
We solve~\eqref{eq:ridge-loss} in closed form after centring $\mathbf{x}'$ and
$y$ (intercept handled separately); at inference the deployed rule is still
the argmax in~\eqref{eq:selection}.
Unlike the ranking objectives above, training optimises squared error on $y$
directly; validation and test nonetheless score the \emph{selected} kernel
through regret~\eqref{eq:regret}.

\ifconf
\subsection{Selection Regret (The Target Metric)}
\else
\subsubsection{Selection Regret (The Target Metric)}
\fi

For any scorer $f_\theta$ the \emph{regret} of group $g$ is
\ifconf
\begin{equation}
  R(g;f_\theta) \;=\; 1 - y\bigl(g,\hat{c}(g)\bigr)
  \;=\; 1 - \frac{\widehat{T}\!\bigl(g,\hat{c}(g)\bigr)}{T^{\star}_g},
  \qquad
  \hat{c}(g)=\arg\max_{c\in\mathcal{C}(g)} f_\theta\!\bigl(\mathbf{x}(g,c)\bigr).
  \label{eq:regret}
\end{equation}
\else
\begin{equation}
  R(g;f_\theta) \;=\; 1 - y\bigl(g,\hat{c}(g)\bigr)
  \;=\; 1 - \frac{\widehat{T}\!\bigl(g,\hat{c}(g)\bigr)}{T^{\star}_g},
\end{equation}
\begin{equation}
  \hat{c}(g)=\arg\max_{c\in\mathcal{C}(g)} f_\theta\!\bigl(\mathbf{x}(g,c)\bigr).
  \label{eq:regret}
\end{equation}
\fi
$R(g)\in[0,1)$ with $R(g)=0$ iff the selector picks a measured maximiser in $\mathcal{C}(g)$.
Aggregate reports include the mean $\bar{R}=\frac{1}{|\mathcal{G}|}\sum_g R(g)$, median, 95th percentile, and maximum across groups, as well as
\begin{align}
  \mathrm{Top1}(f_\theta) &= \frac{1}{|\mathcal{G}|}\sum_g \mathbf{1}\bigl[\mathrm{rank}(\hat{c}(g))=1\bigr],
  \\
  \mathrm{Top5}(f_\theta) &= \frac{1}{|\mathcal{G}|}\sum_g \mathbf{1}\bigl[\mathrm{rank}(\hat{c}(g))\le 5\bigr],
  \\
  \mathrm{Within5\%}(f_\theta) &= \frac{1}{|\mathcal{G}|}\sum_g \mathbf{1}\bigl[R(g;f_\theta)\le 0.05\bigr],
\end{align}
where $\mathrm{rank}(\cdot)$ is the rank by measured $\widehat{T}$ within the group (minimum-rank tie breaking).
Regret is the quantity we minimise during hyperparameter search and of which we maximise interpretability.

\ifconf
\subsection{Hyperparameter Search}
\else
\subsubsection{Hyperparameter Search}
\fi

We tune $\theta$ with Optuna's Tree-structured Parzen Estimator (TPE), minimising $\mathrm{CV}$ in~\eqref{eq:cv-regret}.
Unless an objective is pinned for an ablation arm, each trial also selects the training loss (ranking versus pointwise regression) as a categorical hyperparameter so TPE can allocate budget to the better-surrogate family.
Trials are executed in parallel on multiple GPUs against a shared SQLite study so workers contribute to one global search.
Table~\ref{tab:hp-xgb} and Table~\ref{tab:hp-mlp} list the searched ranges; conditional knobs apply only when the trial's loss is a ranking objective.

\begin{table}[t]
\centering
\footnotesize
\setlength{\tabcolsep}{3pt}
\renewcommand{\arraystretch}{1.0}
\caption{Hyperparameter search space for Model~A (gradient-boosted trees).}
\label{tab:hp-xgb}
\begin{tabular}{@{}p{0.34\columnwidth} p{0.30\columnwidth} p{0.30\columnwidth}@{}}
\toprule
Knob & Range / values & Notes \\
\midrule
$\eta$ (learning rate) & $[5\times 10^{-3},\,0.3]$ log-uniform & \\
max\_depth & $\{4,\ldots,12\}$ & \\
min\_child\_weight & $[1,\,30]$ log-uniform & \\
subsample & $[0.5,\,1.0]$ & row subsampling \\
colsample\_bytree & $[0.5,\,1.0]$ & column subsampling \\
$\lambda$ (L2) & $[0.01,\,20]$ log-uniform & \\
$\alpha$ (L1) & $[10^{-4},\,10]$ log-uniform & \\
$\gamma$ (min split loss) & $[0,\,5]$ & \\
$n_{\mathrm{estimators}}$ & $\{300,\ldots,1200\}$ & boosting rounds \\
loss & $\{\mathrm{ndcg},\,\mathrm{pairwise},\,\mathrm{mse}\}$ & optional categorical \\
rel\_top\_bands & $\{2,\ldots,5\}$ & ranking losses only \\
\midrule
CV folds $K$ & $5$ & shape-grouped \\
\bottomrule
\end{tabular}
\end{table}

\begin{table}[t]
\centering
\footnotesize
\setlength{\tabcolsep}{3pt}
\renewcommand{\arraystretch}{1.0}
\caption{Hyperparameter search space for Model~B (MLP).}
\label{tab:hp-mlp}
\begin{tabular}{@{}p{0.34\columnwidth} p{0.30\columnwidth} p{0.30\columnwidth}@{}}
\toprule
Knob & Range / values & Notes \\
\midrule
loss & $\{\mathrm{LambdaRank},\,\mathrm{RankNet},$ $\mathrm{mse}\}$ & optional categorical \\
learning rate & $[10^{-4},\,5\times 10^{-3}]$ log-uniform & Adam \\
hidden topology & $10$ preset tuples & depths $2$--$4$, widths $32$--$1024$ \\
dropout & $[0,\,0.4]$ & after each ReLU \\
weight decay & $\{0,10^{-7},10^{-6},10^{-5},10^{-4},$ $10^{-3}\}$ & \\
epochs & $\{20,\ldots,150\}$ & \\
batch size & $\{1024,2048,4096,8192\}$ & MSE only \\
groups per batch & $\{32,64,128,256\}$ & ranking only \\
gradient clip (max norm) & $\{0,0.5,1,5,10\}$ & ranking only \\
max pairs per group & $\{2\mathrm{k},\ldots,64\mathrm{k}\}$ & ranking only \\
gain transform & $\{\mathrm{linear},\,\mathrm{exp}\}$ & LambdaRank only \\
\midrule
CV folds $K$ & $3$ during search & shape-grouped \\
train seeds & $2$ per fold & averaged before reporting $\mathrm{CV}$ \\
pruner & median, 10 startup trials & early-stops weak MLP trials \\
\bottomrule
\end{tabular}
\end{table}

For Model~B the search budget is deliberately lighter per trial ($3$-fold $\times$ two seeds); the winning configuration is re-validated offline at $5$-fold cross-validation with three independent initialisation seeds before the final production fit.
After search completes, the best trial's hyperparameters are \emph{locked} and the model is refit once on all of $\mathcal{P}_{\mathrm{tr}}$.

\ifconf
\subsection{Shape-Grouped Cross-Validation}
\else
\subsubsection{Shape-Grouped Cross-Validation}
\fi

Hyperparameters are chosen by $K$-fold cross-validation on $\mathcal{P}_{\mathrm{tr}}$ only.
Folds partition \emph{base shapes} $(M,N,K)$: all four layouts of a shape belong to the same fold, so validation never sees a training shape under any layout.
Let $\mathcal{S}$ be the set of distinct base shapes in $\mathcal{P}_{\mathrm{tr}}$ and $\{\mathcal{S}^{(k)}\}_{k=1}^{K}$ a partition of $\mathcal{S}$.
For fold $k$, train $f_\theta^{(k)}$ on all groups whose base shape lies in $\mathcal{S}\setminus\mathcal{S}^{(k)}$ using the chosen training objective ($\mathcal{L}_{\mathrm{MSE}}$, $\mathcal{L}_{\mathrm{RankNet}}$, $\mathcal{L}_{\mathrm{LambdaRank}}$, or the tree ranker analogue), then evaluate on validation groups $\mathcal{G}^{(k)}$ by computing~\eqref{eq:regret} via~\eqref{eq:selection}.
The CV score is
\begin{equation}
  \mathrm{CV}(f_\theta) \;=\; \frac{1}{K}\sum_{k=1}^{K}
  \frac{1}{|\mathcal{G}^{(k)}|}\sum_{g\in\mathcal{G}^{(k)}} R\!\bigl(g; f_\theta^{(k)}\bigr).
  \label{eq:cv-regret}
\end{equation}
Bayesian optimisation (Optuna TPE) minimises $\mathrm{CV}$ over tree depth, learning rate, regularisation, ensemble size, hidden widths, dropout, pair caps, and---when not fixed---the choice among ranking versus regression objectives.
$\mathcal{P}_{\mathrm{ev}}$ is never used in~\eqref{eq:cv-regret}.

\ifconf
\subsection{Final Fit and Test Protocol}
\else
\subsubsection{Final Fit and Test Protocol}
\fi

After hyperparameters are fixed, we refit $f_\theta$ once on \emph{all} training groups in $\mathcal{P}_{\mathrm{tr}}$ (no held-out fold).
The reported generalisation result is the evaluation of~\eqref{eq:regret}--\eqref{eq:selection} on the $68$ oracle groups in $\mathcal{P}_{\mathrm{ev}}$, with breakdowns by problem regime (square, tall, wide, skinny-$K$, large).
Baselines computed on the same measured data include uniform random selection, whose expected regret is $\mathbb{E}_g[1-\bar{y}(g)]$ with $\bar{y}(g)=\frac{1}{|\mathcal{C}(g)|}\sum_c y(g,c)$, and a single fixed configuration chosen by best mean $y$ over groups where it appears in at least $80\%$ of evaluation groups.

\ifconf
\subsection{Relation between Training Loss and Validation Metric}
\else
\subsubsection{Relation between Training Loss and Validation Metric}
\fi

Measured throughput always defines the supervision signal ($y$ and $\gamma$), but the default training objectives are ranking losses that penalise incorrect \emph{orderings} within $\mathcal{C}(g)$ rather than absolute error in $\widehat{T}$.
Cross-validation and final testing, by contrast, always score the \emph{deployed} rule~\eqref{eq:selection} through regret~\eqref{eq:regret}.
This deliberate mismatch---optimise surrogate ranking loss, validate selection regret---matches the compiler use case: only the identity of the argmax matters, and mis-ordering among top candidates is far more costly than mis-predicting throughput on clearly suboptimal tiles.

\fi


\section{Evaluation}
\label{sec:evaluation}

We run our experiments on
\ifconf
\else
a gh200 node of the Daint supercomputer on CSCS Alps,
\fi
equipped with 4 GH200 GPUs running SUSE Linux Enterprise Server 15 SP6.
We compile CUTLASS from the commit \texttt{3476ddb7} using CUDA 13.1.
The environment runs Python 3.12.3 with nvidia-matmul-heuristics-0.1.0.27 as the baseline; we are unable to compare against other related baselines~\citec{swann2025tritonblas, yu2023tailoring} since they are not implemented for the Hopper architecture.
\ifconf
\else

\subsection{Training}

We train two model families, an MLP and XGBoost, under both the full hardware-aware representation and a structural-only ablation. For the MLP, we compare MSE, RankNet, and LambdaRank objectives; for XGBoost, we compare squared-error regression, pairwise ranking, and NDCG listwise ranking. We monitor validation NDCG@10 during training.

Figure~\ref{fig:training-mlp} and Figure~\ref{fig:training-xgb} show the validation trajectories. All objectives converge stably. With the full feature set, MSE achieves the highest validation NDCG@10 for both MLP (0.968) and XGBoost (0.960). On structural features, MSE again performs best, reaching 0.969 for MLP and 0.929 for XGBoost.

\begin{figure}
  \centering
    \centering
    \includegraphics[width=\linewidth]{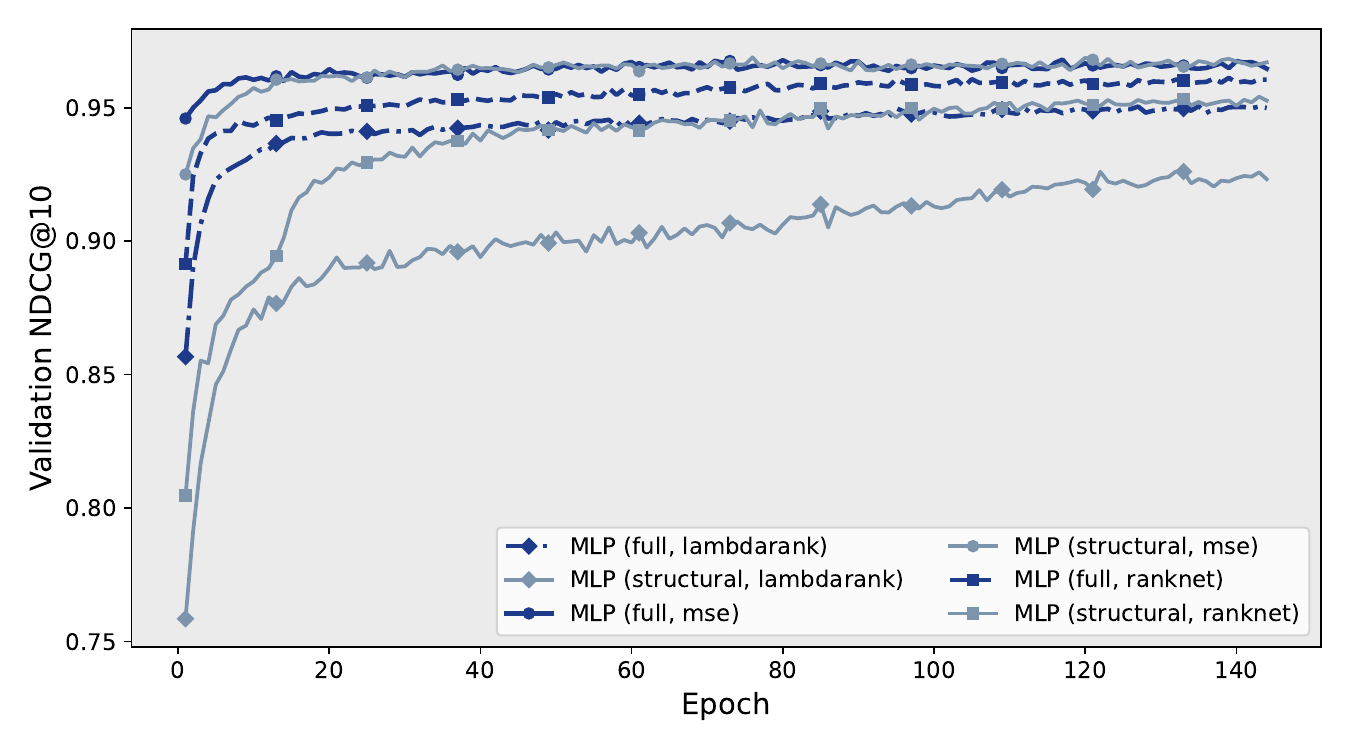}
    \caption{MLP validation NDCG@10.}
    \label{fig:training-mlp}
\end{figure}
\begin{figure}
    \centering
    \includegraphics[width=\linewidth]{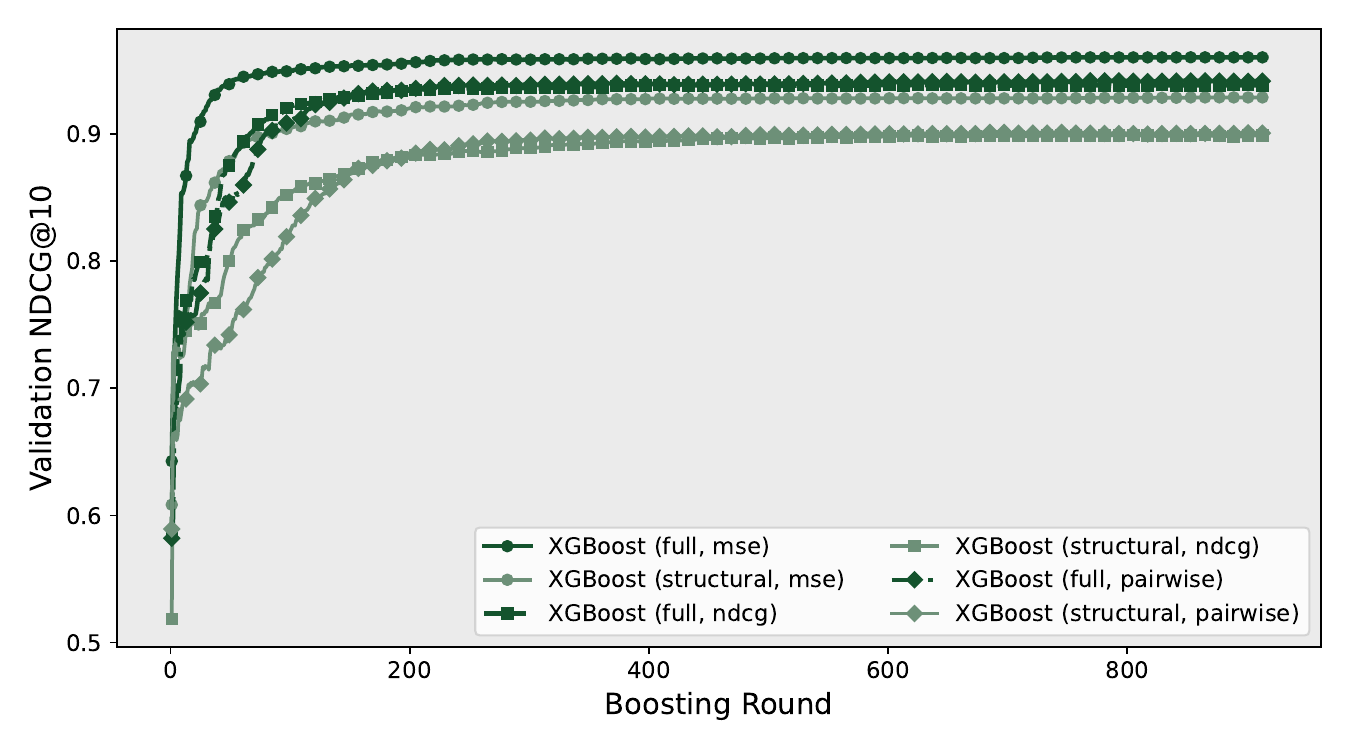}
    \caption{XGBoost validation NDCG@10.}
    \label{fig:training-xgb}
\end{figure}
\fi
\ifconf
\else

Validation NDCG is not perfectly aligned with held-out selection regret. In particular, MLP RankNet attains slightly lower validation NDCG than MSE but achieves the lowest held-out regret among the MLP variants. We therefore report the best held-out configuration for each model family and retain the objective-level comparison as a training analysis rather than treating validation NDCG as a direct proxy for final selection quality.

Hardware-aware features improve held-out selection consistently across objectives. The gain is modest for the MLP and larger for XGBoost, suggesting that tree models benefit more strongly from exposing the derived hardware behavior explicitly.

Figure~\ref{fig:full-vs-struct-obj} compares full and structural features across all objectives. 
\begin{figure}[t]
  \centering
  \includegraphics[width=\linewidth]{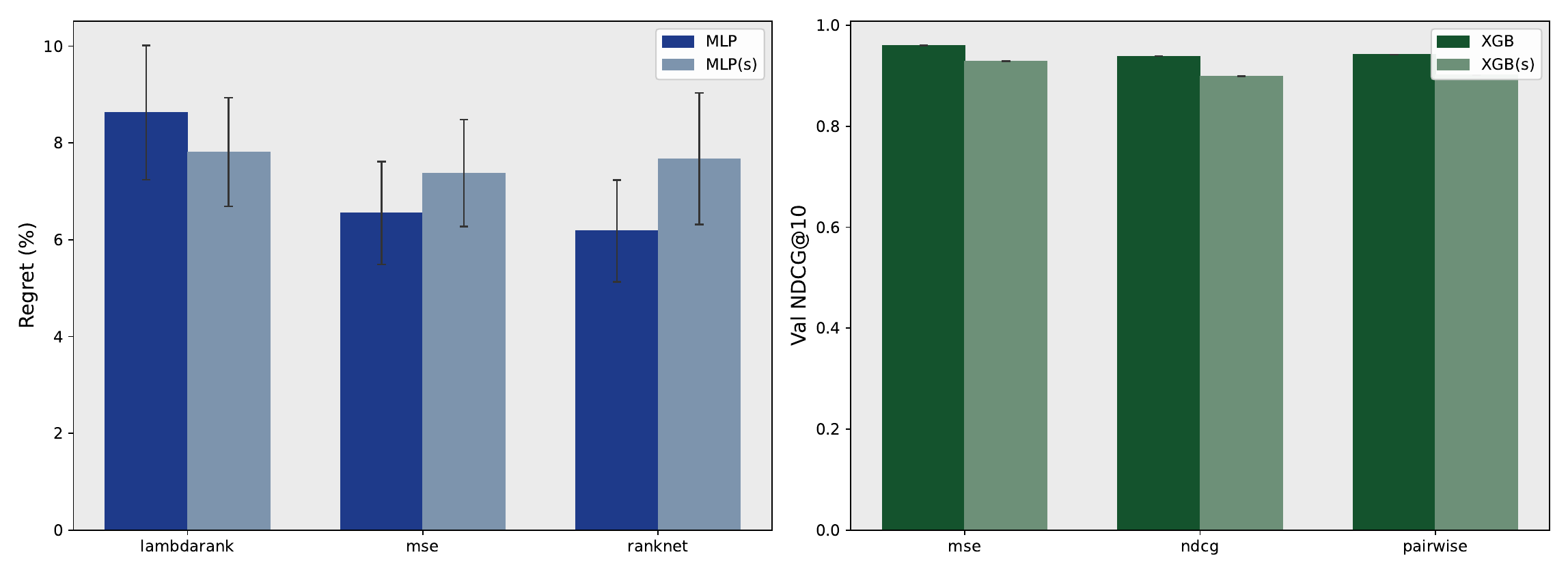}
  \caption{Full and structural features across training objectives.}
  \label{fig:full-vs-struct-obj}
\end{figure}

\fi

\subsection{Evaluation on the Exhaustive Dataset}

\ifconf
\else
\begin{figure}
    \centering
    \includegraphics[width=\linewidth]{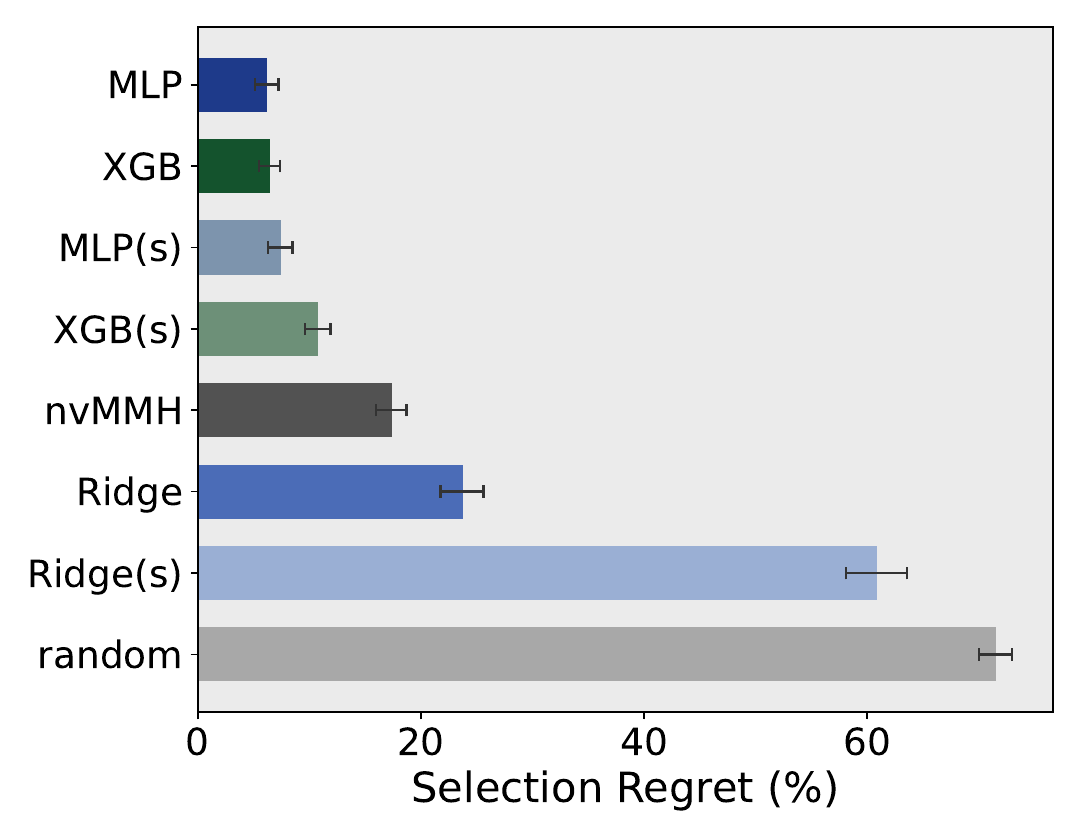}
    \caption{Mean selection regret.}
    \label{fig:eval-main}
\end{figure}
\begin{figure}
    \centering
    \includegraphics[width=\linewidth]{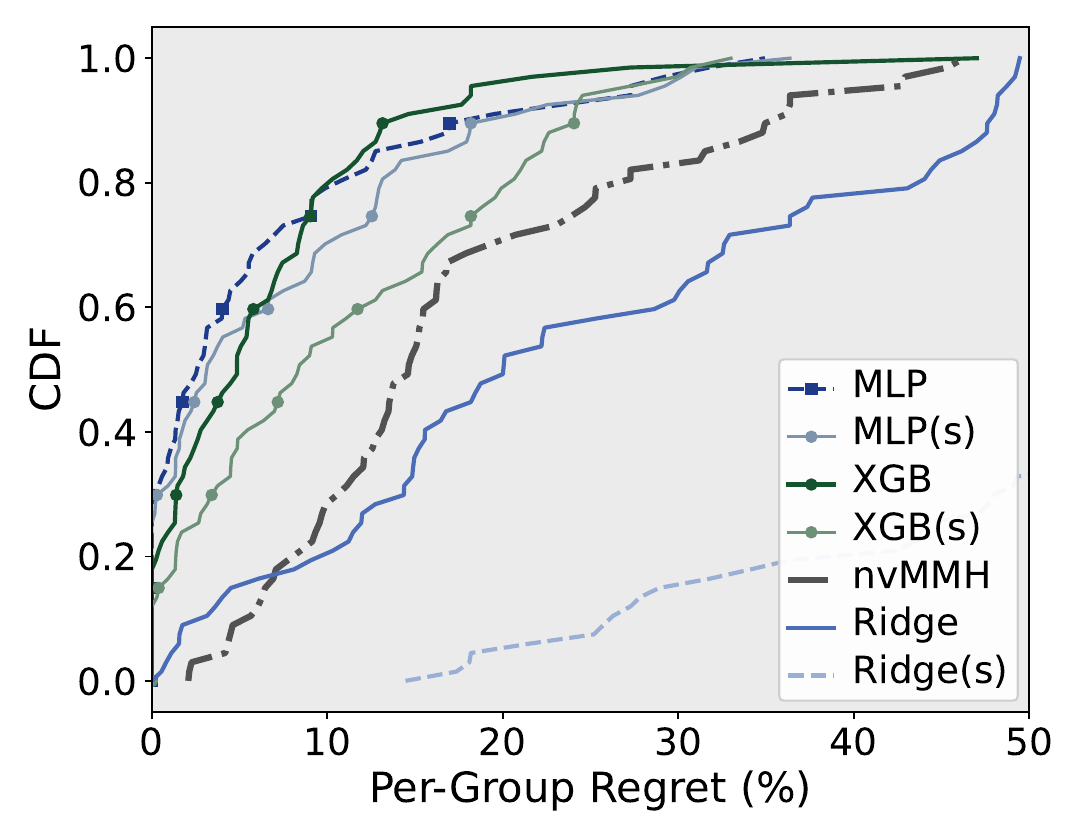}
    \caption{CDF of per-group selection regret.}
    \label{fig:eval-cdf}
\end{figure}
\fi

We consider an MLP with \(1.76\,\mathrm{M}\) trainable parameters and an XGBoost ensemble of 916 trees with maximum depth~7.
\ifconf
\else

Table~\ref{tab:eval-main} summarizes the main results.
\fi
On the held out evaluation dataset (Figure~\ref{fig:eval-main}), the best hardware-aware MLP reduces mean regret to \(6.2\%\), compared with \(17.3\%\) for the top-1 nvMMH recommendation and \(71.5\%\) for a random valid candidate. Hardware-aware XGBoost MSE follows closely at \(6.4\%\). Hardware-aware features reduce mean regret on the best MLP from \(7.4\%\) to \(6.2\%\); for XGBoost, they reduce it from \(10.7\%\) to \(6.4\%\). The larger XGBoost gain indicates that the derived features substantially simplify the mapping from configuration parameters to performance.

\ifconf
\else

\begin{table}[t]
  \centering
  \caption{Selection quality on 68 exhaustively measured shape--layout groups.}
  \label{tab:eval-main}
  \footnotesize
  \setlength{\tabcolsep}{4pt}
  \begin{tabular}{@{}p{0.34\linewidth}rrrrr@{}}
    \toprule
    Method & Mean & Median & Within 1\% & Within 5\% & Top-1 \\
    \midrule
    MLP (full)       & 6.2\%  & 2.6\%  & 36.8\% & 63.2\% & 27.9\% \\
    MLP (structural)     & 7.4\%  & 3.1\%  & 32.4\% & 55.9\% & 26.5\% \\
    XGBoost (full)       & 6.4\%  & 4.9\%  & 25.0\% & 52.9\% & 19.1\% \\
    XGBoost (structural) & 10.7\% & 8.3\%  & 17.6\% & 39.7\% & 13.2\% \\
    nvMMH (best-of-$K$)      & 12.0\% & 10.2\% & 0.0\%  & 25.0\% & 0.0\% \\
    nvMMH (top-1)            & 17.3\% & 14.6\% & 0.0\%  & 10.3\% & 0.0\% \\
    Ridge (full)         & 23.7\% & 20.0\% & 4.4\%  & 16.2\% & 1.5\% \\
    Ridge (structural)       & 60.8\% & 63.6\% & 0.0\%  & 0.0\%  & 0.0\% \\
    Analytical score         & 63.7\% & 63.8\% & 0.0\%  & 0.0\%  & 0.0\% \\
    Random                   & 71.5\% & ---     & ---     & ---     & --- \\
    \bottomrule
  \end{tabular}
\end{table}

The learned selectors are also substantially more likely to return near-oracle kernels. MLP RankNet selects a kernel within \(5\%\) of the oracle on \(63.2\%\) of groups, and XGBoost MSE does so on \(52.9\%\). In comparison, nvMMH top-1 reaches this threshold on only \(10.3\%\) of groups. Even an oracle choice among nvMMH's top-\(K\) recommendations reaches it on only \(25.0\%\), showing that much of the remaining gap arises from the candidate ranking itself rather than from the final choice within the shortlist.

We next ask whether the hardware-aware representation could support strong kernel
selection without a nonlinear learned model. We compare against two simpler
alternatives: a validation-selected analytical score that combines a small set
of mechanistic hardware terms, and a ridge regressor that linearly scores
candidates using the same feature representation as our learned selectors
(Table~\ref{tab:eval-main}).

The analytical score performs poorly, reaching \(63.7\%\) mean regret and never
selecting a configuration within \(5\%\) of the oracle across the 68 exhaustive
evaluation groups. Individual hardware features correlate with throughput in
training, yet this compact composition of them fails to rank the
performance tail: knowing which mechanisms matter is not the same as knowing how
they interact. Under the same linear model family, structural features alone
yield \(60.8\%\) regret, while adding the hardware-aware representation reduces
this to \(23.7\%\). Thus the proposed features expose substantial selection
signal even without a large nonlinear model. However, this global linear
combination remains worse than nvMMH (\(17.3\%\)) and far behind the nonlinear
MLP and XGBoost selectors (\(6.2\%\) and \(6.4\%\)).

These results separate representation from model expressivity. Hardware-aware
features make the problem considerably easier to learn, but a single global
linear mapping is insufficient for strong performance in this setting. The
nonlinear learned models are therefore important for translating mechanistic
features into accurate kernel rankings.
\fi

Figure~\ref{fig:eval-cdf} shows that MLP and XGBoost place most of their
mass at low regret: hardware-aware MLP and XGBoost have medians of
\(2.6\%\) and \(4.9\%\), compared with \(14.6\%\) for nvMMH, and hardware-aware
models sit above their structural counterparts throughout the curve.
\ifconf
\else
Figure~\ref{fig:eval-within5} reports the fraction of groups within \(5\%\) of oracle throughput.

\begin{figure}[t]
  \centering
  \includegraphics[width=\linewidth]{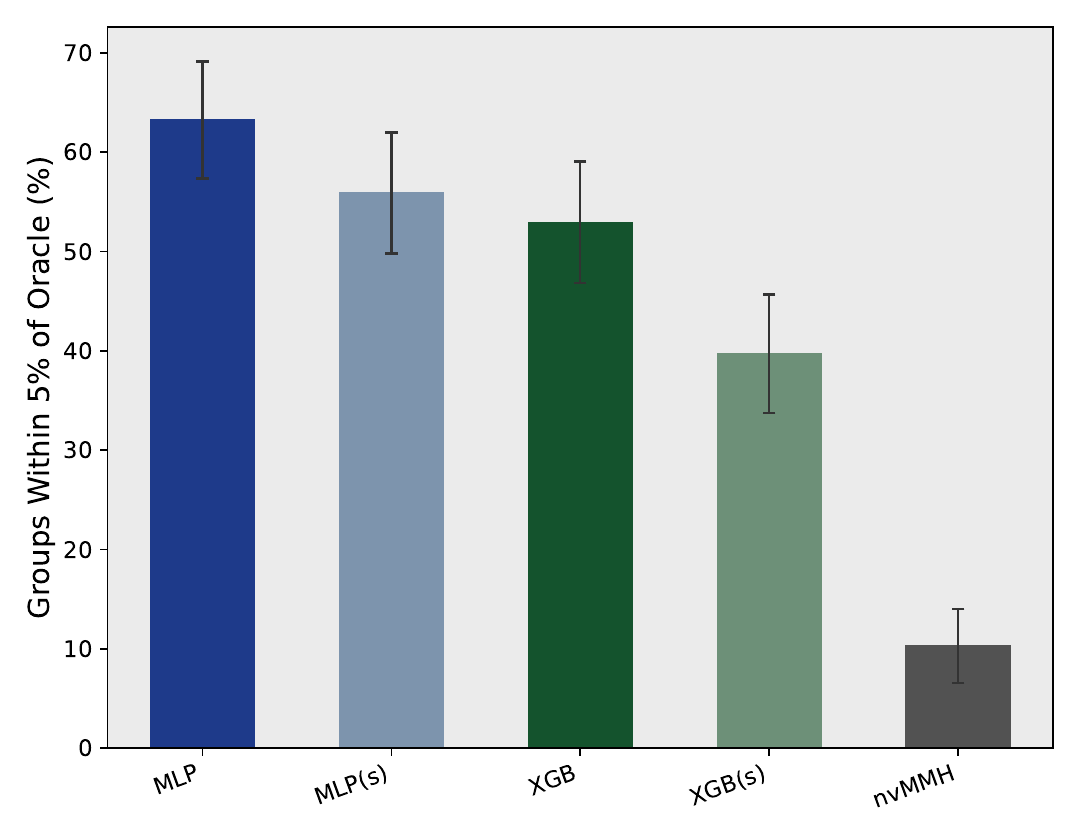}
  \caption{Fraction of evaluation groups selected within \(5\%\) of oracle throughput.}
  \label{fig:eval-within5}
\end{figure}

\paragraph{Regime breakdown.}
Selection quality varies by problem geometry. MLP RankNet achieves mean regret of \(0.9\%\) on wide GEMMs and \(3.3\%\) on tall GEMMs, while square and skinny-\(K\) problems remain more difficult. The same trend appears for XGBoost and the structural ablations. nvMMH performs worse across all regimes, with mean regret between approximately \(12\%\) and \(19\%\).

\begin{figure}[t]
  \centering
  \includegraphics[width=\linewidth]{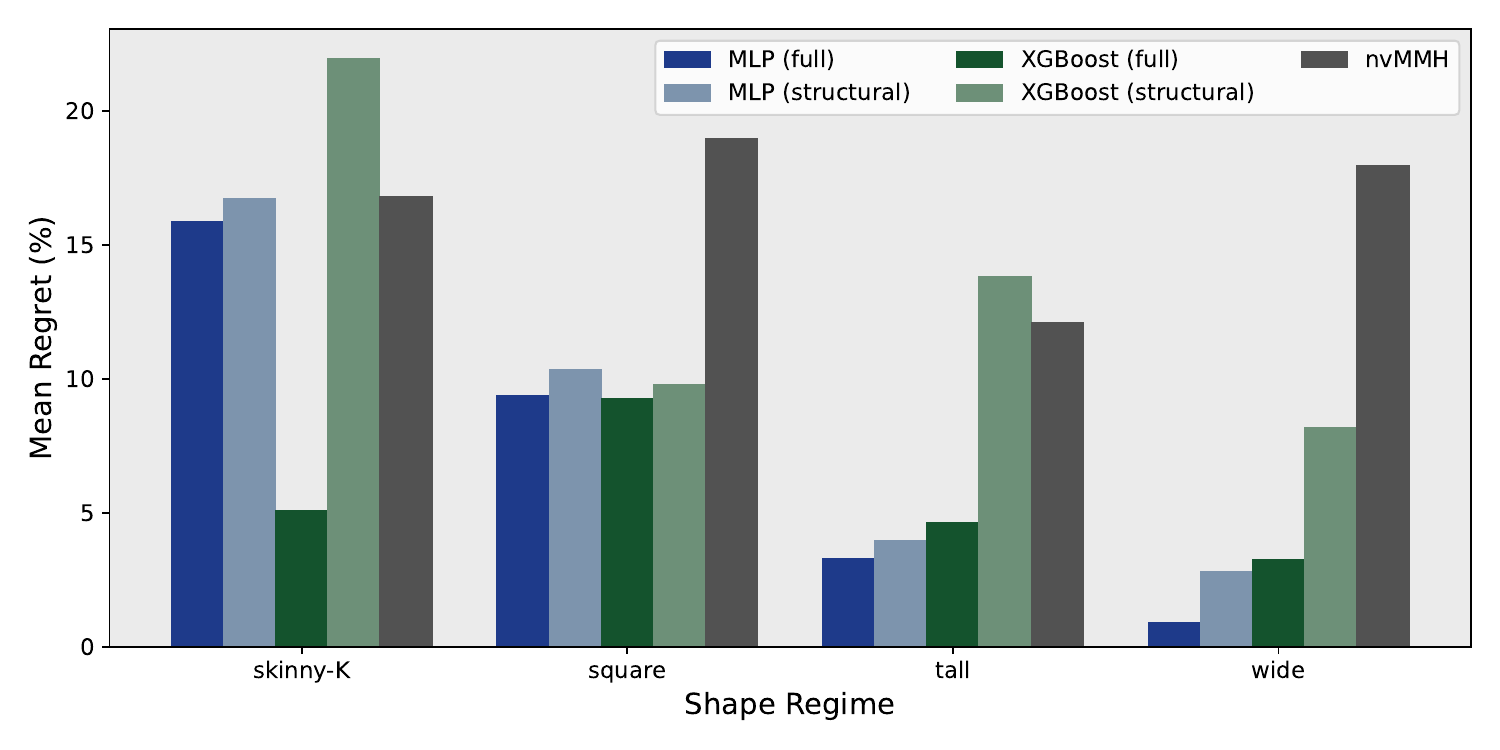}
  \caption{Mean regret by GEMM shape regime.}
  \label{fig:eval-regime-nvmmh}
\end{figure}

\begin{figure}[t]
  \centering
  \includegraphics[width=\linewidth]{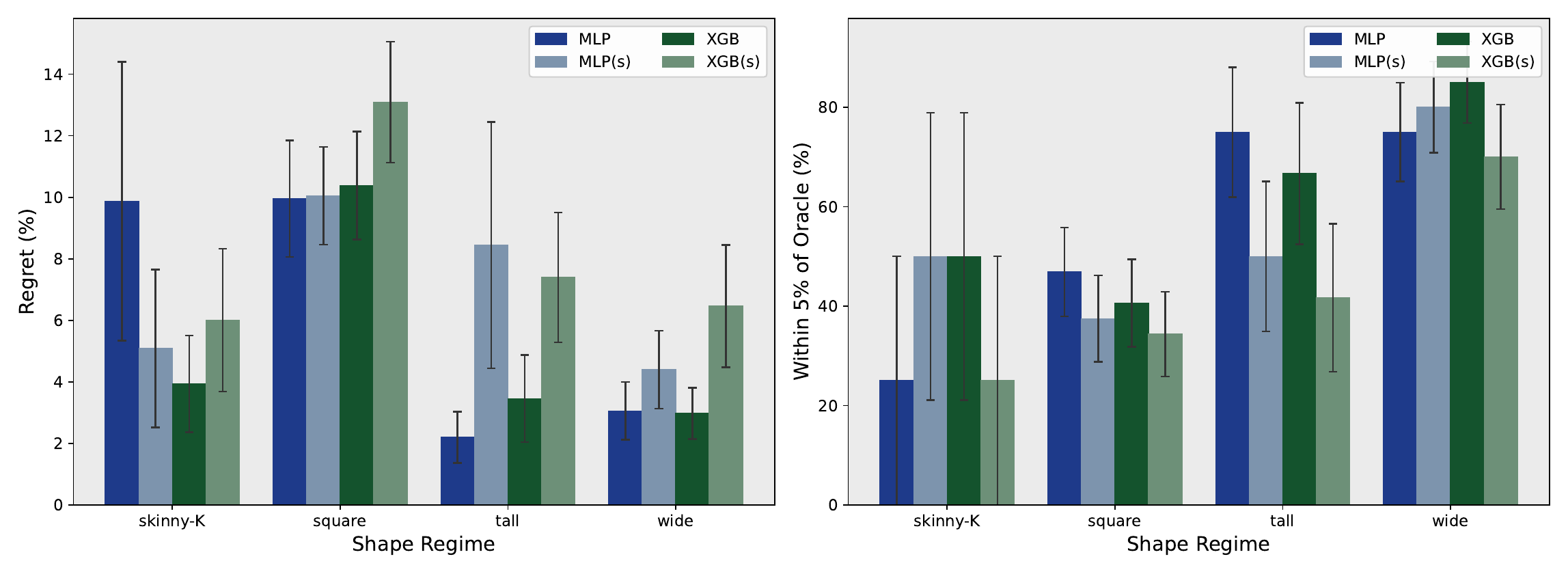}
  \caption{Regret and within-\(5\%\) rate by regime for the best full and structural variants.}
  \label{fig:eval-regime-best}
\end{figure}
\fi

\subsection{Model Size Ablations}

We next ask whether the benefit of hardware-aware features persists as model
capacity changes. We fix the MSE objective and sweep MLP widths from
$16\!\times\!8$ to $1024\!\times\!1024\!\times\!512\!\times\!256$, and
XGBoost \texttt{max\_depth} over
$\{2,3,4,6,8,11,14\}$, using three seeds per configuration and both feature
sets. Figure~\ref{fig:capacity-mlp} and Figure~\ref{fig:capacity-xgb} report
held-out mean selection regret.

\ifconf
\else
\begin{figure}
    \centering
    \includegraphics[width=\linewidth]{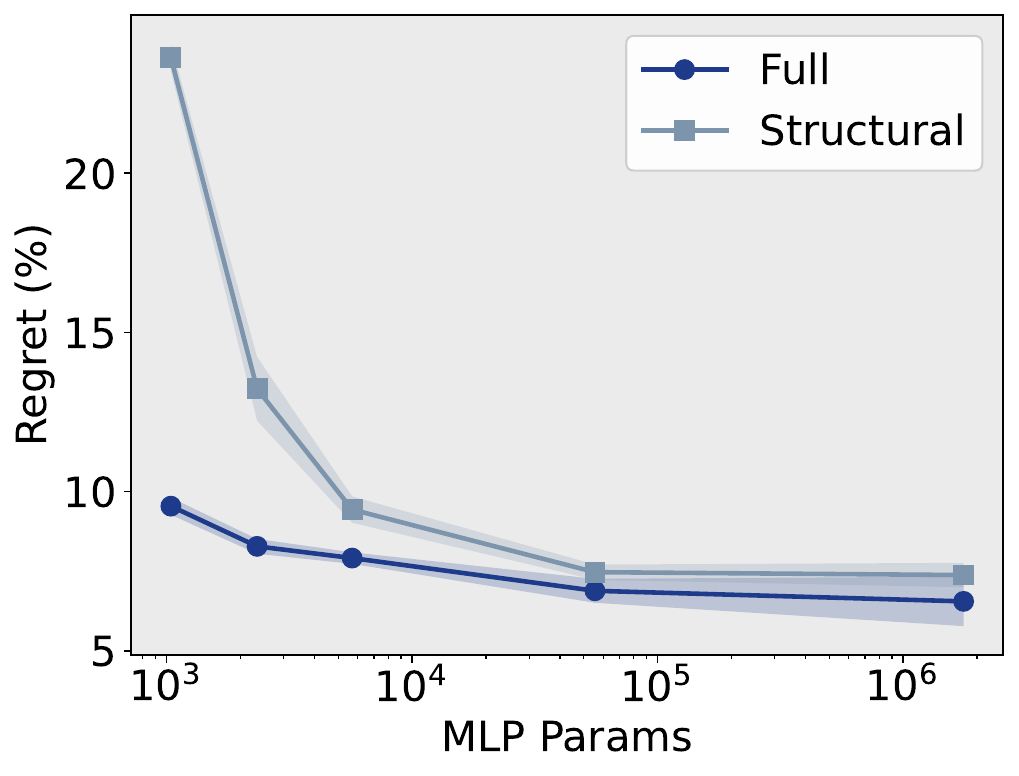}
    \caption{MLP of different parameter counts.}
    \label{fig:capacity-mlp}
\end{figure}
\begin{figure}
    \centering
    \includegraphics[width=\linewidth]{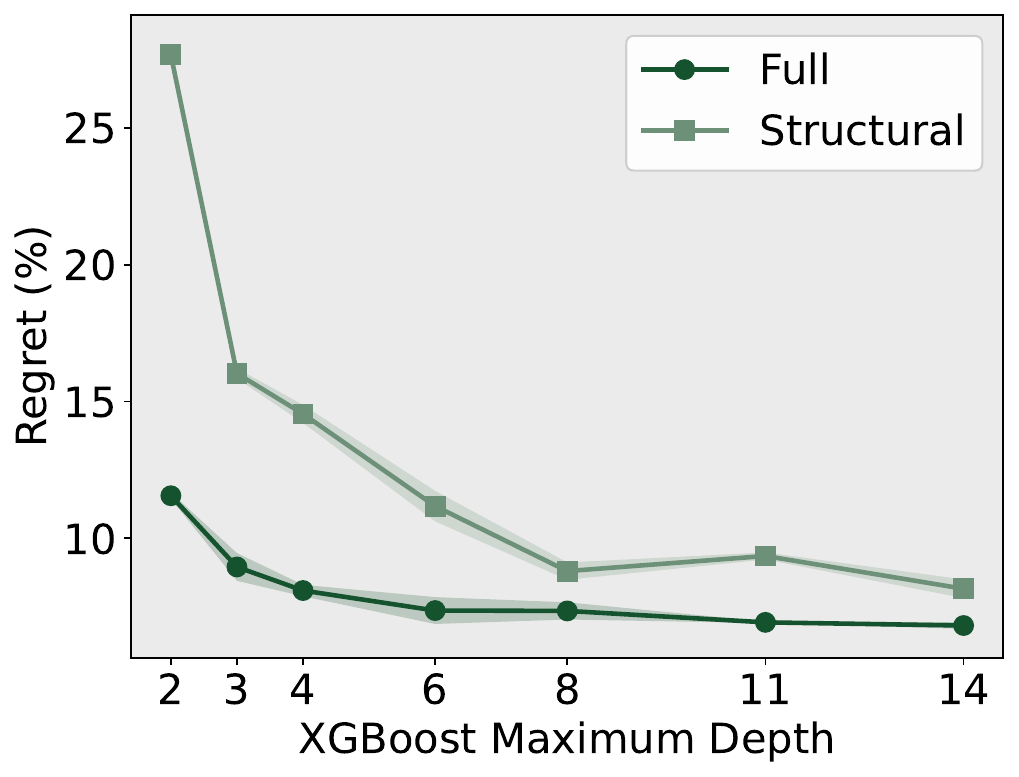}
    \caption{XGB of different depths.}
    \label{fig:capacity-xgb}
\end{figure}
\fi

Hardware-aware features substantially reduce the model capacity required for
accurate kernel selection. At the smallest MLP sizes, the structural model
incurs up to $14\%$ higher regret, whereas the hardware-aware representation
already captures much of the performance structure of the search space.
As capacity increases, structural models gradually recover this gap, and only
the largest MLP ($\sim$1.7\,M parameters) matches the full representation. This
effect is even more pronounced for tree models. XGBoost benefits from the
hardware-aware representation throughout the entire depth sweep and retains an
advantage even at the largest tested depth. In this sense, the
hardware-aware features do not merely improve final accuracy; they make
kernel-selection itself easier to learn.


\subsection{Generalization and Broad GEMM Evaluation}
\label{sec:gemm-eval}

\ifconf\else
The exhaustive evaluation measures selection quality against a true in-space oracle, but necessarily covers only a small number of problems.
\fi
We
\ifconf\else
therefore
\fi
evaluate whether the same selectors generalize at scale to \(8{,}000\) previously unseen BF16 GEMMs.
\ifconf\else Half of the shapes are \(32\)-aligned, while the remainder are only \(8\)-aligned, introducing off-grid dimensions that are sparsely represented in the training sweep.
The broad evaluation uses \(2{,}000\) \((M,N,K)\) triples held out from the production BF16 autotuning databases and evaluates each under all four operand layouts (TN, TT, NN, NT), yielding \(8{,}000\) problems. Half of the shapes are fully \(32\)-aligned; the remainder are \(8\)-aligned but not \(32\)-aligned. The latter set tests generalization to off-grid dimensions that are largely absent from the regular training sweep. The performance results are in Table~\ref{tab:gemm-broad}.

We compare four learned selectors---MLP and XGBoost MSE, each with the full hardware-aware representation and the structural-only ablation---against nvMMH. Each learned model returns a single rank-1 CUTLASS configuration. nvMMH emits a rank-1 tile and cluster recommendation but does not completely specify the CUTLASS mainloop, epilogue, and scheduler configuration. We therefore materialize all valid variants consistent with the nvMMH recommendation and report the best measured variant. This gives nvMMH an optimistic comparison relative to the learned selectors, which each receive only one selected kernel.

All candidates are compiled and benchmarked on the same GH200 node using the same warmup, iteration-count, and operand-rotation protocol as the training measurements. To isolate kernel ranking from tile-scheduler choices, every method uses the same rank-1 rasterization, swizzle, and split-\(K\) settings during benchmarking. nvMMH retains its recommended tile geometry, while the learned selectors retain their independently selected tile, schedule, cluster, and pipeline configuration.

\fi
Figure~\ref{fig:roofline-broad-gemm} places the selected kernels against the
GH200 roofline.
\ifconf\else
For the roofline analysis, we use a GH200 peak BF16 tensor-core throughput of \(989.5\,\mathrm{Tflop/s}\) and memory bandwidth of \(4\,\mathrm{TB/s}\). Arithmetic intensity is computed from the mathematical GEMM work and minimum operand traffic. Each light point corresponds to one measured GEMM, while the bold curves show binned median throughput.
\fi
The hardware-aware MLP and XGBoost selectors produce nearly
identical performance envelopes despite their very different model
architectures, and both consistently outperform nvMMH across the full
arithmetic-intensity range. The improvement persists from memory-bound GEMMs
through the transition region and into compute-bound problems, indicating that
the learned representation captures performance effects beyond any single
bottleneck regime. 
%
%
\ifconf
\else
\begin{figure*}
  \begin{subfigure}{0.48\linewidth}
    \centering
    \includegraphics[width=\linewidth]{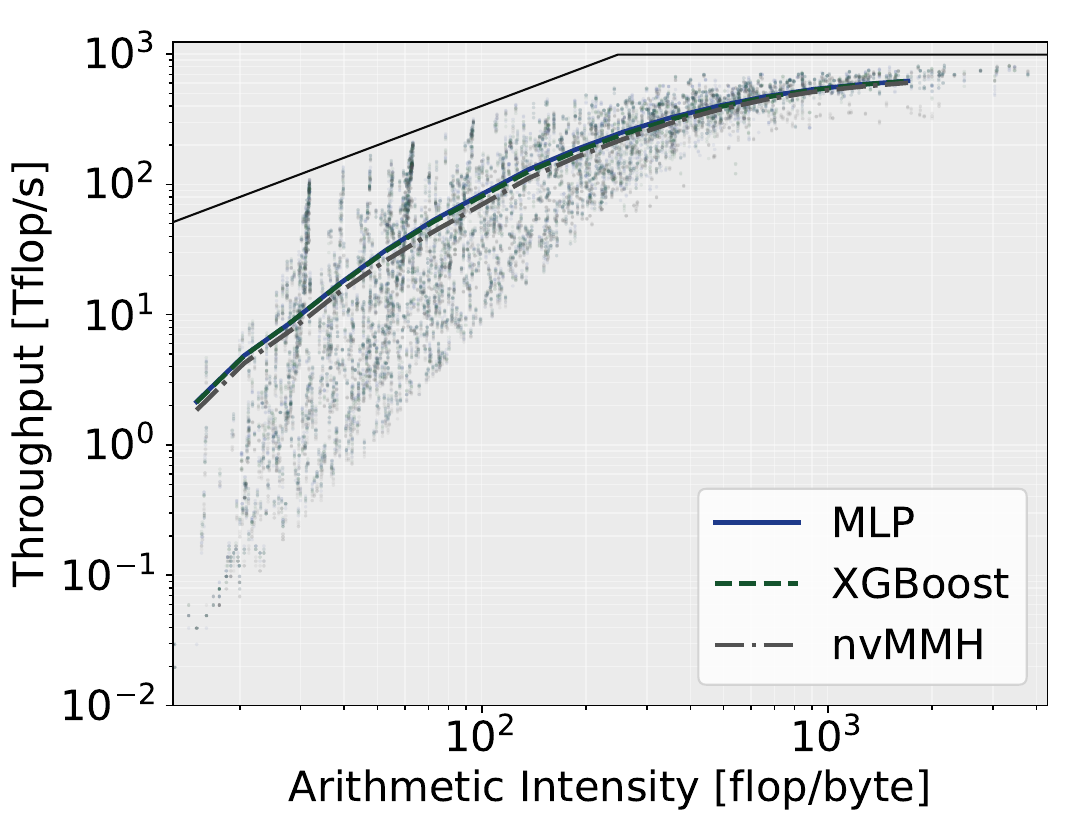}
    \caption{GEMM evaluation against GH200 roofline.}
    \label{fig:roofline-broad-gemm}
  \end{subfigure}
  ~
  \begin{subfigure}{0.48\linewidth}
    \centering
    \includegraphics[width=\linewidth]{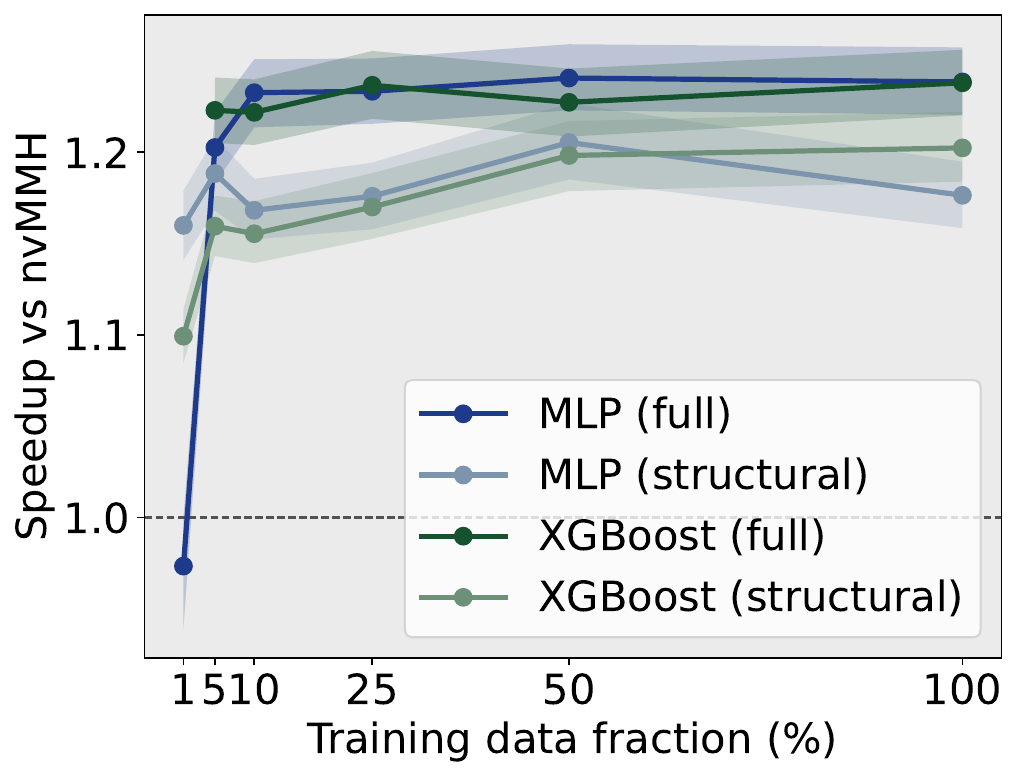}
    \caption{Cross-precision transfer to FP32.}
    \label{fig:transfer-fp32}
  \end{subfigure}
  
   \begin{subfigure}{0.48\linewidth}
    \centering
    \includegraphics[width=\linewidth]{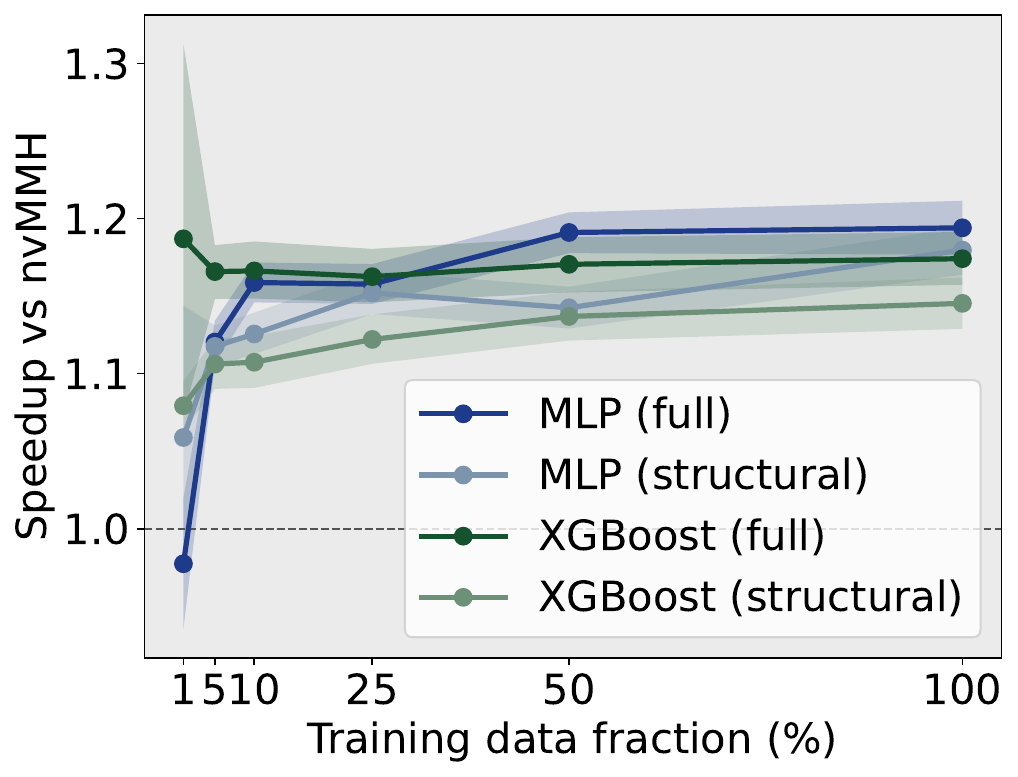}
    \caption{Cross-precision transfer to FP8.}
    \label{fig:transfer-fp8}
  \end{subfigure}
  ~
  \begin{subfigure}{0.48\linewidth}
    \centering
    \includegraphics[width=\linewidth]{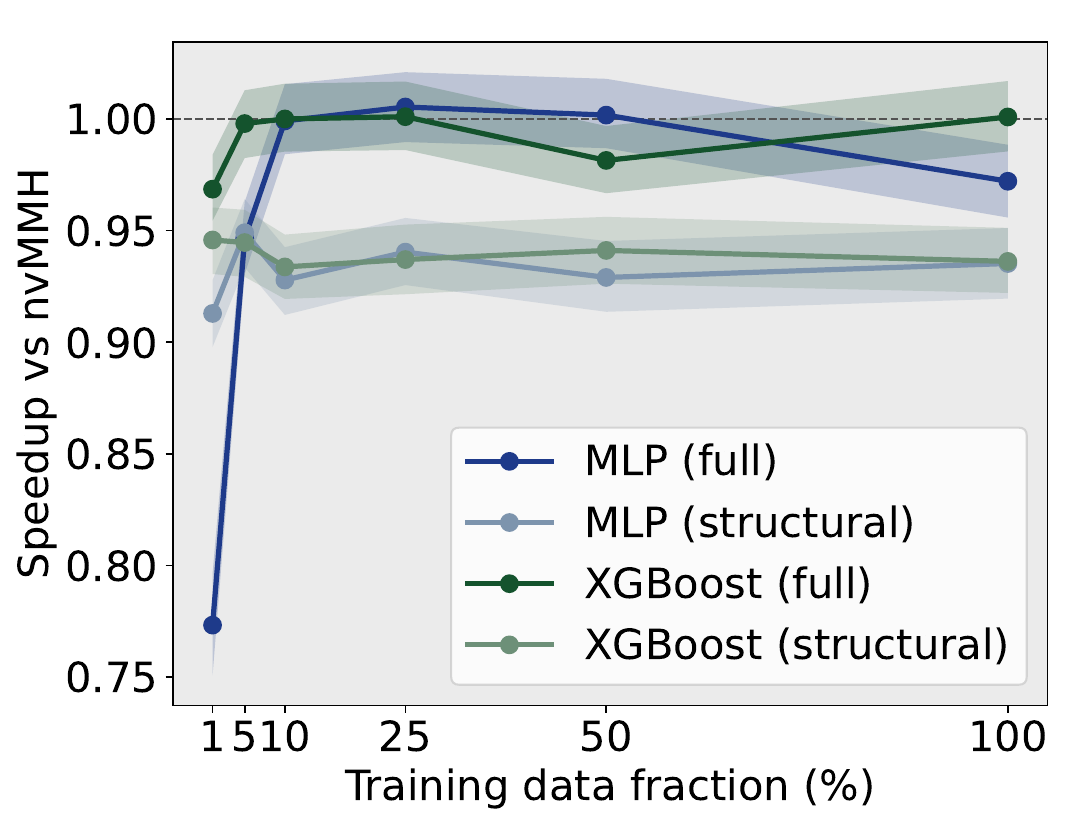}
    \caption{Cross-epilogue transfer.}
    \label{fig:transfer-fusion}
  \end{subfigure}
  \caption{Model Generalization and Transfer experiments. }
\end{figure*}
\fi
\ifconf
\else

\begin{table}[t]
  \centering
  \caption{Broad GEMM evaluation on \(8{,}000\) held-out problems.}
  \label{tab:gemm-broad}
  \footnotesize
  \setlength{\tabcolsep}{4pt}
  \begin{tabular}{@{}llrrrr@{}}
    \toprule
    Model & Features & Geo-mean \% roof & vs.\ nvMMH & Win \% \\
    \midrule
    MLP MSE     & Full            & 11.9\% & 1.14$\times$ & 79.6\% \\
    MLP MSE     & Structural only & 11.5\% & 1.10$\times$ & 73.9\% \\
    \midrule
    XGBoost MSE & Full            & 11.8\% & 1.13$\times$ & 77.4\% \\
    XGBoost MSE & Structural only & 11.4\% & 1.09$\times$ & 70.1\% \\
    \midrule
    nvMMH & --- & 37.6 & 10.5\% & --- & --- \\
    \bottomrule
  \end{tabular}
\end{table}
\fi

\ifconf
\else
All methods achieve \(100\%\) execution coverage on the \(8{,}000\) problems. The hardware-aware MLP reaches \(42.9\,\mathrm{Tflop/s}\) geometric-mean throughput compared with \(37.6\,\mathrm{Tflop/s}\) for nvMMH, corresponding to a \(1.14\times\) improvement. Hardware-aware XGBoost reaches \(42.4\,\mathrm{Tflop/s}\), or \(1.13\times\) nvMMH. The structural variants remain competitive at \(41.2\) and \(41.0\,\mathrm{Tflop/s}\), but both are consistently below their hardware-aware counterparts.

Measured per problem, the full-feature MLP outperforms nvMMH on \(79.6\%\) of cases and full-feature XGBoost on \(77.4\%\). Even the structural variants win on \(73.9\%\) and \(70.1\%\), respectively. Together with the exhaustive-oracle evaluation, this shows that the learned selectors' lower selection regret translates directly into higher realized throughput over a substantially broader problem distribution.
\fi

\subsection{Transfer Learning Experiments}
\label{sec:transfer}

We evaluate whether a selector trained on BF16 can be adapted efficiently
to new precisions. We transfer the BF16 models to FP32 and FP8 E4M3 GEMMs
using nested subsets of the corresponding target-dtype training corpus, and
evaluate on held-out problems against nvMMH.

Figures~\ref{fig:transfer-fp32} and~\ref{fig:transfer-fp8} show that only modest
target-dtype supervision is required to recover strong performance. On FP32,
hardware-aware selectors already reach approximately \(1.20\times\)--\(1.22\times\)
geometric-mean speedup over nvMMH using only \(5\%\) of the available target
shapes, and improve to approximately \(1.24\times\) with the full target
corpus. The corresponding structural models generally trail the
hardware-aware representation, particularly in the low-data regime.
FP8 E4M3 presents a noisier, more difficult transfer setting, but the same trend
persists. Hardware-aware MLPs reach \(1.12\times\) speedup with \(5\%\) of the
target shapes and \(1.19\times\) with the full corpus, while
XGBoost remains around \(1.15\times\)--\(1.17\times\) across the sweep.
Together, these results show that the selectors retain useful
structure across changes in numerical precision, and that exposing
candidate-induced hardware behavior improves adaptation when target-domain
measurements are limited.

We repeat the same finetuning protocol for FP16 epilogue-fusion GEMMs, fusing ReLU, GeLU, and/or bias, warm-starting
from the unfused BF16 checkpoints.
Figure~\ref{fig:transfer-fusion} summarizes the results on held-out
fusion GEMMs.
With only \(10\%\) of the fusion corpus, hardware-aware MLP and XGBoost already
match nvMMH (\(1.00\times\) and \(1.00\times\) geometric-mean speedup,
respectively) and outperform the structural ablation (\(0.93\times\)).
\ifconf
\else

\subsubsection{Transfer Learning Details}
\label{sec:app-transfer}

For each target dtype, FP32 TN and FP8 E4M3 TN, we construct nested subsets
containing \(\{1,5,10,25,50,100\}\%\) of the 593 target-dtype base training
shapes. The same \((M,N,K)\) subsets are used for FP32 and FP8, while the
measured candidate kernels differ according to dtype-specific validity and
compilation behavior. The fractions therefore measure target-domain shape
coverage rather than equal numbers of kernel measurements.

We fix the MSE objective and reuse the model architectures selected for the
BF16 study. MLP transfer initializes from the corresponding BF16 checkpoint,
while XGBoost is warm-started from the BF16 model by adding 200 trees. We
evaluate both the complete hardware-aware representation and the structural
ablation. Evaluation uses a disjoint held-out TN set and the same benchmarking
protocol as Section~\ref{sec:gemm-eval}. Learned selectors emit one rank-1
kernel, while nvMMH is evaluated using its best valid schedule realization.

At full target-dtype coverage, FP32 transfer reaches \(1.24\times\)
geometric-mean speedup for both hardware-aware MLP and XGBoost, with win rates
of \(74.6\%\) and \(78.4\%\), respectively. Structural variants remain strong
but consistently lower, reaching \(1.18\times\) and \(1.20\times\).
FP8 remains more challenging, particularly in execution coverage, but the
hardware-aware MLP and XGBoost still achieve \(1.19\times\) and
\(1.17\times\) geometric-mean speedup over nvMMH.

\begin{table}[t]
  \centering
  \caption{Cross-precision transfer at \(100\%\) of the target TN training
  shapes.}
  \label{tab:transfer-100}
  \footnotesize
  \setlength{\tabcolsep}{4pt}
  \begin{tabular}{@{}llrrr@{}}
    \toprule
    Dtype & Method & Coverage & Speedup & Win \% \\
    \midrule
    FP32 &
    MLP (full) & 89.5\% & 1.24$\times$ & 74.6\% \\
    FP32 &
    XGBoost (full) & 96.7\% & 1.24$\times$ & 78.4\% \\
    FP32 &
    MLP (structural) & 96.1\% & 1.18$\times$ & 71.8\% \\
    FP32 &
    XGBoost (structural) & 91.7\% & 1.20$\times$ & 68.6\% \\
    \midrule
    FP8 E4M3 &
    MLP (full) & 65.4\% & 1.19$\times$ & 47.4\% \\
    FP8 E4M3 &
    XGBoost (full) & 73.0\% & 1.17$\times$ & 51.2\% \\
    FP8 E4M3 &
    MLP (structural) & 87.5\% & 1.18$\times$ & 64.6\% \\
    FP8 E4M3 &
    XGBoost (structural) & 66.0\% & 1.15$\times$ & 43.6\% \\
    \bottomrule
  \end{tabular}
\end{table}

The low-data regime is particularly informative. At \(5\%\) of the target
shape pool, the hardware-aware FP32 models already exceed nvMMH by roughly
\(20\%\)--\(22\%\), despite observing only 30 target-dtype GEMM geometries.
The corresponding FP8 models also exceed nvMMH, demonstrating that much of the
useful structure learned from BF16 survives the precision change and can be
adapted with comparatively little target-domain data.

We apply the same finetuning protocol to the FP16 epilogue-fusion GEMMs.
Selectors are warm-started from the unfused BF16 checkpoints and finetuned on
nested subsets of the 593-shape fusion training corpus, using the same
\(\{1,5,10,25,50,100\}\%\) fractions as the precision-transfer study.
We evaluate \(1{,}000\) held-out problems with in-vocabulary fusion kinds
assigned at evaluation time.

At full fusion training coverage, hardware-aware XGBoost matches nvMMH
(\(1.00\times\) geometric-mean speedup), while the
hardware-aware MLP reaches \(0.97\times\).
Structural variants remain below at \(0.94\times\)--\(0.94\times\).
All configurations achieve \(100\%\) execution coverage.

The low-data fusion regime mirrors the precision-transfer trend. With only
\(10\%\) of the fusion corpus (59 shapes), hardware-aware MLP and XGBoost
already reach \(1.00\times\) geometric-mean speedup over nvMMH, while
structural models remain lower.
The hardware-aware MLP peaks at \(1.01\times\) with \(25\%\) coverage before
settling near parity.
\fi

\subsection{DeepBench Case Study}
\label{sec:deepbench}

\ifconf
\else
To test whether the same selectors generalize to an \emph{externally} defined
workload, we evaluate on dense GEMMs from DeepBench~\cite{baidu2016deepbench}, a
public catalogue of speech-recognition and language-modeling kernels compiled
for 2016-era training and inference stacks.
Our catalogue contains \(151\) unique \((M,N,K,\mathrm{layout})\) problems
after deduplication.  The evaluation set contains
\(107\) training, \(39\) inference-server, and \(5\) inference-device problems.

\begin{figure}[t]
\centering
\includegraphics[width=\linewidth]{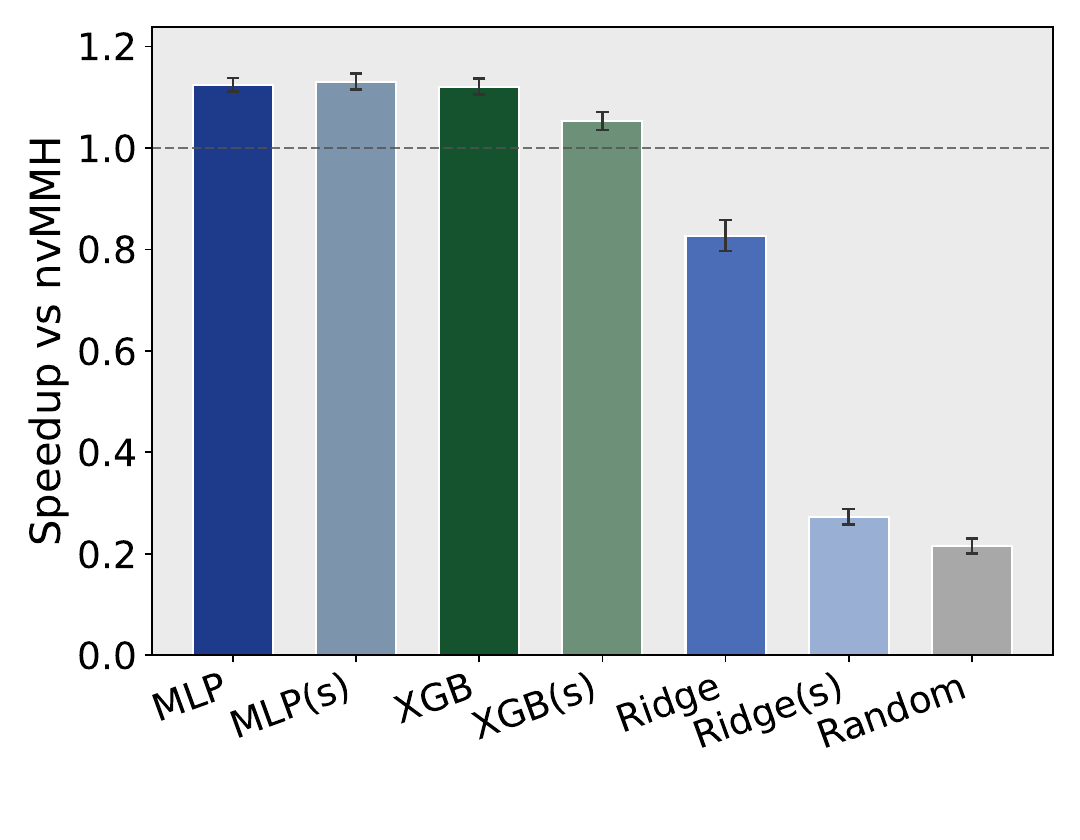}
\caption{Speedup vs.\ nvMMH.}
\label{fig:deepbench-bars}
\end{figure}

Table~\ref{tab:deepbench} and Figure~\ref{fig:deepbench-bars} summarize
the results. All learned selectors achieve \(100\%\) execution coverage on the
retained problems. Hardware-aware MLP and XGBoost improve geometric-mean
throughput by \(12\%\)--\(13\%\) over nvMMH and win on \(85\%\)--\(96\%\) of
problems. The structural MLP matches the full model (\(1.13\times\)),
consistent with the observation that structural features extrapolate more
robustly when problem geometry departs from the training grid. Ridge and random
baselines remain well below nvMMH, confirming that the gains require learned
nonlinear ranking rather than a linear surrogate or chance.

The improvement is largest on inference-server GEMMs
(\(1.24\times\)--\(1.29\times\) for MLP and XGBoost over nvMMH), which
contain the batched speech and language-modeling layers DeepBench was designed
to stress. Training-split problems show a smaller but still positive margin
(\(\approx 1.07\times\)--\(1.09\times\)). 

\begin{table}[t]
\centering
\caption{DeepBench throughput generalization on \(151\) problems. Speedup is
geometric-mean throughput relative to nvMMH.}
\label{tab:deepbench}
\footnotesize
\setlength{\tabcolsep}{4pt}
\begin{tabular}{@{}llrr@{}}
\toprule
Model & Features & vs.\ nvMMH & Win \% \\
\midrule
MLP MSE & Full & 1.12$\times$ & 96.0\% \\
MLP MSE & Structural only & 1.13$\times$ & 92.7\% \\
XGBoost MSE & Full & 1.12$\times$ & 84.8\% \\
XGBoost MSE & Structural only & 1.05$\times$ & 59.6\% \\
Ridge & Full & 0.83$\times$ & 40.4\% \\
Ridge & Structural only & 0.27$\times$ & 0.0\% \\
Random & --- & 0.22$\times$ & 0.0\% \\
\midrule
nvMMH & --- & --- & --- \\
\bottomrule
\end{tabular}
\end{table}

\begin{table}[t]
\centering
\caption{DeepBench speedup vs.\ nvMMH by catalogue split.}
\label{tab:deepbench-splits}
\footnotesize
\setlength{\tabcolsep}{4pt}
\begin{tabular}{@{}lrrrr@{}}
\toprule
Split & $n$ & MLP (full) & MLP (struct.) & XGBoost (full) \\
\midrule
Training & 107 & 1.09$\times$ & 1.09$\times$ & 1.07$\times$ \\
Inference server & 39 & 1.25$\times$ & 1.28$\times$ & 1.29$\times$ \\
Inference device & 5 & 1.08$\times$ & 1.00$\times$ & 1.04$\times$ \\
\bottomrule
\end{tabular}
\end{table}
\fi

\ifconf
\else
\ifconf
\section{Additional Training and Evaluation Results}
\label{sec:app-eval-details}

This section provides additional analysis of the model-training objectives,
feature ablations, and the held-out exhaustive evaluation discussed in
Section~\ref{sec:evaluation}.

\subsection{Training Objective Comparison}
\label{sec:app-eval-training}

We train both the MLP and XGBoost models using three objectives.
For the MLP, we compare mean-squared error (MSE), RankNet, and LambdaRank;
for XGBoost, we compare squared-error regression, pairwise ranking, and
NDCG listwise ranking.  Each objective is evaluated with both the full
hardware-aware representation and the structural-only feature set.

Figure~\ref{fig:training-mlp} and Figure~\ref{fig:training-xgb} show
validation NDCG@10 throughout training.  All objectives converge stably.
With the full feature set, MSE achieves the highest validation NDCG@10 for
both MLP (0.968) and XGBoost (0.960).  With structural features, MSE again
performs best, reaching 0.969 for MLP and 0.929 for XGBoost.

\begin{figure}[t]
    \centering
    \includegraphics[width=0.8\linewidth]{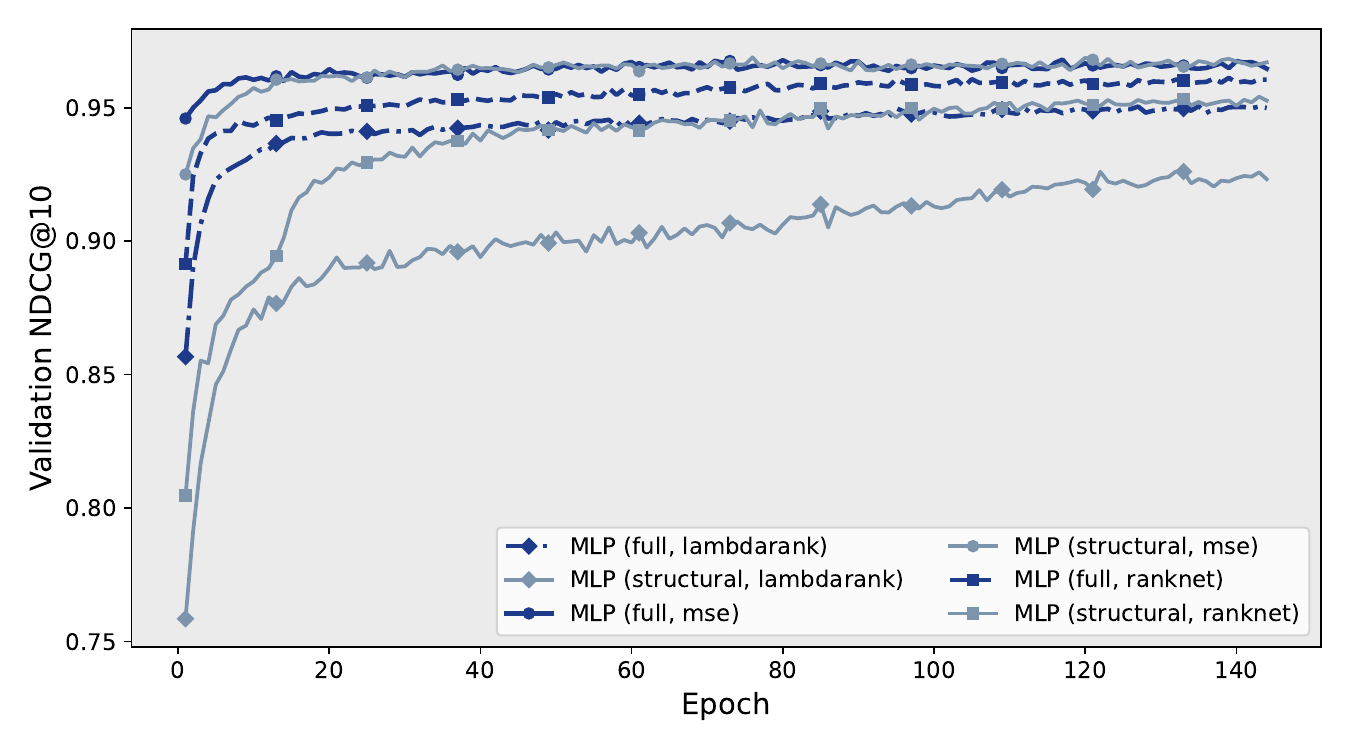}
    \caption{MLP validation NDCG@10.}
    \label{fig:training-mlp}
\end{figure}

\begin{figure}[t]
    \centering
    \includegraphics[width=0.8\linewidth]{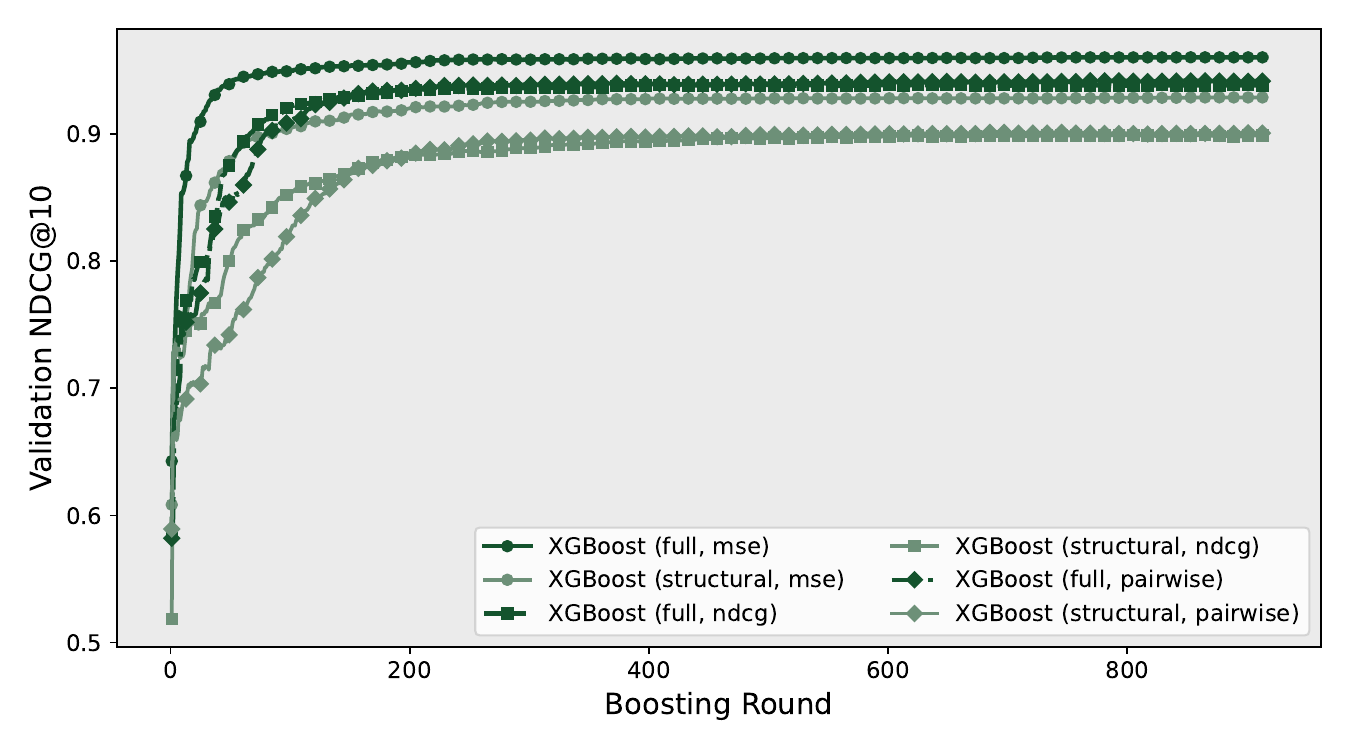}
    \caption{XGBoost validation NDCG@10.}
    \label{fig:training-xgb}
\end{figure}

Validation NDCG@10 is not perfectly aligned with held-out selection regret.
In particular, MLP RankNet achieves slightly lower validation NDCG than MSE,
but attains the lowest mean regret on the exhaustive evaluation set.  This
suggests that validation ranking quality is useful for monitoring training,
but does not fully capture the cost of errors among the highest-performing
candidates.

Figure~\ref{fig:full-vs-struct-obj} compares the full and structural
representations across all objectives.  Hardware-aware features consistently
reduce held-out regret, although the magnitude of the improvement depends on
the model family. The gain is relatively modest for the MLP, while XGBoost
benefits substantially more from the derived hardware features.

\subsection{Detailed Exhaustive Evaluation}
\label{sec:app-exhaustive}

Table~\ref{tab:eval-main} reports detailed selection quality on the evaluation dataset.

\begin{table}[t]
  \centering
  \caption{Selection quality on 68 exhaustively measured shape--layout groups.}
  \label{tab:eval-main}
  \footnotesize
  \setlength{\tabcolsep}{4pt}
  \begin{tabular}{@{}p{0.34\linewidth}rrrrr@{}}
    \toprule
    Method & Mean & Median & Within 1\% & Within 5\% & Top-1 \\
    \midrule
    MLP (full)       & 6.2\%  & 2.6\%  & 36.8\% & 63.2\% & 27.9\% \\
    MLP (structural)     & 7.4\%  & 3.1\%  & 32.4\% & 55.9\% & 26.5\% \\
    XGBoost (full)       & 6.4\%  & 4.9\%  & 25.0\% & 52.9\% & 19.1\% \\
    XGBoost (structural) & 10.7\% & 8.3\%  & 17.6\% & 39.7\% & 13.2\% \\
    nvMMH (best-of-$K$)      & 12.0\% & 10.2\% & 0.0\%  & 25.0\% & 0.0\% \\
    nvMMH (top-1)            & 17.3\% & 14.6\% & 0.0\%  & 10.3\% & 0.0\% \\
    Ridge (full)         & 23.7\% & 20.0\% & 4.4\%  & 16.2\% & 1.5\% \\
    Ridge (structural)       & 60.8\% & 63.6\% & 0.0\%  & 0.0\%  & 0.0\% \\
    Analytical score         & 63.7\% & 63.8\% & 0.0\%  & 0.0\%  & 0.0\% \\
    Random                   & 71.5\% & ---     & ---     & ---     & --- \\
    \bottomrule
  \end{tabular}
\end{table}

\begin{figure}[t]
  \centering
  \includegraphics[width=\linewidth]{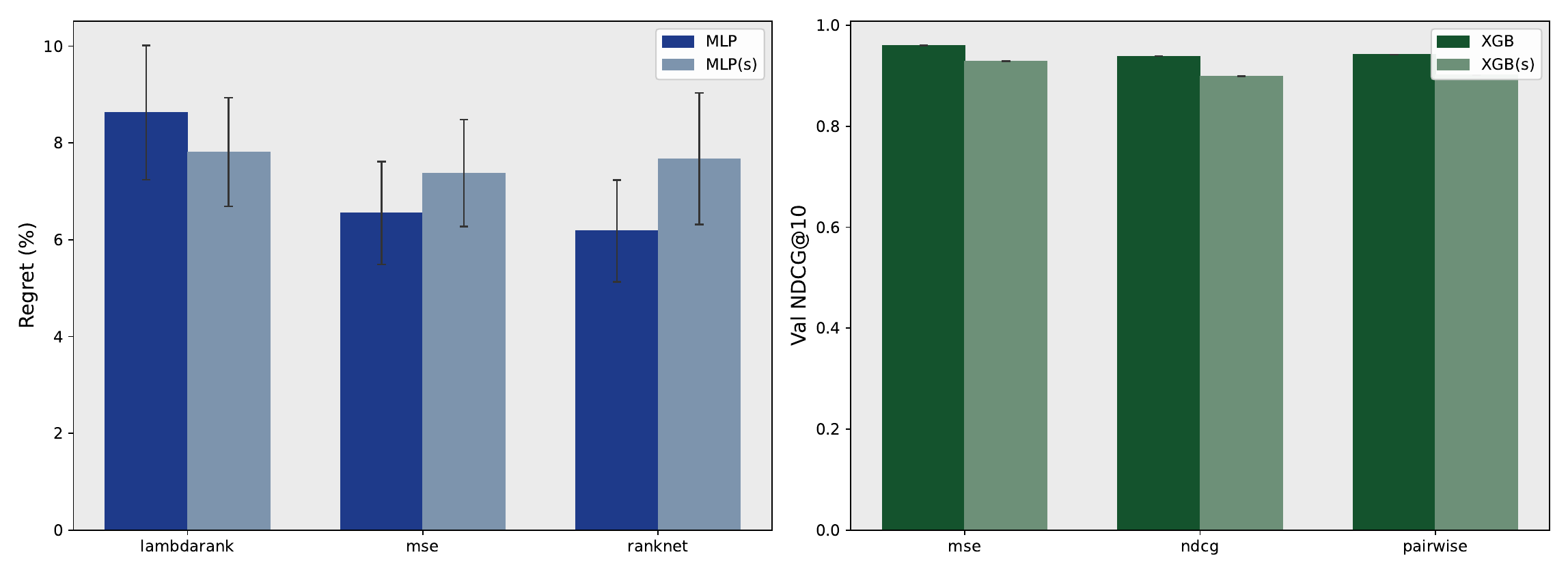}
  \caption{Comparison of full hardware-aware and structural-only features
  across training objectives.}
  \label{fig:full-vs-struct-obj}
\end{figure}

The learned selectors are substantially more likely to recover near-oracle
configurations.  MLP RankNet selects a kernel within \(5\%\) of the oracle
on \(63.2\%\) of groups, while XGBoost MSE does so on \(52.9\%\).
By comparison, the top-1 nvMMH recommendation is within \(5\%\) of the
oracle on only \(10.3\%\) of groups.

We additionally report an oracle over nvMMH's top-\(K\) candidate list.
Even when selecting the fastest measured candidate from that shortlist,
only \(25.0\%\) of groups fall within \(5\%\) of oracle throughput.
This indicates that a substantial fraction of nvMMH's error arises before
the final choice among its recommended candidates.

\subsection{Why is a learned model necessary?}
We next ask whether the hardware-aware representation can support strong kernel
selection without a nonlinear learned model. We compare against two simpler
alternatives: a validation-selected analytical score combining a small set of
mechanistic hardware terms, and a ridge regressor that linearly scores
candidates using the same full or structural feature representations as the
learned selectors (Table~\ref{tab:eval-main}).

The analytical score performs poorly, reaching \(63.7\%\) mean regret and never
selecting a configuration within \(5\%\) of the oracle across the 68 exhaustive
evaluation groups. This is despite several of its constituent hardware-aware
features showing clear within-group correlation with throughput on the training
set. Thus, identifying individually informative mechanisms is not sufficient:
a compact fixed composition of those mechanisms does not recover the
high-performance tail of the configuration space.

The ridge baseline isolates the value of the representation under a fixed linear
model family. Using structural features alone yields \(60.8\%\) mean regret,
while adding the hardware-aware features reduces regret to \(23.7\%\). This
large improvement shows that the proposed representation exposes substantial
selection signal even to a simple linear predictor. Nevertheless, the linear
hardware-aware model remains worse than nvMMH (\(17.3\%\)) and far behind the
nonlinear MLP and XGBoost selectors (\(6.2\%\) and \(6.4\%\)).

Together, these results separate representation quality from model expressivity.
The hardware-aware features make the ranking problem substantially easier, but
a single global linear mapping is insufficient for strong selection. The
remaining gap indicates that accurate ranking depends on nonlinear and
problem-dependent interactions between work decomposition, memory behavior,
resource pressure, and pipeline behavior, which the MLP and XGBoost models are
able to capture.

\subsection{Near-oracle selection rate.}
\label{sec:app-within5}

Figure~\ref{fig:eval-within5} reports the fraction of evaluation groups for which the selected kernel is within \(5\%\) of the exhaustive oracle. MLP RankNet with hardware-aware features reaches this threshold on \(63.2\%\) of groups, compared with \(55.9\%\) for the structural MLP, \(52.9\%\) for hardware-aware XGBoost, and \(39.7\%\) for structural XGBoost. nvMMH reaches the same threshold on only \(10.3\%\) of groups with its top-1 recommendation, or \(25.0\%\) when selecting the best measured kernel from its top-\(K\) shortlist. This metric complements mean regret by showing how frequently each selector returns a practically near-optimal configuration rather than relying on a small number of large or small errors.

\begin{wrapfigure}{l}{0.49\textwidth}
\centering
\includegraphics[width=0.49\textwidth]{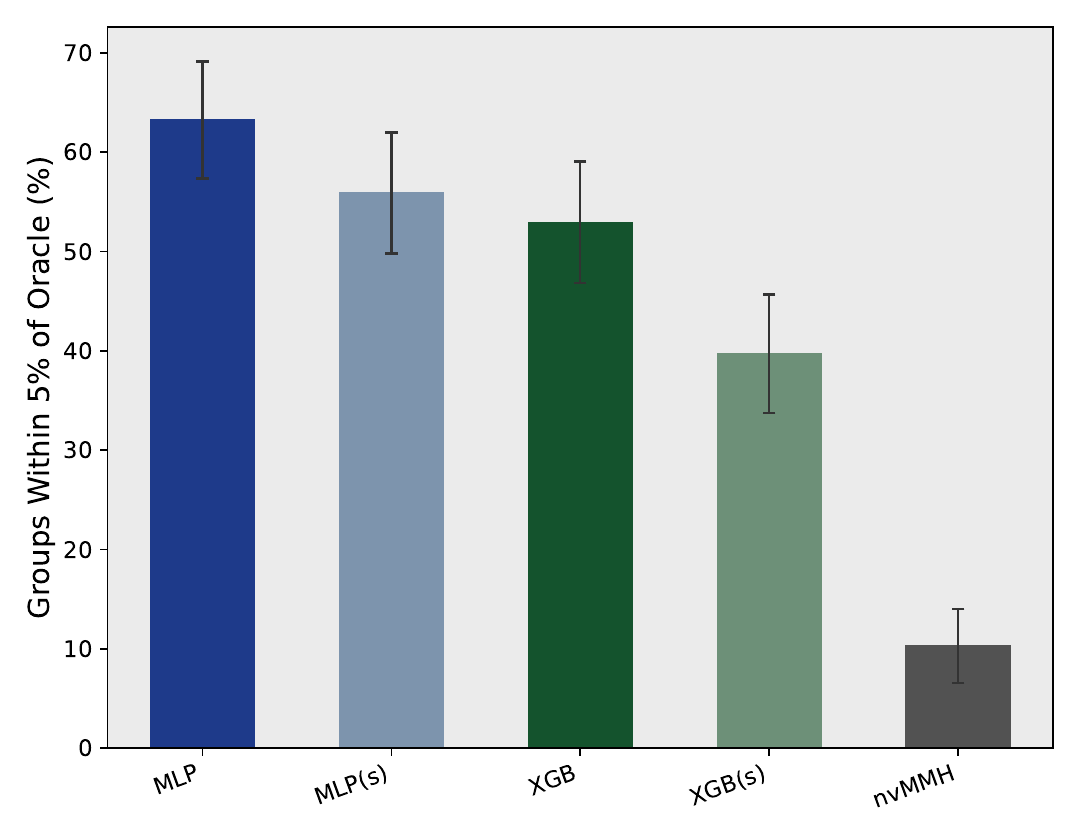}
\caption{Fraction of exhaustive evaluation groups selected within \(5\%\)
  of oracle throughput.}
\label{fig:eval-within5}
\end{wrapfigure}

\subsection{Performance by GEMM Regime}
\label{sec:app-eval-regime}

Selection quality depends on problem geometry.
nvMMH performs worse across all geometric regimes, with mean regret between
approximately \(12\%\) and \(19\%\).  The same qualitative pattern appears
for both full and structural feature sets, although the hardware-aware
representation consistently reduces regret.

Figure~\ref{fig:eval-regime-best} compares the best full and structural
variants of each model family.  The hardware-aware representation improves
both mean regret and the fraction of near-oracle selections, with the largest
gain for XGBoost.

\begin{figure}[t]
  \centering
  \includegraphics[width=\linewidth]{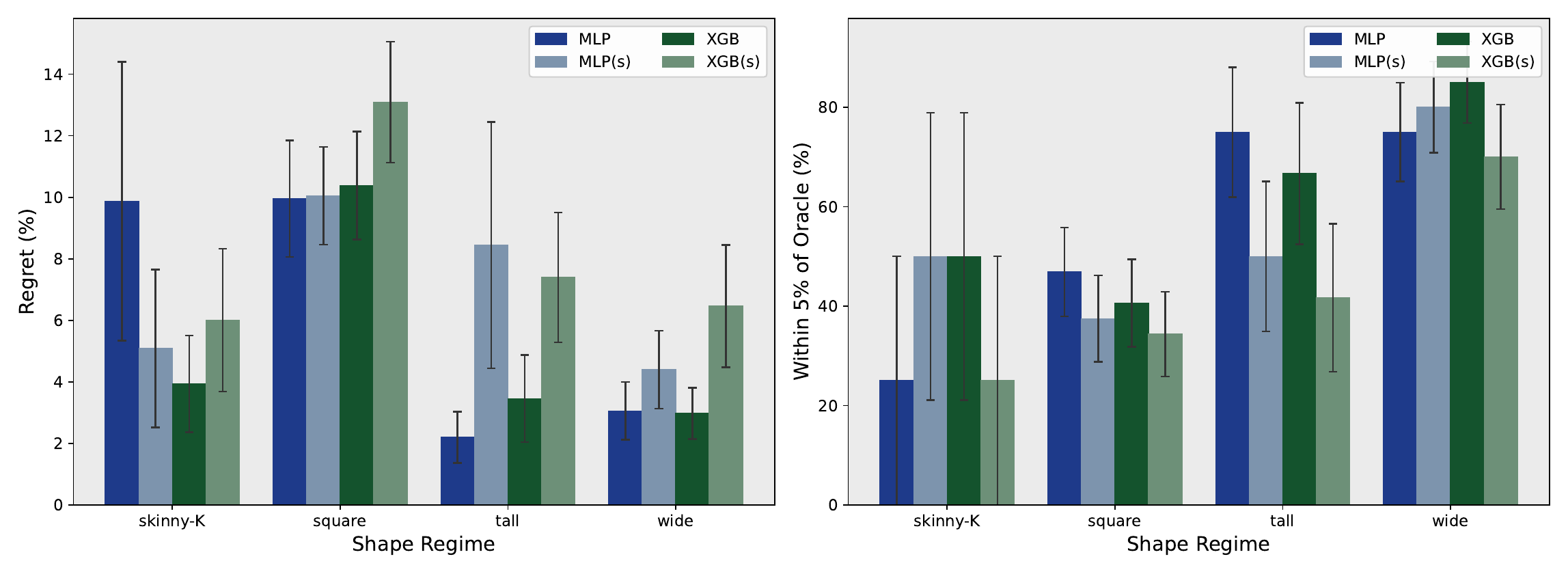}
  \caption{Selection regret and within-\(5\%\) rate by regime for the best
  full and structural variants of each model family.}
  \label{fig:eval-regime-best}
  \vspace{-1em}
\end{figure}

\subsection{Broad GEMM Evaluation Details}
\label{sec:app-eval-gemms}

\begin{table}[t]
  \centering
  \caption{Broad GEMM evaluation on \(8{,}000\) held-out problems.}
  \label{tab:gemm-broad}
  \footnotesize
  \setlength{\tabcolsep}{4pt}
  \begin{tabular}{@{}llrrrr@{}}
    \toprule
    Model & Features & Geo-mean [Tflop/s] & Geo-mean \% roof & vs.\ nvMMH & Win \% \\
    \midrule
    MLP MSE     & Full            & 42.9 & 11.9\% & 1.14$\times$ & 79.6\% \\
    MLP MSE     & Structural only & 41.2 & 11.5\% & 1.10$\times$ & 73.9\% \\
    \midrule
    XGBoost MSE & Full            & 42.4 & 11.8\% & 1.13$\times$ & 77.4\% \\
    XGBoost MSE & Structural only & 41.0 & 11.4\% & 1.09$\times$ & 70.1\% \\
    \midrule
    nvMMH & --- & 37.6 & 10.5\% & --- & --- \\
    \bottomrule
  \end{tabular}
\end{table}

The broad evaluation uses \(2{,}000\) \((M,N,K)\) triples held out from the production BF16 autotuning databases and evaluates each under all four operand layouts (TN, TT, NN, NT), yielding \(8{,}000\) problems. Half of the shapes are fully \(32\)-aligned; the remainder are \(8\)-aligned but not \(32\)-aligned. The latter set tests generalization to off-grid dimensions that are largely absent from the regular training sweep. The performance results are in Table~\ref{tab:gemm-broad}.

We compare four learned selectors---MLP and XGBoost, each with the full hardware-aware representation and the structural-only ablation---against nvMMH. Each learned model returns a single rank-1 CUTLASS configuration. nvMMH emits a rank-1 tile and cluster recommendation but does not completely specify the CUTLASS mainloop, epilogue, and scheduler configuration. We therefore materialize all valid variants consistent with the nvMMH recommendation and report the best measured variant. This gives nvMMH an optimistic comparison relative to the learned selectors, which each receive only one selected kernel.

All candidates are compiled and benchmarked on the same GH200 node using the same warmup, iteration-count, and operand-rotation protocol as the training measurements. To isolate kernel ranking from tile-scheduler choices, every method uses the same rank-1 rasterization, swizzle, and split-\(K\) settings during benchmarking. nvMMH retains its recommended tile geometry, while the learned selectors retain their independently selected tile, schedule, cluster, and pipeline configuration.

For the roofline analysis in Figure~\ref{fig:roofline-broad-gemm}, we use a GH200 peak BF16 tensor-core throughput of \(989.5\,\mathrm{Tflop/s}\) and memory bandwidth of \(4\,\mathrm{TB/s}\). Arithmetic intensity is computed from the mathematical GEMM work and minimum operand traffic. Each light point corresponds to one measured GEMM, while the bold curves show binned median throughput.

All methods achieve \(100\%\) execution coverage on the \(8{,}000\) problems. The hardware-aware MLP reaches \(42.9\,\mathrm{Tflop/s}\) geometric-mean throughput compared with \(37.6\,\mathrm{Tflop/s}\) for nvMMH, corresponding to a \(1.14\times\) improvement. Hardware-aware XGBoost reaches \(42.4\,\mathrm{Tflop/s}\), or \(1.13\times\) nvMMH. The structural variants remain competitive at \(41.2\) and \(41.0\,\mathrm{Tflop/s}\), but both are consistently below their hardware-aware counterparts.

Measured per problem, the full-feature MLP outperforms nvMMH on \(79.6\%\) of cases and full-feature XGBoost on \(77.4\%\). Even the structural variants win on \(73.9\%\) and \(70.1\%\), respectively. Together with the exhaustive-oracle evaluation, this shows that the learned selectors' lower selection regret translates directly into higher realized throughput over a substantially broader problem distribution.

\subsection{Transfer Learning Details}
\label{sec:app-transfer}

For each target dtype, (FP32 and FP8 E4M3), we construct nested subsets
containing \(\{1,5,10,25,50,100\}\%\) of the 593 target-dtype base training
shapes. The same \((M,N,K)\) subsets are used for FP32 and FP8, while the
measured candidate kernels differ according to dtype-specific validity and
compilation behavior. The fractions therefore measure target-domain shape
coverage rather than equal numbers of kernel measurements. Table~\ref{tab:transfer-005}
reports aggregate statistics at \(5\%\) of the
training shapes (30 of 593 base shapes).

We reuse the model architectures selected for the
BF16 study. MLP transfer initializes from the corresponding BF16 checkpoint,
while XGBoost is warm-started from the BF16 model by adding 200 trees. We
evaluate both the complete hardware-aware representation and the structural
ablation. Evaluation uses a disjoint held-out set and the same benchmarking
protocol as Section~\ref{sec:gemm-eval}. Learned selectors emit one rank-1
kernel, while nvMMH is evaluated using its best valid schedule realization.

At full target-dtype coverage, FP32 transfer reaches \(1.24\times\)
geometric-mean speedup for both hardware-aware MLP and XGBoost, with win rates
of \(74.6\%\) and \(78.4\%\), respectively. Structural variants remain strong
but consistently lower, reaching \(1.18\times\) and \(1.20\times\).
FP8 remains more challenging, particularly in execution coverage, but the
hardware-aware MLP and XGBoost still achieve \(1.19\times\) and
\(1.17\times\) geometric-mean speedup over nvMMH.

\begin{table}[t]
  \centering
  \caption{Cross-precision transfer at \(5\%\) of the target training
  shapes.}
  \vspace{-0.5em}
  \label{tab:transfer-005}
  \footnotesize
  \setlength{\tabcolsep}{4pt}
  \begin{tabular}{@{}llrrrr@{}}
    \toprule
    Dtype & Method & Coverage & Geo-mean speedup & Win \% \\
    \midrule
    FP32 & MLP (full)       & 77.1\% & 1.20$\times$ & 58.7\% \\
     & XGBoost (full) & 97.4\% & 1.22$\times$ & 76.3\% \\
     & MLP (struct.)    & 80.8\% & 1.19$\times$ & 62.3\% \\
     & XGBoost (struct.) & 98.8\% & 1.16$\times$ & 69.4\% \\
    \midrule
    FP8 E4M3 & MLP (full)       & 100.0\% & 1.12$\times$ & 69.8\% \\
     & XGBoost (full) & 68.2\% & 1.17$\times$ & 47.3\% \\
     & MLP (struct.)    & 100.0\% & 1.12$\times$ & 69.7\% \\
     & XGBoost (struct.) & 71.2\% & 1.11$\times$ & 40.7\% \\
    \bottomrule
  \end{tabular}
\end{table}

The low-data regime is particularly informative. At \(5\%\) of the target
shape pool, the hardware-aware FP32 models already exceed nvMMH by roughly
\(20\%\)--\(22\%\), despite observing only 30 target-dtype GEMM geometries.
The corresponding FP8 models also exceed nvMMH, demonstrating that much of the
useful structure learned from BF16 survives the precision change and can be
adapted with comparatively little target-domain data.

\paragraph{Fusion transfer.}
We apply the same finetuning protocol to FP16 epilogue-fusion GEMMs.
Selectors are warm-started from the unfused BF16 checkpoints and finetuned on
nested subsets of the 593-shape fusion training corpus, using the same
\(\{1,5,10,25,50,100\}\%\) fractions as the precision-transfer study.
We evaluate \(1{,}000\) held-out problems with in-vocabulary fusion kinds
assigned at evaluation time.

At full fusion training coverage, hardware-aware XGBoost matches nvMMH
(\(1.00\times\) geometric-mean speedup), while the
hardware-aware MLP reaches \(0.97\times\).
Structural variants remain below at \(0.94\times\).
All configurations achieve \(100\%\) execution coverage.

The low-data fusion regime mirrors the precision-transfer trend. With only
\(10\%\) of the fusion corpus (59 shapes), hardware-aware MLP and XGBoost
already reach \(1.00\times\) geometric-mean speedup over nvMMH, while
structural models remain lower.
The hardware-aware MLP peaks at \(1.01\times\) with \(25\%\) coverage before
settling near parity.

\subsection{DeepBench Study Details}

Our catalogue contains \(151\) unique \((M,N,K,\mathrm{layout})\) problems
after deduplication and constraining $M,N,K\geq 32$.  The evaluation set contains
\(107\) training, \(39\) inference-server, and \(5\) inference-device problems.
%
%
Table~\ref{tab:deepbench} summarizes
the results. All learned selectors achieve \(100\%\) execution coverage on the
retained problems. Hardware-aware MLP and XGBoost improve geometric-mean
throughput by \(12\%\)--\(13\%\) over nvMMH and win on \(85\%\)--\(96\%\) of
problems. The structural MLP matches the full model (\(1.13\times\)),
consistent with the observation that very large MLP models with structural features extrapolate rather
robustly. Ridge and random
baselines remain well below nvMMH, confirming that the gains require learned
nonlinear ranking.
The improvement is largest on inference-server GEMMs
(\(1.24\times\)--\(1.29\times\) for MLP and XGBoost over nvMMH), which
contain the batched speech and language-modeling layers DeepBench was designed
to stress. Training-split problems show a smaller but still positive margin
(\(\approx 1.07\times\)--\(1.09\times\)). 
%

\begin{table}[t]
\centering
\caption{DeepBench throughput speedup vs. nvMMH.}
\label{tab:deepbench}
  \vspace{-0.5em}
\footnotesize
\setlength{\tabcolsep}{4pt}
\begin{tabular}{@{}llrr@{}}
\toprule
Model & Features & vs.\ nvMMH & Win \% \\
\midrule
MLP MSE & Full & 1.12$\times$ & 96.0\% \\
MLP MSE & Structural only & 1.13$\times$ & 92.7\% \\
XGBoost MSE & Full & 1.12$\times$ & 84.8\% \\
XGBoost MSE & Structural only & 1.05$\times$ & 59.6\% \\
Ridge & Full & 0.83$\times$ & 40.4\% \\
Ridge & Structural only & 0.27$\times$ & 0.0\% \\
Random & --- & 0.22$\times$ & 0.0\% \\
\midrule
nvMMH & --- & --- & --- \\
\bottomrule
\end{tabular}
\end{table}
\begin{table}[t!]
\centering
\caption{DeepBench speedup vs.\ nvMMH by catalogue split.}
\vspace{-1em}
\label{tab:deepbench-splits}
\footnotesize
\setlength{\tabcolsep}{4pt}
\begin{tabular}{@{}lrrrr@{}}
\toprule
Split & $n$ & MLP (full) & XGBoost (full) \\
\midrule
Training & 107 & 1.09$\times$  & 1.07$\times$ \\
Inference server & 39 & 1.25$\times$ 1.29$\times$ \\
Inference device & 5 & 1.08$\times$ & 1.04$\times$ \\
\bottomrule
\end{tabular}
\vspace{-2em}
\end{table}
\fi

\fi

\section{Related Work}



Learned cost models guide iterative search over tensor-program spaces \citec{li2020adatune, zheng2020ansor, choi2022learning, zhang2026wavetune, bai2025learned} in frameworks like AutoTVM \citec{chen2019learningoptimizetensorprograms, wu2023autotuningapachetvmbasedscientific} and Ansor \citec{zheng2020ansor}. While corpora like TenSet \citec{zheng2021tenset} and optimization methods like AdaTune \citec{li2020adatune} or TLP \citec{zhai2023tlp} improve search efficiency, our model scores the entire valid CUTLASS configuration catalogue in a single batch to directly select the optimal implementation. While device-transfer frameworks~\citec{ryu2021metatune, zhai2023tlp, bai2025learned} rely on complex learned program representations, instead we show that explicit, mechanism-level hardware features provide a strong inductive bias that improves transfer accuracy with fewer target-domain measurements.

While predictive models like ISAAC~\citec{tillet2018inputawareautotuningcomputeboundhpc}, CUTLASS-tailor~\citec{yu2023tailoring}, and related low-bit selectors~\citec{guo2024low, guo2026leveraging} show that GEMM execution cost is learnable from configuration descriptors~\citec{xiaoteng2024understandinggemmperformanceenergy}, our model uses analytical hardware-behavior features to directly rank and select optimal configurations. Manual analytical models~\citec{ran2026kernelbandsteeringllmbasedkernel, jarmusch2026microbenchmarkdrivenanalyticalperformancemodeling} and heuristics~\citec{Improvin92:online, NVIDIAMa44:online, swann2025tritonblas, zhang2025hexcute} avoid training data by hardcoding hardware-mapping rules, but we learn these complex interactions from behavioral features instead.

\ifconf
\else
Architectural innovations like Stream-K~\citec{osama2023streamkworkcentricparalleldecomposition} and FlashAttention-3~\citec{shah2024flashattention3fastaccurateattention} demonstrate that GPU efficiency relies on work partitioning and pipeline coordination rather than just raw computation. We model these underlying mechanisms as compile-time features to select optimal configurations from the vast spaces exposed by CUTLASS.
\fi

\section{Conclusion}

We presented a hardware-aware approach to learned CUTLASS kernel selection. Rather than asking a model to infer GPU behavior from raw configuration parameters, we augment each candidate with statically computable estimates of induced execution behavior. Using 4.9 million measured CUTLASS kernels, the resulting representation achieves mean selection regret of \(6.2\%\) with an MLP and \(6.4\%\) with XGBoost, compared with \(17.3\%\) for the top-1 nvMMH recommendation. Its benefit across both model families indicates that the improvement comes from the representation rather than a particular predictor architecture.

More broadly, our results suggest that learned performance models benefit from representing what a configuration does to the hardware, rather than only what the configuration is. Raw parameters often hide effects whose meaning changes with workload and execution regime, while fully analytical models require brittle architecture-specific rules. Hardware-aware representations provide an intermediate abstraction: architectural knowledge exposes the relevant mechanisms, while data learns their interactions. As accelerator configuration spaces grow, such behavioral representations may provide a reusable basis for performance models across precisions, operators, and architectures. The central implication is that better performance models may come less from larger models than from better representations aligned with hardware execution.

\ifconf
\else
\section*{AI Acknowledgment}
\fi




In this work, we used generative AI tools to assist with the implementation and debugging of benchmarking and analysis code, interpret intermediate experimental results, and assist with language editing of the manuscript. We did not use generative AI tools to generate synthetic experimental data, formulate or prove mathematical claims, or fabricate experimental measurements or results. Tasks involving surveys, interviews, transcription of research material, and qualitative or thematic data analysis were not applicable to this work.

All AI-assisted research ideas, methodological suggestions, code, analyses, and manuscript text were reviewed by the authors. AI-assisted code was inspected and validated through compilation, testing, and comparison against measured data before being used in the experimental pipeline. Literature suggestions and technical claims were checked against the corresponding papers or documentation. Experimental results were obtained from the authors' own measurement infrastructure, and their interpretation and the final scientific conclusions were determined by the authors. We take responsibility for the final content of this work, including all text, claims, code, and artifacts produced with the aid of generative AI.

\ifconf
\else
\section*{Code}

Our code is available online at \url{https://github.com/spcl/cutlass-selector}.
\fi

\ifconf
\else
\subsection*{Acknowledgments}
This project received support by the ERC PSAP project (Grant Agreement No. 101002047). The research was conducted as part of the FastTrackAI project at the Singapore-ETH Centre, which was established collaboratively between ETH Zurich and the National Research Foundation, Singapore. Additionally, this research is supported by the National Research Foundation, Singapore (NRF), and the Ministry of Digital Development and Information (MDDI) under the AI Visiting Professorship (Award No. AIVP-2025-005). We also thank the Swiss National Supercomputing Center (CSCS) for providing the computational resources used in this work. 
\fi

\ifconf
\else
\bibliographystyle{acmref}
\fi
\sloppy \bibliography{iclr2027_conference}

@inproceedings{li2020adatune,
author = {Li, Menghao and Zhang, Minjia and Wang, Chi and Li, Mingqin},
title = {AdaTune: adaptive tensor program compilation made efficient},
year = {2020},
isbn = {9781713829546},
publisher = {Curran Associates Inc.},
address = {Red Hook, NY, USA},
booktitle = {Proceedings of the 34th International Conference on Neural Information Processing Systems},
articleno = {1241},
numpages = {13},
location = {Vancouver, BC, Canada},
series = {NIPS '20}
}

@inproceedings{zheng2020ansor,
author = {Zheng, Lianmin and Jia, Chengfan and Sun, Minmin and Wu, Zhao and Yu, Cody Hao and Haj-Ali, Ameer and Wang, Yida and Yang, Jun and Zhuo, Danyang and Sen, Koushik and Gonzalez, Joseph E. and Stoica, Ion},
title = {Ansor: generating high-performance tensor programs for deep learning},
year = {2020},
isbn = {978-1-939133-19-9},
publisher = {USENIX Association},
address = {USA},
booktitle = {Proceedings of the 14th USENIX Conference on Operating Systems Design and Implementation},
articleno = {49},
numpages = {17},
series = {OSDI'20}
}

@article{choi2022learning,
  title={Learning from distinctive candidates to optimize reduced-precision convolution program on tensor cores},
  author={Choi, Junkyeong and Kwon, Hyucksung and Lee, Woongkyu and Choi, Jungwook and Lim, Jieun},
  journal={arXiv preprint arXiv:2202.06819},
  year={2022}
}

@article{zhang2026wavetune,
  title={WaveTune: Wave-aware Bilinear Modeling for Efficient GPU Kernel Auto-tuning},
  author={Zhang, Kaixuan and Ding, Chutong and Qian, Shiyou and Wang, Luping and Cao, Jian and Xue, Guangtao and Huang, Cheng and Yang, Guodong and Zhang, Liping},
  journal={arXiv preprint arXiv:2604.10187},
  year={2026}
}

@inproceedings{wu2023autotuningapachetvmbasedscientific,
author = {Wu, Xingfu and Paramasivam, Praveen and Taylor, Valerie},
title = {Autotuning Apache TVM-based Scientific Applications Using Bayesian Optimization},
year = {2023},
isbn = {9798400707858},
publisher = {Association for Computing Machinery},
address = {New York, NY, USA},
url = {https://doi.org/10.1145/3624062.3626079},
doi = {10.1145/3624062.3626079},
booktitle = {Proceedings of the SC '23 Workshops of the International Conference on High Performance Computing, Network, Storage, and Analysis},
pages = {29–35},
numpages = {7},
location = {Denver, CO, USA},
series = {SC-W '23}
}

@article{bai2025learned,
  title={A learned performance model with transfer learning across GPUs on tensorized instructions},
  author={Bai, Yang and Li, Mingjun and Xu, Wendong and Yu, Bei},
  journal={IEEE Transactions on Parallel and Distributed Systems},
  volume={36},
  number={9},
  pages={1904--1919},
  year={2025},
  publisher={IEEE}
}

@inproceedings{zheng2021tenset,
  title={Tenset: A large-scale program performance dataset for learned tensor compilers},
  author={Zheng, Lianmin and Liu, Ruochen and Shao, Junru and Chen, Tianqi and Gonzalez, Joseph E and Stoica, Ion and Ali, Ameer Haj},
  booktitle={Thirty-fifth Conference on Neural Information Processing Systems Datasets and Benchmarks Track (Round 1)},
  year={2021}
}

@inproceedings{zhai2023tlp,
author = {Zhai, Yi and Zhang, Yu and Liu, Shuo and Chu, Xiaomeng and Peng, Jie and Ji, Jianmin and Zhang, Yanyong},
title = {TLP: A Deep Learning-Based Cost Model for Tensor Program Tuning},
year = {2023},
isbn = {9781450399166},
publisher = {Association for Computing Machinery},
address = {New York, NY, USA},
url = {https://doi.org/10.1145/3575693.3575737},
doi = {10.1145/3575693.3575737},
booktitle = {Proceedings of the 28th ACM International Conference on Architectural Support for Programming Languages and Operating Systems, Volume 2},
pages = {833–845},
numpages = {13},
location = {Vancouver, BC, Canada},
series = {ASPLOS 2023}
}

@misc{nvidia_cutlass,
author = {Thakkar, Vijay and Ramani, Pradeep and Cecka, Cris and Shivam, Aniket and Lu, Honghao and Yan, Ethan and Kosaian, Jack and Hoemmen, Mark and Wu, Haicheng and Kerr, Andrew and Nicely, Matt and Merrill, Duane and Blasig, Dustyn and Atluri, Aditya and Qiao, Fengqi and Majcher, Piotr and Springer, Paul and Hohnerbach, Markus and Wang, Jin and Gupta, Manish},
month = {1},
title = {CUTLASS},
url = {https://github.com/NVIDIA/cutlass/tree/v3.0.0},
year = {2023}
}

@misc{CUTLASS368:online,
author = {Vijay Thakkar and Cris Cecka and Tejash Shah and Jay Shah and Paul VanKoughnett and Ryo Asai},
  title = {CUTLASS 3.x: Orthogonal, Reusable, and Composable Abstractions for GEMM Kernel Design | NVIDIA Technical Blog},
  url = "https://developer.nvidia.com/blog/cutlass-3-x-orthogonal-reusable-and-composable-abstractions-\\for-gemm-kernel-design/",
month = {7},
year = {2025},
}

@article{bikshandi2023developing,
  title={Developing CUDA Kernels for Accelerated Matrix Multiplication on NVIDIA Hopper Architecture using the CUTLASS Library},
  author={Bikshandi, Ganesh and Shah, Jay},
  journal={Colfax Research},
  year={2023}
}

@misc{Improvin92:online,
author = {Harrison Barclay and Ian Tramble and  Michał Kukuła and Michel Migdal and Nicholai Tukanov and Leopold Cambier},
  title = {Improving GEMM Kernel Auto-Tuning Efficiency on NVIDIA GPUs with Heuristics and CUTLASS 4.2 | NVIDIA Technical Blog},
  url = "https://developer.nvidia.com/blog/improving-gemm-kernel-auto-tuning-efficiency-on-nvidia-gpus-wi\\th-heuristics-and-cutlass-4-2/",
month = {9},
year = {2025},
}

@article{swann2025tritonblas,
  title={tritonblas: Triton-based analytical approach for gemm kernel parameter selection},
  author={Swann, Ryan and Osama, Muhammad and Guo, Xiaohu and Nelson, Bryant and Zhang, Lixun and Brown, Alex and Ong, Yen and Yazdani, Ali and Siddens, Sean and Dasika, Ganesh and others},
  journal={arXiv preprint arXiv:2512.04226},
  year={2025}
}

@article{ryu2021metatune,
  title={Metatune: Meta-learning based cost model for fast and efficient auto-tuning frameworks},
  author={Ryu, Jaehun and Sung, Hyojin},
  journal={arXiv preprint arXiv:2102.04199},
  year={2021}
}

@INPROCEEDINGS{luo2024benchmarkingdissectingnvidiahopper,
  author={Luo, Weile and Fan, Ruibo and Li, Zeyu and Du, Dayou and Wang, Qiang and Chu, Xiaowen},
  booktitle={2024 IEEE International Parallel and Distributed Processing Symposium (IPDPS)}, 
  title={Benchmarking and Dissecting the Nvidia Hopper GPU Architecture}, 
  year={2024},
  volume={},
  number={},
  pages={656-667},
  doi={10.1109/IPDPS57955.2024.00064}}

@inproceedings{tillet2018inputawareautotuningcomputeboundhpc,
author = {Tillet, Philippe and Cox, David},
title = {Input-aware auto-tuning of compute-bound HPC kernels},
year = {2017},
isbn = {9781450351140},
publisher = {Association for Computing Machinery},
address = {New York, NY, USA},
url = {https://doi.org/10.1145/3126908.3126939},
doi = {10.1145/3126908.3126939},
booktitle = {Proceedings of the International Conference for High Performance Computing, Networking, Storage and Analysis},
articleno = {43},
numpages = {12},
location = {Denver, Colorado},
series = {SC '17}
}

@INPROCEEDINGS{guo2024low,
  author={Guo, Hong and Guo, Nianhui and Meinel, Christoph and Yang, Haojin},
  booktitle={2024 IEEE International Symposium on Parallel and Distributed Processing with Applications (ISPA)}, 
  title={Low-bit CUTLASS GEMM Template Auto-tuning using Neural Network}, 
  year={2024},
  volume={},
  number={},
  pages={394-401},
  doi={10.1109/ISPA63168.2024.00057}}

@INPROCEEDINGS{yu2023tailoring,
  author={Yu, Yongseung and Son, Donghyun and Lee, Younghyun and Park, Sunghyun and Ryu, Giha and Cho, Myeongjin and Seo, Jiwon and Park, Yongjun},
  booktitle={2023 IEEE 41st International Conference on Computer Design (ICCD)}, 
  title={Tailoring CUTLASS GEMM using Supervised Learning}, 
  year={2023},
  volume={},
  number={},
  pages={465-474},
  doi={10.1109/ICCD58817.2023.00077}}

@ARTICLE{guo2026leveraging,
  author={Guo, Hong and Guo, Nianhui and Meinel, Christoph and Yang, Haojin},
  journal={Big Data Mining and Analytics}, 
  title={Leveraging Large-Scale Data for Efficient Low-Bit CUTLASS GEMM Optimization via Neural Networks}, 
  year={2026},
  volume={9},
  number={2},
  pages={632-652},
  doi={10.26599/BDMA.2025.9020065}}

@misc{xiaoteng2024understandinggemmperformanceenergy,
      title={Understanding GEMM Performance and Energy on NVIDIA Ada Lovelace: A Machine Learning-Based Analytical Approach}, 
      author={Xiaoteng Liu and Pavly Halim},
      year={2024},
      eprint={2411.16954},
      archivePrefix={arXiv},
      primaryClass={cs.DC},
      url={https://arxiv.org/abs/2411.16954}, 
}

@misc{ran2026kernelbandsteeringllmbasedkernel,
      title={KernelBand: Steering LLM-based Kernel Optimization via Hardware-Aware Multi-Armed Bandits}, 
      author={Dezhi Ran and Shuxiao Xie and Mingfang Ji and Anmin Liu and Mengzhou Wu and Yuan Cao and Yuzhe Guo and Hao Yu and Linyi Li and Yitao Hu and Wei Yang and Tao Xie},
      year={2026},
      eprint={2511.18868},
      archivePrefix={arXiv},
      primaryClass={cs.LG},
      url={https://arxiv.org/abs/2511.18868}, 
}

@misc{jarmusch2026microbenchmarkdrivenanalyticalperformancemodeling,
      title={Microbenchmark-Driven Analytical Performance Modeling Across Modern GPU Architectures}, 
      author={Aaron Jarmusch and Sunita Chandrasekaran},
      year={2026},
      eprint={2605.04178},
      archivePrefix={arXiv},
      primaryClass={cs.DC},
      url={https://arxiv.org/abs/2605.04178}, 
}

@misc{NVIDIAMa44:online,
  author = {NVIDIA},
  title = {NVIDIA Matmul Heuristics — nvMatmulHeuristics},
  url = "https://docs.nvidia.com/cuda/nvidia-matmul-heuristics/",
  month = {},
  year = {2025},
}

@misc{zhang2025hexcute,
      title={Hexcute: A Compiler Framework for Automating Layout Synthesis in GPU Programs}, 
      author={Xiao Zhang and Yaoyao Ding and Bolin Sun and Yang Hu and Tatiana Shpeisman and Gennady Pekhimenko},
      year={2026},
      eprint={2504.16214},
      archivePrefix={arXiv},
      primaryClass={cs.LG},
      url={https://arxiv.org/abs/2504.16214}, 
}

@inproceedings{osama2023streamkworkcentricparalleldecomposition,
author = {Osama, Muhammad and Merrill, Duane and Cecka, Cris and Garland, Michael and Owens, John D.},
title = {Stream-K: Work-Centric Parallel Decomposition for Dense Matrix-Matrix Multiplication on the GPU},
year = {2023},
isbn = {9798400700156},
publisher = {Association for Computing Machinery},
address = {New York, NY, USA},
url = {https://doi.org/10.1145/3572848.3577479},
doi = {10.1145/3572848.3577479},
booktitle = {Proceedings of the 28th ACM SIGPLAN Annual Symposium on Principles and Practice of Parallel Programming},
pages = {429–431},
numpages = {3},
location = {Montreal, QC, Canada},
series = {PPoPP '23}
}

@inproceedings{shah2024flashattention3fastaccurateattention,
author = {Shah, Jay and Bikshandi, Ganesh and Zhang, Ying and Thakkar, Vijay and Ramani, Pradeep and Dao, Tri},
title = {FlashAttention-3: fast and accurate attention with asynchrony and low-precision},
year = {2024},
isbn = {9798331314385},
publisher = {Curran Associates Inc.},
address = {Red Hook, NY, USA},
booktitle = {Proceedings of the 38th International Conference on Neural Information Processing Systems},
articleno = {2193},
numpages = {28},
location = {Vancouver, BC, Canada},
series = {NIPS '24}
}

@misc{soi2025optimalsoftwarepipeliningwarp,
      title={Optimal Software Pipelining and Warp Specialization for Tensor Core GPUs}, 
      author={Rupanshu Soi and Rohan Yadav and Fredrik Kjolstad and Alex Aiken and Maryam Mehri Dehnavi and Michael Garland and Michael Bauer},
      year={2025},
      eprint={2512.18134},
      archivePrefix={arXiv},
      primaryClass={cs.PL},
      url={https://arxiv.org/abs/2512.18134}, 
}

@inproceedings{chen2019learningoptimizetensorprograms,
author = {Chen, Tianqi and Zheng, Lianmin and Yan, Eddie and Jiang, Ziheng and Moreau, Thierry and Ceze, Luis and Guestrin, Carlos and Krishnamurthy, Arvind},
title = {Learning to optimize tensor programs},
year = {2018},
publisher = {Curran Associates Inc.},
address = {Red Hook, NY, USA},
booktitle = {Proceedings of the 32nd International Conference on Neural Information Processing Systems},
pages = {3393–3404},
numpages = {12},
location = {Montr{\'e}al, Canada},
series = {NIPS'18}
}

@misc{baidu2016deepbench,
  author = {Baidu},
  title = {DeepBench: Benchmarking deep learning operations on different hardware},
  year = {2016},
  publisher = {GitHub},
  journal = {GitHub repository},
  howpublished = {\url{https://github.com/baidu-research/DeepBench}}
}

\end{document}